\documentclass[]{interact}

\usepackage{booktabs}
\usepackage[caption=false]{subfig}
\usepackage[numbers,sort&compress]{natbib}

\usepackage[usenames,dvipsnames,table]{xcolor}
\usepackage{hyperref}
\hypersetup{
    colorlinks=false,
    bookmarks=false,
    hidelinks=true,
    linkcolor=MidnightBlue,
    filecolor=magenta,   
    urlcolor=MidnightBlue,
    citecolor=MidnightBlue
}

\bibpunct[, ]{[}{]}{,}{n}{,}{,}
\makeatletter%
\def\NAT@def@citea{\def\@citea{\NAT@separator}}
\makeatother%

\newcommand{\fixed}[1]{\textcolor{black}{#1}}

\usepackage[normalem]{ulem}

\newcommand{\switchlanguage}[2]{%
  \ifx\paperlanguage\empty%
  #1%
  \else%
  #2%
  \fi%
}

\begin{document}

\articletype{REVIEW ARTICLE}
\title{Real-World Robot Applications of Foundation Models: A Review}

\author{
    Kento Kawaharazuka\textsuperscript{a}\textsuperscript{*}, 
    Tatsuya Matsushima\textsuperscript{b}\textsuperscript{*},\\
    Andrew Gambardella\textsuperscript{b}, 
    Jiaxian Guo\textsuperscript{b}, 
    Chris Paxton\textsuperscript{c},
    and Andy Zeng\textsuperscript{d}
    \thanks{\indent Email: \url{kawaharazuka@jsk.imi.i.u-tokyo.ac.jp} and \url{matsushima@weblab.t.u-tokyo.ac.jp} \\ This is a preprint of an article whose final and definitive form has been published in ADVANCED ROBOTICS 2024, copyright Taylor \& Francis and Robotics Society of Japan, is available online at: \url{http://www.tandfonline.com/Article} DOI; \url{https://doi.org/10.1080/01691864.2024.2408593}.}
    \\\vspace{6pt}
    \textsuperscript{*} First two authors equally contributed to this work\\
    \textsuperscript{a} Department of Mechano-Informatics, The University of Tokyo, Japan\\
    \textsuperscript{b} School of Engineering, The University of Tokyo, Japan \\
    \textsuperscript{c} Meta AI Research \\
    \textsuperscript{d} Google DeepMind 
}

\maketitle

\begin{abstract}
Recent developments in foundation models, like Large Language Models (LLMs) and Vision-Language Models (VLMs), trained on extensive data, facilitate flexible application across different tasks and modalities.
Their impact spans various fields, including healthcare, education, and robotics.
This paper provides an overview of the practical application of foundation models in real-world robotics, with a primary emphasis on the replacement of specific components within existing robot systems.
The summary encompasses the perspective of input-output relationships in foundation models, as well as their role in perception, motion planning, and control within the field of robotics.
This paper concludes with a discussion of future challenges and implications for practical robot applications.
\end{abstract}


\section{Introduction}

Recent advancements in artificial intelligence have markedly expanded the operational capabilities of robots, enabling them to undertake a diverse range of activities  \cite{open_x_embodiment_rt_x_2023,zitkovich2023rt2,nair2022r3m,ahn2022saycan,liang2023codeaspolicies}. Although initially robots' deployment was primarily limited to mass production environments \cite{goel2020robotics,hagele2016industrial,kondo1998robotics,koren1985robotics,de2018industrial,ren2020agricultural}, the applicability of industrial robots has now branched into areas of small-batch and high-variety production, including indoor spaces and disaster sites \cite{gates2007robot,kidd2008robots,murphy2016disaster,burke2004moonlight}. This proliferation is not merely limited to an increase in environmental diversity; it also extends to an expanded repertoire of tasks, encompassing everyday activities like tidying~\cite{abdo2015robot,matsushima2022world,wu2023tidybot}, washing \cite{marchetti2022pet,mir2018portable}, wiping \cite{leidner2019cognition,thosar2018review}, and cooking \cite{bollini2013interpreting,sugiura2010cooking}.

Machine learning has provided a way to meet the needs of such robot systems.
However, training each model only on domain-specific data is insufficient for the diverse robots, tasks, and environments.
There is a growing need for the development of robots that can be applied to various bodies, tasks, and environments using a single, pre-trained system or module.

One solution to this challenge is the introduction of foundation models \cite{bommasani2021opportunities}.
Foundation models are models that are trained on large amounts of data and can be easily applied to a wide range of downstream tasks through in-context learning, fine-tuning, or even in a zero-shot manner \cite{dong2022survey,OpenAIGPT42023}.
Prominent examples include Large Language Models (LLMs) like GPT~\cite{OpenAIGPT42023} and Vision-Language Models (VLMs) like CLIP~\cite{radford2021clip}, where language is a glue to combine various types of modalities.
The impact of these foundation models has been remarkable, with several review papers discussing their influence across diverse fields \cite{wang2024large, zeng2023large, firoozi2023foundation, hu2023generalpurpose}.
\citet{wang2024large} and \citet{zeng2023large} have conducted surveys on the application of large language models in robotics, while \citet{firoozi2023foundation} and \citet{hu2023generalpurpose} have conducted surveys on a broader scale, focusing on the application of foundation models in robotics.
In this paper, we summarize the applicability of foundation models to real-world robots, aiming to accelerate their adoption in practical robot applications.
In comparison to the other survey papers, we provide a summary of how to replace specific components in existing robot systems with foundation models, from the perspective of input-output relationships of the foundation models, as well as perception, motion planning, and control in robotics.

\begin{figure}[tb]
  \centering
  \includegraphics[width=\linewidth]{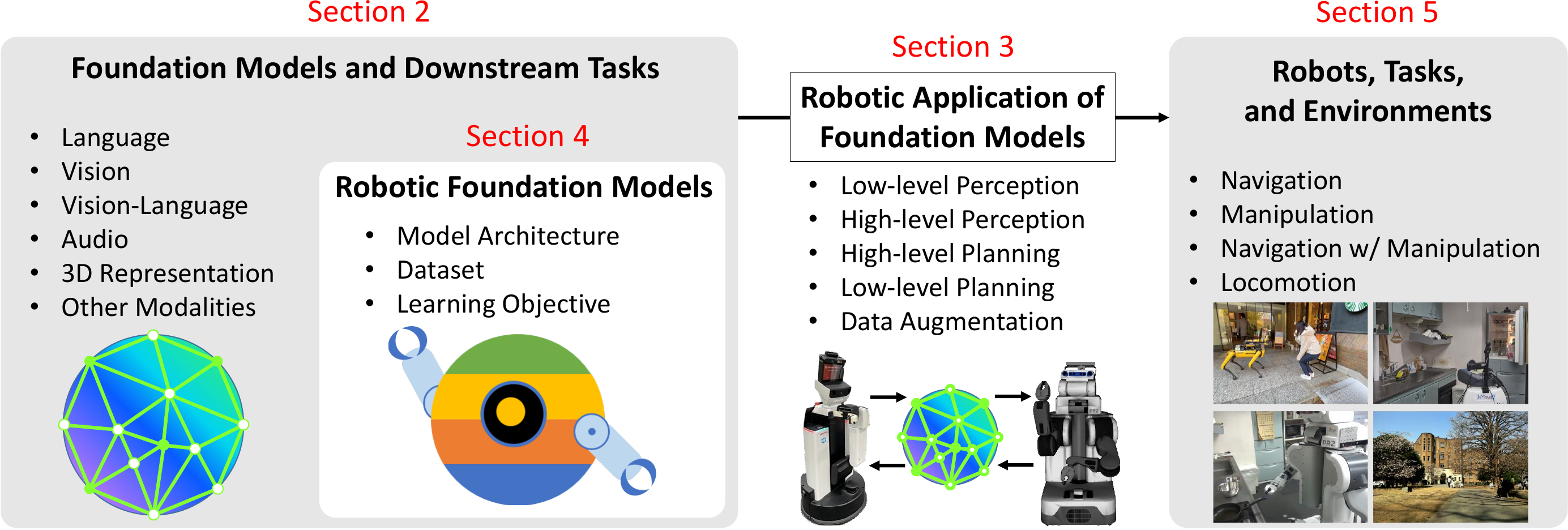}
  \caption{The structure of this study. In~\autoref{sec:foundation-models}, we overview the characteristics of foundation models and introduce common downstream tasks. In~\autoref{sec:fm_application_robotics}, we categorize studies of applications of foundation models in robotics. In~\autoref{sec:fm_for_robotics}, we introduce prior work on creating foundation models for robotics, so-called robotic foundation models. In~\autoref{sec:robot_task_environment}, we overview robots, tasks, and environments used for applications of foundation models in robotics.}
  \label{figure:concept}
\end{figure}

The structure of this study is shown in~\autoref{figure:concept}.
In~\autoref{sec:foundation-models}, we will describe the foundation models themselves.
In particular, we will classify foundation models in terms of what kind of modalities they use, e.g., vision~\cite{yang2022diffusion,yuan2021florence}, language~\cite{brown2020language,chowdhery2023PaLM,zhang2022opt,zeng2022glm,scao2022bloom,touvron2023llama,anil2023palm} and so on, and what kind of downstream tasks to which they can be applied.
In~\autoref{sec:fm_application_robotics}, we will describe how to apply foundation models to robotics based on current applications~\cite{nair2022r3m,Thibault2021ToolLanguage,zitkovich2023rt2}.
In general, robots need to be equipped with a perception module, a planning module, and a control module.
From this perspective, we classify the ways in which one can apply foundation models to real-world robotics into low-level perception, high-level perception, high-level planning, and low-level planning.
Additionally, we will explain data augmentation for robotics when training a mapping to directly connect low-level perception and low-level planning.
In~\autoref{sec:fm_for_robotics}, we will describe foundation models that include robot embodiment, the robotic foundation model, including discussions about how to make these robotic foundation models in terms of model architecture, dataset, and learning objective.
In~\autoref{sec:robot_task_environment}, we will describe the robots, tasks, and environments where foundation models are used.
We classify the tasks into navigation, manipulation, navigation with manipulation, locomotion, and communication.
Finally, we will discuss future challenges and present our conclusions.

\section{Foundation Models}
\label{sec:foundation-models}
The term \emph{foundation model} was first introduced in~\cite{bommasani2021opportunities}.
In this survey, we will simply describe the types of foundation models used in robotic applications, as well as downstream tasks, deferring to~\cite{bommasani2021opportunities} for a discussion of foundation models themselves.

In 2012, deep learning gained mainstream attention from the machine learning community with the winning model from the ILSVRC-2012 competition~\cite{krizhevsky2012imagenet}.
In 2017, the Transformer model was introduced by~\cite{vaswani2017attention}, leading to significant advancements in the fields of natural language processing (NLP)~\citep{Devlin2019} and computer vision~\citep{dosovitskiy2021an}.
In 2021, a model that has been trained on a large amount of data and can be easily applied to a wide range of downstream tasks has come to be referred to as a ``foundation model''~\cite{bommasani2021opportunities}.
Foundation models are characterized by three main characteristics:
\begin{itemize}
    \item In-context learning
    \item Scaling law
    \item Homogenization
\end{itemize}
\emph{In-context learning} enables the accomplishment of new tasks with just a few examples, without the need for retraining or fine-tuning.
\emph{Scaling laws} allow for continued performance improvements as data, computational resources, and model sizes are increased.
\emph{Homogenization} allows for certain foundation model architectures to handle diverse modalities in a unified manner.

\begin{figure}[tb]
  \centering
  \includegraphics[width=\linewidth]{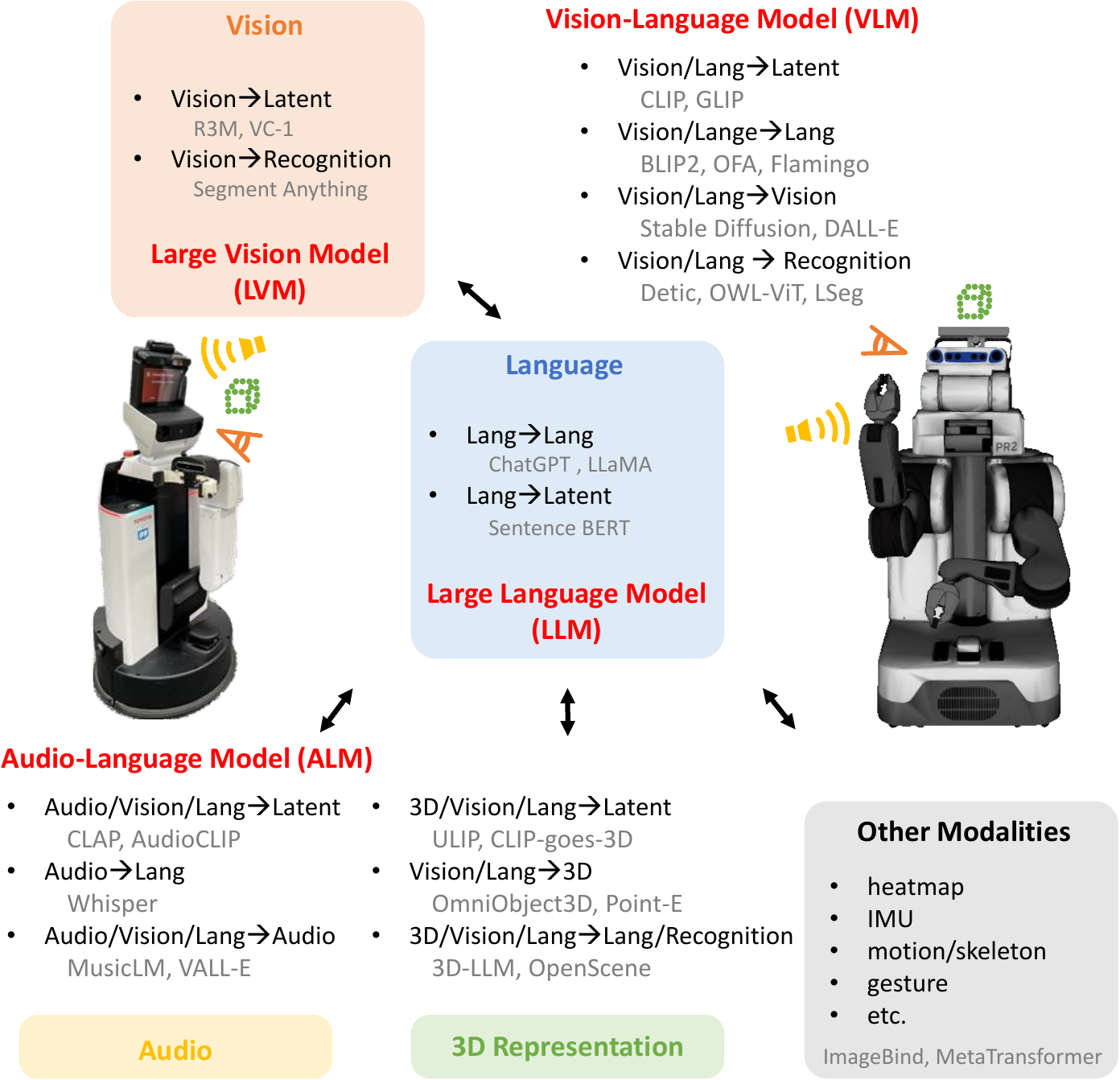}
  \caption{The overview of foundation models classified by the modalities such as language, vision, audio, and 3D representation, and by the network input and output.}
  \label{figure:foundation-models}
\end{figure}

In this chapter, we classify foundation models from the perspective of their applicability in robotics.
The most critical criterion for robots to leverage foundation models is the choice of which modalities to use.
This chapter discusses the types of foundation models and the downstream tasks that they can perform from the perspectives of language, vision, audio, 3D representation, and various other modalities.
In the context of utilizing each modality, we further classify foundation models from the perspective of network inputs and outputs.
The overview is shown in \autoref{figure:foundation-models}.
Note that we do not aim to comprehensively cover foundation models here; our focus remains on addressing differences in modalities and classification of foundation models.

\subsection{Foundation Models for Language} 
\textit{Large Language Models} (LLMs) often refer to large deep neural networks trained on vast amounts of data. 
From the viewpoint of large deep neural networks, Transformers are common architectures for recent LLMs, which may refer to decoder-only autoregressive architectures that consist of multiple Transformer blocks trained to predict the next token.
In terms of the data, the datasets for training LLMs are Internet-scale; for example, the pre-training of GPT-3~\citep{GwernGPT3} uses CommonCrawl covering Internet data from years 2016 to 2019, which amounts to 45TB of compressed plain text and 410 billion tokens after filtering.
Codex~\citep{Codex2021} leverages substantial code data from the snapshot of 54 million public software repositories hosted on GitHub in May 2020 to enable code-writing capabilities with LLMs.
LLMs are the basis of recent foundation models, which can generalize to diverse domains and combine multiple types of data.
The network input and output of LLMs primarily fall into two categories:
\begin{itemize}
    \item \texttt{Language $\to$ Language}
    \item \texttt{Language $\to$ Latent}
\end{itemize}

Regarding \texttt{Language $\to$ Language}, various LLMs have been developed, including GPT-3~\citep{brown2020language}, LLaMA~\citep{touvron202302llama,touvron2023llama}, and others.
These models can be prompted in order to induce \emph{Chain of Thought (CoT)}~\citep{kojima2022large,wei2022chain} which allows them to achieve more complex inference capabilities by going through intermediate reasoning steps while solving a problem.
Furthermore, this \texttt{language} is not limited to natural language but can also handle various formats such as program code, markup languages (e.g., HTML, XML), and data structures (e.g., JSON, YAML)~\citep{Codex2021}.
In-context learning can also lead to a significant improvement in its performance through the use of examples presented in a few-shot manner.

As for \texttt{Language $\to$ Latent}, models like \fixed{BERT~\citep{Devlin2019}, RoBERTa~\citep{Liu2019a} }have been developed that can transform language into latent space vectors.
Also, one can obtain latent space vectors from large language models which can perform \texttt{Language $\to$ Language} conversions, by extracting features from intermediate layers.
By projecting into the latent space, it becomes possible to measure the distance between two sentences, enabling one to extract similar sentences for tasks such as retrieval.

\fixed{Additionally, it is important to highlight the significance of using language models in robotics.
There are several advantages to employing language models in robots, with a key benefit being that language models encompass human common sense.
This enables robots to autonomously perform tasks such as motion planning and recognition. 
Moreover, knowledge can be imparted to the robots through in-context learning.
However, a major challenge is hallucination.
Since language models do not always produce correct outputs, various methods are being developed to ensure reliability.}

\subsection{Foundation Models for Vision} 
The foundation models for images, Large Vision Models (LVM), can mainly be divided into two types based on network input and output:
\begin{itemize}
    \item \texttt{Vision $\to$ Latent}
    \item \texttt{Vision $\to$ Recognition}
\end{itemize}
Regarding \texttt{Vision $\to$ Latent}, various techniques have been developed to project images from diverse situations into some latent space. Representative examples include R3M~\citep{nair2022r3m} and VC-1~\citep{majumdar2023where}.
These methods are designed to transform image information from various data sources into lower-dimensional vectors, which can be considered as feature extraction of high-level representations.

As for \texttt{Vision $\to$ Recognition}, in this context, \texttt{Recognition} refers to performing tasks such as semantic segmentation, instance segmentation, and bounding box extraction on images.
Typically, recognition is based on language labels, and there are not many foundation models that perform these recognitions solely based on image information.
Among them, Segment Anything~\citep{kirillov2023segany} is a model specialized in image segmentation, capable of segmenting not only the entire image but also specific areas specified by points or boxes.
More advanced applications like Tracking Anything~\cite{yang2023track} and Faster Segment Anything~\cite{zhang2023faster} have also been developed.
\fixed{It is important to note that \texttt{Vision $\to$ Latent} learning can be self-supervised, making it easy to construct large datasets.
However, \texttt{Vision $\to$ Recognition} learning requires manually collecting large amounts of data, making it less scalable.}

Of course, some models perform \texttt{Vision $\to$ Vision} or \texttt{Latent $\to$ Vision} tasks.
To effectively utilize these models as foundation models for various tasks, however, conditioning on language is often essential.
Therefore, in the following section on Vision and Language, we will introduce examples of such models.

\subsection{Foundation Models for Vision and Language}
While individual language or image processing has limited capabilities, combining vision and language allows us to build diverse foundation models.
These Vision-Language Models (VLMs) are trained with the Internet-scale massive datasets as well as LLMs.
For instance, CLIP~\citep{radford2021clip} is trained on 400 million image-text pair datasets from the Internet.
When classifying VLMs from the network input and output, they mainly fall into the following four categories.
\begin{itemize}
    \item \texttt{Vision + Language $\to$ Latent}
    \item \texttt{Vision + Language $\to$ Language}
    \item \texttt{Vision + Language $\to$ Vision}
    \item \texttt{Vision + Language $\to$ Recognition}
\end{itemize}

\texttt{Vision + Language $\to$ Latent} involves converting images and text into latent space vectors, which allows one to calculate the similarity between them for downstream tasks such as retrieval.
Prominent models include CLIP~\citep{radford2021clip} and GLIP~\citep{li2022grounded}. While CLIP~\citep{radford2021clip} computes correspondences between a single image-text pair, GLIP~\citep{li2022grounded} can calculate correspondences between multiple regions of an image with their text descriptions.

\texttt{Vision + Language $\to$ Language} enables tasks like image captioning (IC), visual question answering (VQA), and visual entailment (VE).
Prominent models include BLIP2~\citep{li2023blip}, Flamingo~\cite{alayrac2022flamingo}, OFA~\citep{wang2022ofa}, and Unified-IO~\citep{lu2023unifiedio}.
BLIP2 is dedicated to \fixed{IC and} VQA, whereas OFA and Unified-IO serve as a foundation model capable of VQA, IC, Visual Grounding (VG), Text-to-Image Generation (TIG), and more within a single network.
Flamingo supports in-context learning for both vision and language.
\fixed{GPT-4 Vision (GPT-4V)~\citep{OpenAIGPT42023} can analyze images that users input into the model and generate the answer as texts, which can apply to IC, VWA, and VE. Extending GPT-4V, GPT-4o (omni) is a multimodal model that can reason across audio, vision, and text.}

\texttt{Vision + Language $\to$ Vision} facilitates tasks like image editing and image generation through language.
Prominent models include Stable Diffusion~\cite{StableDiffusion}, DALL-E~\cite{ramesh2021zero}, and the previously mentioned OFA~\citep{wang2022ofa} and Unified-IO~\citep{lu2023unifiedio}.
They incorporate Transformer-based architectures and diffusion architectures into the models.

\texttt{Vision + Language $\to$ Recognition} enables tasks like semantic segmentation, instance segmentation, and bounding box extraction from images and language.
Prominent models include Detic~\citep{zhou2022detecting}, OWL-ViT~\citep{minderer2022simple}, \fixed{UniVL~\citep{liu2021unified},} LSeg~\citep{li2022languagedriven}, and DinoV2~\citep{oquab2023dinov2}, which \fixed{are applicable to open-vocabulary recognition}.

The \texttt{Vision} category encompasses not only images but also videos.
Some approaches involve inputting videos frame by frame into these models and integrating the results, while others like XCLIP~\cite{ma2022xclip} and StableVideo~\citep{chai2023stablevideo} directly accept videos as input or produce video outputs.

\subsection{Foundation Models for Audio}
By incorporating auditory input into an existing Vision-Language Model, it becomes possible to tackle a wider range of tasks.
These Audio-Language Models (ALMs) can be categorized as follows based on network input and output.
\begin{itemize}
    \item \texttt{Audio + Vision + Language $\to$ Latent}
    \item \texttt{Audio + Language $\to$ Language}
    \item \texttt{Audio + Vision + Language $\to$ Audio}
\end{itemize}

\texttt{Audio + Vision + Language $\to$ Latent} does not only represent the correspondence between images and language like in CLIP and GLIP, but also encodes sound into latent space vectors, enabling similarity calculations.
Prominent models developed for this purpose include CLAP~\citep{elizalde2023clap} and AudioCLIP~\citep{guzhov2022audioclip}.

\texttt{Audio + Language $\to$ Language} primarily enables Speech-to-Text recognition.
Representative models in this category include Whisper~\citep{radford2023robust}.
With Whisper~\citep{radford2023robust}, it is possible to specify easily confused words and perform in-context learning.

\texttt{Audio + Vision + Language $\to$ Audio} mainly facilitates Text-to-Speech and audio conversion.
Prominent models in this domain include MusicLM~\cite{agostinelli2023musiclm} and VALL-E~\cite{wang2023neural}. \fixed{VAST~\citep{chen2024vast} is a multi-modal foundation model, which is applicable to video, audio, subtitle, and texts.}

\subsection{Foundation Models for 3D Representation}
Currently, research incorporating 3D representations such as point clouds and 3D meshes as modalities is on the rise.
3D representations are essential for robot operations.
Classifying these from network input and output can be summarized as follows:
\begin{itemize}
    \item \texttt{3D Representation + Vision + Language $\to$ Latent}
    \item \texttt{Vision + Language $\to$ 3D Representation}
    \item \texttt{3D Representation + Vision + Language $\to$ Recognition}
\end{itemize}

\texttt{3D Representation + Vision + Language $\to$ Latent} enables the transformation of 3D representations into latent space vectors, and therefore allows for similarity calculations between latent vectors, like in previous models like CLIP and CLAP.
Representative models in this category include ULIP~\citep{xue2023ulip} and CLIP-goes-3D~\citep{hegde2023clip}.
It allows for tasks such as classifying objects of interest based on their 3D shapes, images, or language and using them as features for policy learning.

\texttt{Vision + Language $\to$ 3D Representation} outputs the 3D shapes of objects represented as Point Clouds or 3D Meshes based on images or language input.
Prominent models in this category include OmniObject3D~\citep{wu2023omniobject3d} and Point-E~\citep{nichol2022point}.
These can be useful for tasks like comparing the shape of manipulated objects to reality or viewing them from different angles.
Also, TexFusion can output 3D texture with text-guided image diffusion models \cite{cao2023texfusion}.

\texttt{3D Representation + Vision + Language $\to$ Recognition} enables tasks such as segmentation and bounding box extraction in 3D space through language, similar to previous approaches.
Representative algorithms include 3D-LLM \cite{hong20233d},  OpenScene~\citep{peng2023openscene}, and SpatialVLM~\cite{chen2024spatialvlm}, which combine 3D representations with LLMs.

\subsection{Foundation Models for Other Modalities}
Up until now, we have discussed language, vision, audio, and 3D representations, but in reality it is possible to use various types of sensors. 
Such sensors include IMUs, heatmaps, object poses, and skeletal movements including gestures.
Notable models in this regard include ImageBind~\cite{girdhar2023imagebind} and Meta-Transformer~\cite{zhang2023meta}.
These methods, like CLIP, CLAP, and ULIP, enable similarity calculations, but they simultaneously handle a significantly larger number of modalities.
In addition, FoundationPose~\cite{wen2023foundationpose} is a unified foundation model for 6D object pose estimation and tracking, supporting both model-based and model-free setups.
There are numerous techniques for dealing with gestures and skeletal movements, such as Human Motion Diffusion Model~\cite{tevet2023human}, T2M-GPT~\cite{zhang2023generating}, and GestureDiffCLIP~\cite{Ao2023GestureDiffuCLIP}, which generate human motion or gesture through spoken language.

Finally, it is important to note that there are many models with slightly different input and output configurations beyond what we have discussed here.
By determining which modalities are being utilized and what tasks are possible, we can anticipate future application of these models in real-world robotics.

\section{Applications of Foundation Models to Robotics}
\label{sec:fm_application_robotics}
\begin{figure}[tb]
  \centering
  \includegraphics[width=\linewidth]{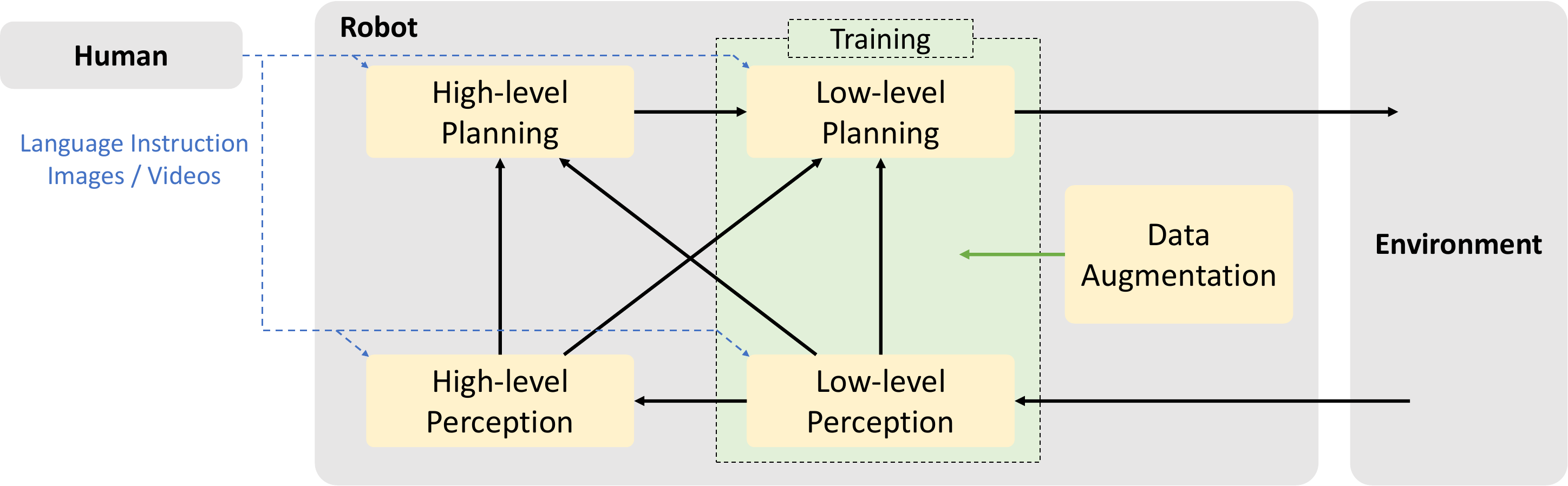}
  \caption{The overview of utilization of foundation models for robots. With foundation models, low-level perception conducts feature extraction or scene recognition, high-level perception conducts reward generation or map construction, high-level planning conducts task planning or code generation, low-level planning conducts footstep generation or command generation, and data augmentation conducts image augmentation or instruction augmentation.}
  \label{figure:applications}
\end{figure}

In general, the behavior of robots is composed of perception, planning, and control.
In this study, we divide perception into two categories: low-level and high-level.
Also, we refer to planning and control as high-level planning and low-level planning, respectively.
With the addition of data augmentation for learning these components, we categorize the utilization of foundation models for robots into the following five categories.
\begin{itemize}
    \item \emph{low-level perception}
    \item \emph{high-level perception}
    \item \emph{high-level planning}
    \item \emph{low-level planning}
    \item \emph{data augmentation}
\end{itemize}
\fixed{It should be noted that this order of five categories takes into account the typical sequence in robotics: perception, planning, and control.}
The relationship between these categories is shown in \autoref{figure:applications}.
Foundation models for low-level perception include semantic segmentation and bounding box extraction in images or 3D representations, and feature extraction in various modalities.
Foundation models for high-level perception involve the transformation and utilization of results obtained from low-level perception into forms such as maps, rewards, and motion constraints.
Foundation models for high-level planning execute higher-level abstract task planning, excluding direct control.
Foundation models for low-level planning execute lower-level motion control, including joint and end effector control.
Foundation models for data augmentation enhance robustness through data augmentation when executing learning that connects low-level perception and low-level planning.

\begin{figure}[tb]
  \centering
  \includegraphics[width=\linewidth]{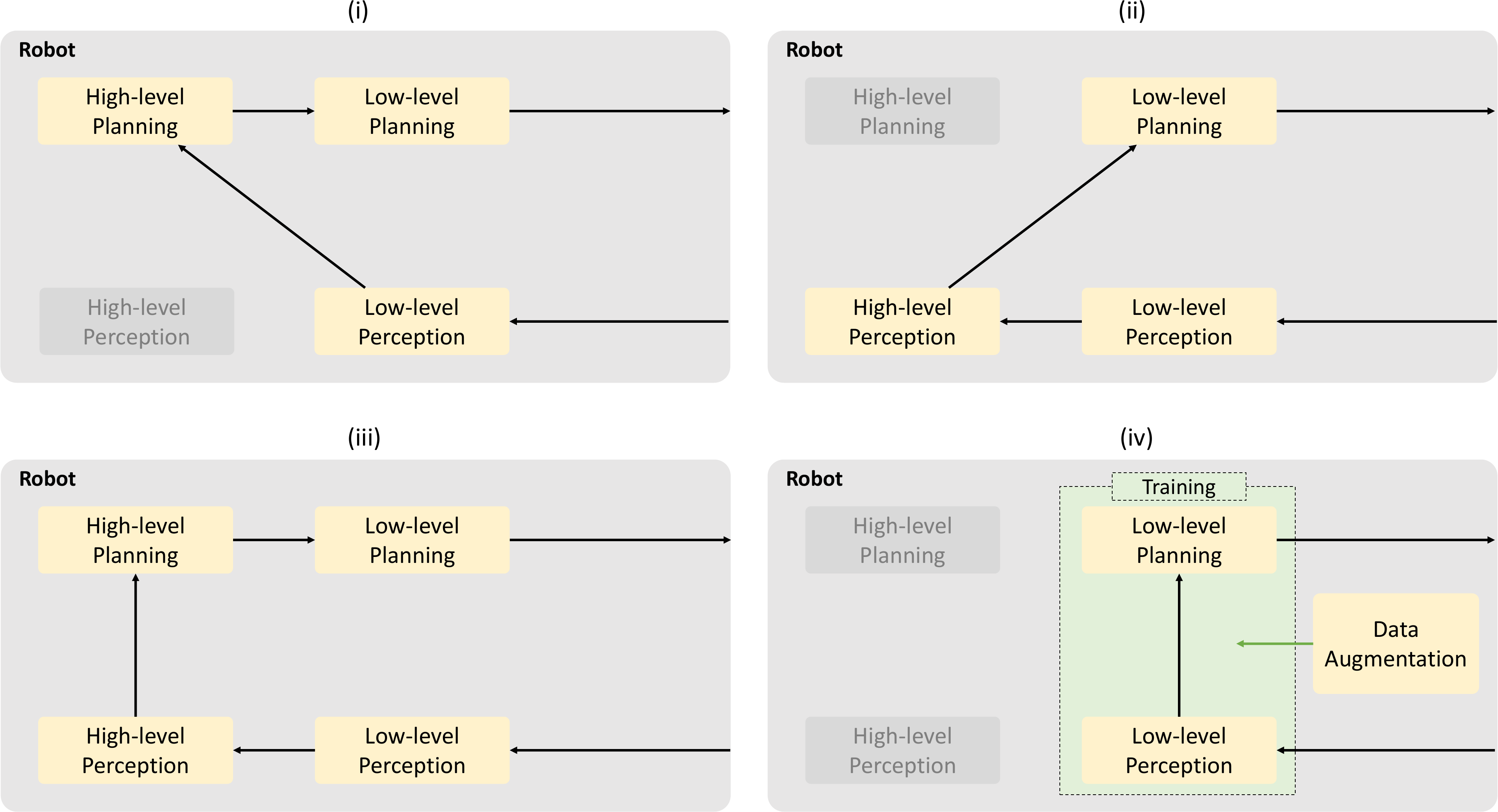}
  \caption{The four types of combinations of low-level perception, high-level perception, high-level planning, low-level planning, and data augmentation with foundation models.}
  \label{figure:applications2}
\end{figure}

In practice, various applications are created by combining these five utilization methods.
They are mainly categorized into four types as shown in \autoref{figure:applications2}. 
\begin{enumerate}
    \renewcommand{\labelenumi}{(\roman{enumi})}
    \item Performing low-level perception and then planning behaviors with high-level planning.
    \item Extracting rewards and motion constraints through low-level perception and high-level perception, and using them for reinforcement learning and trajectory optimization.
    \item Generating maps, scene graphs, and more through low-level perception and high-level perception, and using them as a basis for task planning.
    \item Robustly conducting end-to-end learning that directly correlates feature extraction from low-level perception and control inputs, using data augmentation.
\end{enumerate}
It is important to note that there are also research approaches that do not fit within this framework.

From these perspectives, we select several representative papers and summarize them in \autoref{table:applications}.
The details are discussed subsequently.

\textbf{Low-level Perception}: Representative studies utilizing foundation models for low-level perception include CLIPort~\cite{shridhar2021cliport} and REFLECT~\cite{liu2023reflect}.
CLIPort utilizes the semantic information extracted by CLIP's image and text encoder~\citep{radford2021clip} along with spatial information extracted by Transporter Networks~\citep{zeng2020transporter} to output pick and place positions, serving as an example of using foundation models for feature extraction in low-level perception. 
REFLECT combines object detection by MDETR~\cite{kamath2021mdetr}, object state recognition by CLIP, scene graph description using depth images and heuristics, sound recognition by AudioCLIP~\cite{guzhov2022audioclip}, and a summary of the cause of robot failures along with corrective task plans generated by an LLM, serving as an example of using foundation models for scene recognition including explicit object recognition and sound recognition.
CLIPort falls under category (iv), and REFLECT falls under category (i) in~\autoref{figure:applications2}.

\textbf{High-level Perception}: Representative studies utilizing foundation models for high-level perception include VoxPoser~\cite{huang2023voxposer} and CLIP-Fields~\cite{shafiullah2022clipfields}.
VoxPoser uses an LLM and a VLM to generate affordance maps and constraint maps as programs, and based on these maps, it performs model predictive control, serving as an example of using foundation models for objective function generation and constraint design.
A similar case is using LLM or VLM for reward design in reinforcement learning, such as with Eureka~\cite{ma2023eureka}.
CLIP-Fields utilizes Detic~\citep{zhou2022detecting}, CLIP~\citep{radford2021clip}, and Sentence-BERT~\citep{reimers2019sentence} to embed object labels and object image features for each point cloud in space, enabling navigation, serving as an example of using foundation models for map construction.
VoxPoser falls under category (ii), and CLIP-Fields falls under category (iii) in~\autoref{figure:applications2}.

\textbf{High-level Planning}: Representative studies utilizing foundation models for high-level planning include TidyBot~\cite{wu2023tidybot} and Code as Policies~\cite{liang2023codeaspolicies}.
TidyBot performs object recognition using ViLD~\citep{gu2022open} and object classification using CLIP~\citep{radford2021clip}.
It then infers where to place those objects, considering learned human preferences in conjunction with an LLM, serving as an example of using foundation models for task planning.
Code as Policies generates Python code for pre-defined robot control recognition APIs and control APIs via an LLM, serving as an example of using foundation models for code generation.
Both of these cases fall under category (i) in~\autoref{figure:applications2}.

\textbf{Low-level Planning}: Representative studies utilizing foundation models for low-level planning include SayTap~\cite{tang2023saytap} and General Pattern Machines~\cite{mirchandani2023generalpatternmachines}.
SayTap uses an LLM to output footstep plans and velocity using in-context learning, and then performs reinforcement learning through simulation to obtain joint commands that fulfill these plans, serving as an example of using foundation models for footstep generation.
General Pattern Machines utilizes an LLM as a sequence generator for sequence transformation, completion, and improvement, and is used for reinforcement learning in tasks like Cartpole and action completion in manipulation, serving as an example of using foundation models for direct command generation.
Both of these cases fall under category (i) in~\autoref{figure:applications2}, which consists solely of a low-level planner that does not involve perception.

\textbf{Data Augmentation}: Representative studies utilizing foundation models for data augmentation include GenAug~\cite{chen2023genaug} and DIAL~\cite{xiao2023robotic}.
GenAug extensively augments images using diffusion models~\citep{StableDiffusion} by changing backgrounds, adding distractors, altering object textures, and repositioning different objects to improve the robustness of imitation learning, serving as an example of using foundation models for image data augmentation.
DIAL obtains a large number of language instruction examples through LLM and crowdsourcing, fine-tunes CLIP~\citep{radford2021clip} on a large dataset of trajectory data, and extracts multiple appropriate language instructions for imitation learning, serving as an example of using foundation models for instruction augmentation.
Both of these cases fall under category (iv) in~\autoref{figure:applications2}.

On the other hand, there has been research aiming at creating foundation models for robots that encompass these frameworks, including perception and planning, so-called \emph{Robotic Foundation Models}.
We will provide more details on these in \autoref{sec:fm_for_robotics}.

\begin{table*}[tb]
  \centering
  \caption{Representative studies utilizing foundation models for low-level perception, high-level perception, high-level planning, low-level planning, and data augmentation.}
  \vspace{1.0ex}
  \scalebox{0.55}{
  \begin{tabular}{l c c c c c}
    \toprule
    Papers & Low-level perception & High-level perception & High-level planning & Low-level planning & Data augmentation \\
    \midrule
    CLIPort~\cite{shridhar2021cliport} & \textbf{feature extraction} & - & - & - & - \\
    REFLECT~\cite{liu2023reflect} & \textbf{scene recognition} & - & task planning & - & - \\
    \midrule
    VoxPoser~\cite{huang2023voxposer} & scene recognition & \textbf{objective design} & task planning & - & -\\
    CLIP-Fields~\cite{shafiullah2022clipfields} & feature extraction & \textbf{map construction} & - & - & -\\
    \midrule
    Tidy-Bot~\cite{wu2023tidybot} & scene recognition & - & \textbf{task planning} & - & -\\
    Code as Policies~\cite{liang2023codeaspolicies} & scene recognition & - & \textbf{code generation} & - & -\\
    \midrule
    SayTap~\cite{tang2023saytap} & - & - & - & \textbf{footstep generation} & -\\
    General Pattern~\cite{mirchandani2023generalpatternmachines} & - & - & - & \textbf{command generation} & -\\
    \midrule
    GenAug~\cite{chen2023genaug} & - & - & - & - & \textbf{image augmentation}\\
    DIAL~\cite{xiao2023robotic} & - & - & - & - & \textbf{instruction augmentation}\\
    \bottomrule
  \end{tabular}
  }
  \label{table:applications}
\end{table*}

\subsection{Low-level Perception} \label{subsec:low-level-perception}
We will describe examples of performing low-level perception with foundation models.
As shown in \autoref{table:applications}, its use can mainly be divided into feature extraction and scene recognition.

\subsubsection{Low-level Perception for Feature Extraction}
Visual pretrained models have often been utilized as strong baselines of feature extraction for perception e.g., ResNet~\cite{he2016residual} and MoCo~\cite{he2020momentum}. 
More recently, foundation models, which are trained with more diverse and sometimes multi-modal datasets, are used as pre-trained and general feature extractors, e.g. CLIP~\cite{radford2021clip} and R3M~\cite{nair2022r3m}.
They are leveraged as either off-the-shelf models or fine-tuned with a dataset of downstream tasks.

RRL~\cite{shah2021rrl} uses a pre-trained ResNet trained with ImageNet~\citep{JiaDeng2009} (1,000 classes, 1.28M images) to extract features for a reinforcement learning policy with image input. %
R3M~\cite{nair2022r3m} is a visual pre-trained model for robotics trained with Ego4D datasets~\citep{grauman2022ego4d}, which consists of egocentric human activity video, with the intention of reducing the domain gap between the source datasets and robot manipulation tasks. %
R3M is now one of the most commonly used visual pre-trained models; for instance, CACTI~\cite{mandi2022cacti} uses R3M as a feature extractor for converting camera input to features for imitation learning combined with a ResNet trained with MoCo. %
\fixed{Also, CLIP~\cite{radford2021clip} is used in CLIPort~\cite{shridhar2021cliport} to extract language and image embeddings.}

\subsubsection{Low-level Perception for Scene Recoginition}
Foundation models can be used off-the-shelf for common robot vision tasks of robot systems, such as object detection and semantic segmentation.
Due to being trained with diverse text-image pair datasets, many of these models can be used in open-vocabulary settings without further fine-tuning.
For example, Detic~\cite{zhou2022detecting}, SAM~\cite{kirillov2023segany}, OWL-ViT~\cite{minderer2022simple}, and ViLD~\cite{gu2022open} are open-vocabulary object detection and segmentation models. %
ViLD is often combined with CLIP for open-vocabulary object recognition from robot camera observation in table-top manipulation tasks~\cite{zeng2023socraticmodels} and mobile manipulation tasks~\cite{ahn2022saycan,wu2023tidybot,chen2023nlmapsaycan}.
OFA~\citep{wang2022ofa} is a sequence-to-sequence pre-trained vision language model that allows image captioning and visual question answering (VQA), text-to-image generation.

\fixed{REFLECT~\citep{liu2023reflect} combines object detection with MDETR~\citep{kamath2021mdetr}, object state recognition by CLIP~\citep{radford2021clip}, and sound recognition by AudioCLIP~\cite{guzhov2022audioclip} for a summary of the cause of robot failures.}
GIRAF~\citep{lin2023gestureinformed} utilizes SAM to perform object segmentation on the scene images, and OpenCLIP~\citep{openclip2021}, an open-sourced version of CLIP, for assigning object labels to the segmented images, in combination with supervised gesture detection models.
In~\citet{kawaharazuka2023vqa}, OFA~\citep{wang2022ofa} is leveraged to classify the scene as a binary state, e.g. whether a door is open or not, for robot navigation.
Similarly, in~\citet{kanazawa2023recognition}, CLIP is utilized to detect state changes in food ingredients, such as boiling water and melted butter, in the application of cooking robots.

\subsection{High-level Perception} %
We will describe examples of performing high-level perception with foundation models.
As shown in \autoref{table:applications}, its use can mainly be divided into objective design and map construction.

\subsubsection{High-level Perception for Objective Design}
Foundation models can be utilized for generating goals for downstream planning, generating rewards and costs of the plan, and affordances of controls.

One of the examples of goal generation for planning with foundation models is DALL-E bot~\cite{kapelyukh2023dallebot}, which utilizes a text-conditioned diffusion model for generating goal images for table-top object rearrangement from language instructions.
Also, StructDiffusion~\cite{structdiffusion2023} can offer greater flexibility as it can directly output object placement poses from point cloud observations.
Similarly, SuSIE~\citep{black2023zeroshot} generates subgoals for object manipulation tasks leveraging pre-trained text-conditional image-editing models such as InstructPix2Pix~\citep{brooks2023instructpix2pix}.
UniPi~\cite{du2023learning} generates video using a text-guided video diffusion model with an initial image observation and instruction text, which then can be used in conjunction with a learned inverse model in order to infer actions.

Similarly, foundation models are used to evaluate states of rewards or costs that can be leveraged with classical planners such as Model Predictive Control (MPC) or learned planners such as reinforcement learning policies.
Some algorithms leverage LLMs to generate reward functions from language instructions.
In~\citet{kwon2023rewarddesign}, LLM is leveraged as a proxy of reward functions by providing a few examples or a description of desired behavior. %
Language to Rewards~\cite{yu2023language} utilizes LLMs to describe desired robot motion as text from language instruction and then converts it to code reward functions. The reward function is used to optimize robot motion. %
Dynalang~\cite{lin2023learning} uses an LLM to predict future rewards and utilizes it for model-based reinforcement learning (world models) based on DreamerV3~\cite{hafner2023dreamerv3}. %
ZeST~\citep{cui2022can} utilizes VLMs to specify goals for robot manipulation tasks in a zero-shot manner. %

Foundation models are also leveraged to learn affordances for manipulation tasks.
For example, some works learn affordances for object manipulation from human egocentric video (e.g., EPIC-KITCHENS~\cite{damen2022rescaling} and Ego4D~\cite{grauman2022ego4d} dataset).
RoboAffordances~\cite{bahl2023affordances} learns a contact heatmap and trajectories from human egocentric videos (EPIC-KITCHENS).
Similarly, VIP~\cite{ma2023vip} learns reward functions for object manipulation from the Ego-4D dataset by time contrastive learning and transfers it to real-world robot manipulation settings.
LIV~\cite{ma2023liv} extends VIP by learning a multi-modal (language and image) reward function from text-annotated video (EPIC-KITCHENS), which is used to train language-conditional imitation learning policies.
\fixed{Aside from learning 2D visual affordances, VoxPoser~\cite{huang2023voxposer} represents visual affordances as a 3D Voxel map for motion planning with MPC.}

\subsubsection{High-level Perception for Map Construction}
Vision language models, such as CLIP, are used for extracting semantic information in the mapping, localization, and navigation tasks of mobile robots.
LM-Nav~\cite{shah2022lmnav} uses CLIP as a vision-and-language model for grounding the robot's visual observations to landmarks in order to create graphs of visual landmarks\fixed{.} %
CLIP on Wheels~\cite{gadre2022clip} adapts open-vocabulary models to language-driven zero-shot object navigation (L-ZSON), the task of navigating an environment and finding specific objects without any prior training.
CLIPFields~\cite{shafiullah2022clipfields} uses CLIP for semantic mapping, embedding the visual features from CLIP into a 3D map using neural rendering ~\cite{mildenhall2020nerf,mueller2022instant}.
The model can retrieve the most similar position between the text input and the semantic map. %
Similarly, NLMap-Saycan~\cite{chen2023nlmapsaycan} proposes an open-vocabulary queryable semantic representation using CLIP and ViLD~\cite{gu2022open} to form a semantic map that can be used for navigation with text instructions.
AVL-Maps~\cite{huang23avlmaps} and VL-Maps~\cite{huang23vlmaps} generate a top-down projection map with the embeddings of text, image, and audio information, which a robot can use to navigate to appropriate places.
Furthermore, ConceptFusion~\cite{jatavallabhula2023conceptfusion} is engaged in constructing a 3D map that allows queries through images, sound, and text, while OpenFusion~\cite{yamazaki2024openfusion} is developing a method for performing this process in real-time.
Also, ConceptGraphs~\cite{gu2023conceptgraphs} proposes an open-vocabulary graph-structured representation for 3D scenes, enabling the robot to answer complex scene queries, open-vocabulary pick and place, and traversability estimation.

\subsection{High-level Planning}
We will describe examples of performing high-level planning with foundation models.
Generating high-level motion planning is one of the most effective use cases of LLMs.
As shown in \autoref{table:applications}, its use can mainly be divided into task planning and code generation.

\subsubsection{High-level Planning with Task Planning}
In \citet{huang2022language}, next motion plans are generated as language using GPT-3 (causal LLM), the actionable plan is chosen from action set using BERT (Masked LLM), and then the action is executed.
SayCan~\cite{ahn2022saycan} is one of the most well-known papers connecting LLM and robotics. An LLM outputs the scores (log-likelihoods) of the next action candidates from language instruction, which are multiplied by scores from visual affordance, and the action with the highest combined score is chosen as the action to be executed.
LM-Nav~\cite{shah2022lmnav} outputs landmarks using LLM, matches the landmarks and nodes of vision-navigation map, and executes graph search to navigate to the place specified by natural language.
Inner Monologue~\cite{huang2022inner} updates a task plan with an LLM using feedback from the current state description. This combines a passive scene description, an active scene description, and success detector.
Hulc++~\cite{mees2023hulc2} combines a model-based policy guiding the robot to its vicinity with an affordance heatmap using VLM and language instruction, and language-conditioned imitation learning. For long-term task execution, the language instruction is broken down into multiple subtasks with an LLM.
PIGINet~\cite{yang2023piginet} generates multiple plans using an LLM, scores the feasibility of each plan, and executes the best one.
Grounded Decoding~\cite{huang2023grounded} is the evolutional version of SayCan, which does not generate the entire sentence but generates tokens successively.
LLM-GROP~\cite{ding2023llmgrop} outputs how certain objects are arranged using an LLM. The LLM outputs the symbolic spatial relationship of objects, and then converts it to the geometric spatial relationship whose feasibility levels are evaluated. Finally, the feasibility and efficiency of motion plans are optimized.
AVL-Maps~\cite{huang23avlmaps} and VL-Maps~\cite{huang23vlmaps} execute motion planning of navigation with an LLM. By combining the subgoals and a map generated with the embeddings of language, image, and audio information, a robot can navigate to appropriate places.
Text2Motion~\cite{lin2023text2motion} outputs the probability of each skill, multiplies it by geometric score, and obtains the best plan by sampling multiple plans iteratively. 
PET (Plan, Eliminate, and Track)~\cite{wu2023plan} outputs subtasks with an LLM, eliminates unrelated objects from recognition, and tracks the state change in the objects for detecting sub-task completion.
In REFLECT~\cite{liu2023reflect}, a robot summarizes why the action failed and executes replanning. From MDETR~\cite{kamath2021mdetr}, CLIP, AudioCLIP, 3D scene graph with depth camera, the current state is described, and the failure can be detected.
KITE~\cite{sundaresan2023kite} calculates keypoints in images from language instructions, output actions from an LLM and primitives, and combines them for task execution.
In~\citet{ren2023knowno}, LLM is leveraged to execute a task plan and ask for help using conformal prediction when multiple choices of action exist.
In SayPlan~\cite{rana2023sayplan} a robot navigates by using an LLM and a 3D scene graph. The robot finds objects using a hierarchical 3D scene graph, replanning using a graph simulator and language feedback from failure.
In \citet{ichikura2023diary}, a diary is generated using an LLM and VLM for interaction between a human and a robot. It collects image captioning results, voice, and emotion, and summarizes them by using LLM.
OK-robot~\cite{liu2024okrobot} integrates a variety of learned models trained on publicly available data, such as CLIP, Lang-SAM, AnyGrasp~\cite{fang2023anygrasp}, and OWL-ViT, to pick and drop objects in real-world environments.

\subsubsection{High-level Planner with Code Generation}
Socratic models~\cite{zeng2023socraticmodels} combine VLM, LLM, ALM, and external resources like web search for various applications.
For robot experiments, objects are detected by a VLM, code is generated by an LLM, and motion is generated by CLIPort.
ProgPrompt~\cite{singh2023progprompt} outputs Python code with assertions for checking conditions from task examples and current environment information.
In~\citet{wake2023chatgpt} and~\citet{shirasaka2023self}, an LLM is leveraged to output JSON files of structured action plans.
LLM+P~\cite{liu2023llmplusp} outputs optimal task plans by combining LLMs and PDDL~\cite{aeronautiques1998pddl}.
The task completion rate can be improved by not directly generating language but by generating PDDL codes.
In Voyager~\cite{wang2023voyager}, to acquire new skills for MineDojo, a curriculum is output by an LLM. Feedback from the environment (such as success and failure) is added as prompt, and skill libraries are extended continuously.
Source code for new skills are summarized and embedded into a latent space, which allows them to be retrieved later.
ChatGPT for Robotics~\cite{vemprala2023chatgpt} predefines some APIs and generates source code for some motion controls automatically for Habitat~\citep{puig2023habitat}, AirSim, Tello, and MyCobot.
The platform for sharing prompt engineering results, PromptCraft, is open.
Statler~\cite{yoneda2023statler} explores using LLMs to memorize the world. By deciding the next action using an LLM (world-model writer), and by describing the current state after the action (world-model reader), better performance than code-as-policies can be obtained.
In \citet{obinata2023gpsr}, LLM outputs a sequence of subtasks and hierarchical finite state machines are generated to safely operate a robot over long horizons.
ViLaIn~\cite{shirai2023vision} generates a problem description of PDDL using LLM and VLM.

\subsection{Low-level Planning}
There are a few attempts to construct low-level planners by using foundation models.
They mainly have two types of output: one is for issuing higher-level commands like footstep plans and end effector positions, and the other is for issuing the lowest-level commands, which are the joint angle commands.

SayTap~\cite{tang2023saytap} generates footstep plans and target velocity for a quadruped robot using a large language model and in-context learning.
It acquires policies for converting the obtained footstep plans and velocities into joint angle commands through reinforcement learning, enabling diverse walking patterns.
General Pattern Machines~\cite{mirchandani2023generalpatternmachines} uses large language models to perform sequence transformation, completion, and improvement.
It can be used for tasks such as reinforcement learning tasks like Cartpole and completing manipulation actions from human demonstrations.
Prompt2Walk~\cite{wang2023prompt} constructs a planner utilizing LLMs to generate a gait for a quadruped robot, focusing on outputting direct joint angle commands rather than footstep plans.

\subsection{Data Augmentation}
Data augmentation is indispensable to train agents which can behave in a variety of scenarios.
In particular, the augmentation of images is one of the most important processes for a generalist agent.
Generally, the data augmentation of images is executed by rotating, shifting, zooming, and flipping the image, adding noise to the image, and rescaling the pixels.
However, if we scale the image of a robot that acts based on the view, the relationship between visual information and behavior will be changed.
Therefore, without scaling the image, various changes in the robot view must be prepared.
Also, there can be many different language instructions for a robot to accomplish the same task, meaning that the language instructions should be also augmented.
To solve these problems, current foundation models can be used efficiently.
We introduce several studies that train a generalist agent by augmenting the camera view and language instructions.

CACTI~\cite{mandi2022cacti} has improved the accuracy of imitation learning with R3M or pre-trained ResNet for various tasks and scenes, through image augmentation by stable diffusion~\cite{StableDiffusion}.
ROSIE~\cite{yu2023scaling} has enhanced the robustness of RT-1~\cite{brohan2023rt} by selecting regions with OWL-ViT~\cite{minderer2022simple} and then modifying and augmenting the image using Image Editor~\cite{wang2023imagen} and language prompts.
GenAug~\cite{chen2023genaug} augments images to improve the robustness of imitation learning by using a pre-trained depth-guided text-to-image inpainting model~\cite{StableDiffusion} to change the background, add distractors, modify the object's texture, and place different objects for image augmentation.
FoundationPose~\cite{wen2023foundationpose} adapts LLM-aided texture augmentation with diffusion models for a more realistic appearance and strong generalizability.
Moo~\cite{stone2023moo} separates the action and object information of language instruction, masks images with object information using OWL-ViT, and generalizes to various objects by removing object information from language instruction, leaving only action information.
DIAL~\cite{xiao2023robotic} is an example of instruction augmentation, which obtains many examples of language instructions through LLM or crowdsourcing, extracts appropriate language instructions using fine-tuned CLIP for trajectory data, and uses them for imitation learning.

\section{Progress towards Robotic Foundation Models}
\label{sec:fm_for_robotics}

Aside from studies focusing on leveraging foundation models for robot perception and planning, there are some works aiming to create foundation models for robotics itself, which may be referred to as \emph{robotic foundation models}.
Developing robotic foundation models requires special considerations compared to the robotics domain compared to LLM or VLM.
For example, collecting diverse datasets for robotics is more costly than for LLMs and VLMs, which can be easily trained using data collected from the internet, and this difficulty inhibits the scaling up of robotic foundation models.
Also, robot systems should handle diversified types of data other than just images or text.
\fixed{In general, the design choices are based on reducing human labor to collect those datasets.}

\subsection{Approaches for Robotic Foundation Models}
\label{subsec:robotic-foundation-models}

\begin{table}[tbp]
    \centering
    \caption{Some examples of representative foundation models trained for robotics. The models can be categorized into three: pre-trained visual representations (PVRs) for robotics, vision language models (VLMs) for robotics, and end-to-end control policies and dynamics models.
    For the columns of inputs and outputs, Im, S, L, R, and A denote images, robot states, language tokens, rewards, and actions.
    For the column of architecture, ViT denotes Vision Transformer for vision, and T denotes Transformers. Especially for CNNs, we denote ResNet\citep{he2016residual} as $\text{CNN}_\text{R}$.
    For the training objectives, CE denotes categorical cross-entropy of discrete tokens, and BC denotes behavioral cloning loss on action space.}
    \vspace{1.0ex}
    \scalebox{0.65}{
    \begin{tabular}{lcccccc}
        \toprule
        Model
            & Inputs
            & Outputs
            & Architecture
            & Objective
            \\
        \midrule
        R3M~\citep{nair2022r3m}
            & Im
            & Latent
            & $\text{CNN}_\text{R}$
            & Time Contrastive~\citep{sermanet2018time}
            \\
        MVP~\citep{xiao2022masked,radosavovic2022realworld}
            & Im
            & Im,Latent
            & ViT
            & MAE~\citep{he2022masked}
            \\
        \midrule
        PaLM-E~\citep{driess2023palm}
            & Im,L
            & L
            & ViT+T
            & CE
            \\
        RoboVQA~\citep{sermanet2023robovqa}
            & Im,L
            & L
            & ViT+T
            & CE
            \\
        \midrule
        MT-Opt~\citep{kalashnikov2021scaling}
            & Im,S,A
            & Q-value
            & CNN
            & Q-learning
            \\
        BC-Z~\citep{jang2022bc}
            & Im,L
            & A
            & $\text{CNN}_\text{R}$
            & Regession~(BC)
            \\
        Gato~\citep{reed2022generalist}
            & Im,S,L,A
            & L,A
            & $\text{CNN}_\text{R}$+T
            & CE
            \\
        RT-1~\citep{brohan2023rt}
            & Im,L
            & A
            & CNN+T
            & CE~(BC)
            \\
        RT-2~\citep{zitkovich2023rt2}
            & Im,L
            & A
            & ViT+T
            & CE~(BC)
            \\
        Q-Transformer~\citep{chebotar2023qtransformer}
            & Im,L,A
            & Q-value
            & CNN+T
            & Q-learning
            \\
        RoboCat~\citep{bousmalis2023robocat}
            & Im,S,A
            & Im,A
            & $\text{CNN}_\text{R}$+T
            & CE
            \\
        TDMs~\citep{schubert2023generalist}
            & S,R,A
            & S,R,A
            & T
            & CE
            \\
        \bottomrule
    \end{tabular}
    }
    \label{tab:rfm_model}
\end{table}

We first categorize robotic foundation models based on their applications.
\autoref{tab:rfm_model} shows examples of robotic foundation models.

\subsubsection{Pre-trained Visual Representations for Robotics}
Firstly, pre-trained visual representations (PVRs)~\citep{majumdar2023where} for robotics have been developed and utilized.
Large visual pretraining models for robotics are developed mainly for feature extraction (low-level perception in~\autoref{subsec:low-level-perception}).
For example, RRL~\citep{shah2021rrl} utilizes pretrained CNN (ResNet~\citep{he2016residual}) with ImageNet~\citep{JiaDeng2009} as a feature extractor for vision-based reinforcement learning.
R3M~\citep{nair2022r3m} is a ResNet-based visual pretraining model for robot manipulation trained by ego-centric human activity videos (Ego4D~\citep{grauman2022ego4d}) with time contrastive learning~\citep{sermanet2018time}. %
VIP~\citep{ma2023vip} learns features embedding of the observation utilizing time contrastive learning and defines value functions implicitly as the distance on embedding spaces.
MVP~\citep{xiao2022masked} proposes to learn a visual pre-trained model for imitation learning using Masked Autoencoder~(MAE)~\citep{he2022masked} with Vision Transformer backbones and later applied to real-world settings~\citep{radosavovic2022realworld}. %
VC-1~\citep{majumdar2023where} is a large Vision Transformer-based PVR trained with ImageNet, Ego4D, and robotic datasets of manipulation and navigation.

\subsubsection{Vision Language Models for Robotics}
The second direction of robotic foundation models is to develop (vision) language models for robotics.
PaLM-E~\cite{driess2023palm} is a multimodal language model, which allows multimodal inputs (e.g., images) in addition to text questions, and is trained to answer the questions.
The model is an extension of PaLM~\citep{chowdhery2023PaLM}, a large language model with language inputs.
The text output (answer) can be utilized for multiple downstream tasks, for example, task and motion planning and visual question answering (VQA)\fixed{~\citep{sermanet2023robovqa}}.
RoboVQA~\citep{sermanet2023robovqa} presents a scalable data collection scheme for visual question answering for robotics in addition to a large and diverse cross-embodiment video-and-text dataset.

\subsubsection{End-to-End Control Policies and Dynamics Models}
The third direction is to train control policies with large-scale datasets in an end-to-end manner.
This type of robotic foundation model is sometimes called as~\textbf{Vision-Language-Action} (VLA) model.
For example, MT-Opt~\citep{kalashnikov2021scaling} learns a value function (Q-function) of robot manipulation tasks with a large dataset collected with seven robots and 12 tasks, and utilizes it for fast adaptation of novel tasks, ideally in zero-shot transfer if there are sufficient overlaps among tasks.
BC-Z~\citep{jang2022bc} learns ResNet-based language-conditional policy from large-scale dataset~(more than 25k episodes) of 100 manipulation tasks with 7 DoF manipulators.
The dataset is collected semi-autonomously (shared-autonomy) with HG-Dagger~\citep{kelly2019hg} to reduce data collection costs; they collect about 11K expert demonstrations first and the rest is collected by deploying the learned policy with the aggregated dataset.
VIMA~\citep{jiang2023vima} learns an end-to-end visual manipulation policy regarding images in the same way as text tokens; the visual demonstrations and language instructions are tokenized and input to the model, and the model outputs discretized actions.

\textbf{Robotics Transformers}:
There is a line of works which is called as ``Robotics Transformer''~(RT) that learns robot control policies using transformers.
RT-1~\citep{brohan2023rt} learns language conditional imitation learning policy with large models~(35M parameters).
They collected 130k demonstrations with 13 mobile manipulators (taking 17 months) with about 750 language instructions.
Recently, there have been several works extending RT-1.
For example, Q-Transformer~\citep{chebotar2023qtransformer} extends RT-1 to offline reinforcement learning settings by learning Q-function considering offline temporal difference backup.
RT-2~\cite{zitkovich2023rt2} scales up the model of RT-1 by leveraging Internet-scale vision-language models \cite{driess2023palm,chen2023pali} and co-finetuning the model with the original dataset and datasets collected with robots (seven skills of RT-1 dataset).
The model sizes are with 12B parameters for models based on PaLM-E~\citep{zitkovich2023rt2} and 55B for models based on PaLI-X~\citep{chen2023pali}.

There are several works extending RT-1 and RT-2. For example, RT-Trajectory~\citep{gu2023rt} extend RT-1 architecture to allow human demonstrations of the new target tasks as trajectory sketches, and similarly, RT-Sketch~\citep{anonymous2024rtsketch} is a goal-conditional policy which incorporates human hand-drawing of the trajectory as inputs.
AutoRT~\citep{ahn2024autort} integrates LLMs, VLMs, and end-to-end robot control models like RT-1 and RT-2 to enable robots to gather training datasets in novel environments and scale the model itself.
SARA-RT~\citep{leal2023sara} proposes up-training, a novel fine-tuning methods for robotics, replacing attention blocks with lightweight ones during training, which enables massive models as RT-2 in robotics with less latency.

RT-X~\citep{open_x_embodiment_rt_x_2023} extends the works of RT-1 and RT-2 in terms of the datasets to train with.
They collected datasets of various types of robots, environments, and tasks from more than 20 academic laboratories and trained RT-1 and RT-2 models, which can effectively generalize new robot embodiment and tasks by finetuning.
Based on the models in RT-1 and datasets collected in RT-X, Octo~\citep{octo_2023} proposes the flexible architecture for new modalities of inputs and outputs such as proprioception as input and joint angle as output action.
They incorporate diffusion-based policy models~\citep{chi2023diffusion} as the action head of the model to transform the output token from Transformer models to control actions.

\textbf{Large Models for Learning Dynamics}:~
Large-scale vision-language-action models are used to model not only robot policy but also the dynamics, which are used for planning.
For example, Gato~\citep{reed2022generalist} learns control policies of diverse domains, \emph{e.g.,} Atari \cite{mnih2015atari}, DeepMind Control Suite \cite{tassa2018deepmind}, and real-world manipulation tasks \cite{geng20193d,shridhar2021cliport,gu2017deep,sundaresan2023kite} of a single model.
RoboCat~\citep{bousmalis2023robocat} learns dynamics models of diverse control tasks, such as pick-and-place and stacking blocks.
TDMs~\citep{schubert2023generalist} utilizes Transformer-based sequence models as dynamics models, which can be utilized as a zero-shot generalist and a fine-tuned one for control tasks.
UniSim~\citep{yang2023learning} is a large-scale action-conditional video diffusion models, which can be utilized as simulators of interactions with the real world.

In the following sections, we categorize the recipes for developing robotic foundation models in terms of model architecture (\autoref{subsec:rfm_architecture}), 

\subsection{Model Architectures}
\label{subsec:rfm_architecture}
Traditionally, the choice of deep learning model is dependent on the modalities of the input of the models; for example, CNNs are used for images and RNNs are used for sequential data such as videos and audio.
Recently, as mentioned in~\autoref{sec:foundation-models}, one of the prominent characteristics of modern deep learning network architectures based on Transformers is \textit{homogenization}, which unifies the methodologies of deep learning for various modalities.
Foundation models effectively leverage this feature to enable large-scale multi-modal models that can be shared across various domains.
One of the prominent models for multi-modal input is PerceiverIO~\citep{jaegle2022perceiver}, which generalizes Transformer architecture and scales linearly to the size of inputs and outputs in order to accept various types of inputs such as optical flow and audio as well as texts and images.
Perceiver-Actor~\citep{shridhar2023perceiver} utilizes PerceiverIO as the model of the imitation leaning policy, which allows voxels from RGB-D camera of the environments and language instructions as input and outputs 6-DoF end effector pose, gripper state, and a binary state indicating motion planner for collision avoidance.

The multi-modal inputs are fused into a feature vector to input to the model.
The simplest choice of the fuse is just a concatenation of the features from different modalities.
Recently, FiLM conditioning~\citep{perez2018film} is a popular choice for conditioning for two different types of features.
For example, BC-Z~\citep{jang2022bc,brohan2023rt,bharadhwaj2023roboagent} utilizes FiLM to condition visual input human demonstrations with visual observations.

Models using transformers require tokenizations of images or actions as well as text.
For the image inputs, the inputs are sometimes decomposed into patches and the features of these patches using CNNs~\citep{octo_2023} or vision transformers~\citep{zitkovich2023rt2} are regarded as image tokens.
Instead, RT-1 uses CNN (EfficientNet~\citep{tan2019efficientnet}) without patchfying the input images and uses the flattened feature map as visual tokens.
TokenLearner~\citep{ryoo2021tokenlearner} is often utilized for reducing dimensions to align dimensions of different modalities.
For the case of RT-1~\citep{brohan2023rt}, TokenLearner is used to subsample visual tokens to input transformers with text token inputs.
Action spaces are often uniformly \fixed{discretized} into bins along with the dimension to align with tokens output by transformers~\cite{brohan2023rt,zitkovich2023rt2}, while Octo~\citep{octo_2023} learns to regress the output token to action spaces using diffusion policy~\citep{chi2023diffusion}.

\subsection{Datasets and Their Collection}
\label{subsec:rfm_dataset}
Datasets utilized to train robotic foundation models are divided into two groups; namely \emph{action-free datasets} and \emph{action-conditional datasets}.

With regard to action-conditional datasets, many have been released in the context of action-conditional video prediction tasks and offline reinforcement learning tasks.
For example, RoboNet~\citep{dasari2019robonet} contains videos and robot actions of table-top object manipulation tasks and has been utilized in action-conditional video prediction tasks.
D4RL~\citep{fu2020d4rl} is a dataset containing several robot control tasks collected in simulators and has been leveraged as a benchmark of offline reinforcement learning tasks~\citep{kumar2020conservative,yu2020mopo,matsushima2021deploymentefficient}.

Recently, more diverse and large-scale datasets have been made public and utilized for training end-to-end foundation models.
For example, Bridge Data~\citep{ebert2022bridge,walke2023bridgedata} contains real-world object manipulation data using a 6-DoF WidowX250 robot arm.
Datasets used in RT-1~\citep{brohan2023rt} are publicly released and utilized in subsequent studies~\citep{zitkovich2023rt2,chebotar2023qtransformer}.
The dataset is collected with human experts and a language description of each task is added.
RT-X~\citep{open_x_embodiment_rt_x_2023} utilizes cross-domain datasets, namely Open X-Embodiement~(OXE) datasets, collected with 22 different robots from 21 institutions.
The dataset contains more than 1M episodes of 527 skills (160,266 tasks).

For the ease of collecting expert demonstration datasets with real robots, robot teleoperation systems suitable for collecting datasets have been proposed.
For instance, ALOHA~\citep{zhao2023learning} is a bimanual teleoperation system with low-cost hardware.
Mobile ALOHA~\citep{fu2024mobile} incorporates ALOHA system into a mobile robot to enable teleoperation of bimanual mobile manipulator.
GELLO~\citep{wu2023gello} proposes a framework for teleoperation of robot arms to collect high-quality datasets with low-cost hardware parts.

Action-free datasets are relatively easier to collect and transfer among different types of robot morphologies, but, low-level actions cannot be directly obtained from the learned model.
Since action-conditional datasets are often dependent on the robot, the action is often generalized (e.g. end-effector pose for object manipulation task).
Some studies combine these two groups of datasets for training.
The typical type of action-free dataset for a robotic foundation model is a video dataset.
For example, Ego4D~\citep{grauman2022ego4d} is a large (3,740 hours) egocentric video dataset of human daily behavior collected across 74 locations.

\subsection{Learning Objectives}
\label{subsec:rfm_objective}

In order to train end-to-end robot control models, a simple behavioral cloning (BC) loss on action spaces are often used; for example, BC-Z~\citep{jang2022bc} uses regression loss on action spaces, and RT-1~\citep{brohan2023rt} and RT-2~\citep{zitkovich2023rt2} use categorical cross-entropy loss on discretized action tokens as next token prediction of actions.
MT-Opt~\cite{kalashnikov2021scaling} and Q-Transformer learns value function using TD-Learning as is often with Q-learning in reinforcement learning~\citep{Sutton2018}.
However, some recent works like Gato~\citep{reed2022generalist} learn to predict the entire trajectory not just the next action or Q-value.
There are some works extending the idea of masked prediction models like BERT~\citep{Devlin2019} to robot trajectories such as Unimask~\citep{carroll2022uni} and Masked Trajectory Models~\citep{wu2023masked}.
By masking tokens in \fixed{trajectories} (observations, actions, rewards, returns, etc.) during training, the various learning objectives on robot trajectories such as behavioral cloning, offline reinforcement learning, and forward dynamics, can be written as the same framework.
\fixed{This formulation can be seen as self-supervised learning on robot trajectories.}

\section{Robots, Tasks, and Environments}
\label{sec:robot_task_environment}
\begin{figure}[htb]
  \centering
  \includegraphics[width=0.9\linewidth]{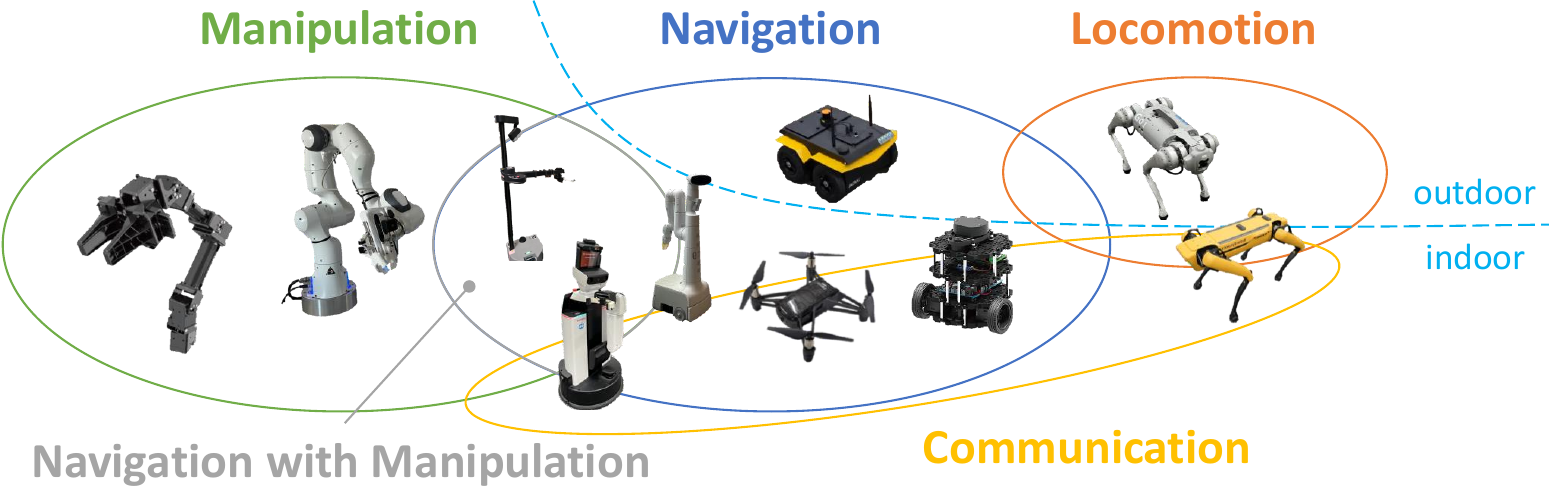}
  \caption{The overview of robots, tasks, and environments used for research with foundation models.}
  \label{figure:robots-tasks}
\end{figure}

From the studies where foundation models are applied, we can know what kind of robots, tasks, and environments are suitable for foundation model-based robotics.
In this study, various studies are classified based on the tasks performed by robots.
They are primarily classified into four categories: navigation, manipulation, locomotion, and communication.
Communication often involves scenarios where robots are not explicitly utilized, and its description is kept to a minimum.
Since navigation and manipulation make up the majority of the tasks, we will discuss five tasks, including those four tasks and navigation with manipulation, which deals with both navigation and manipulation simultaneously, as shown in \autoref{figure:robots-tasks}.
Note that, in this section, we handle only tasks and robots that handle the actual robots, not simulation.
\fixed{Additionally, while there have been many studies on robot behavior using language before the advent of foundation models~\citep{hatori2018interactively,taniguchi2016symbol}, we will focus here solely on the current research landscape utilizing these models.}

\subsection{Navigation}
Research on navigation includes studies that focus on generating maps and those that also involve path planning.
Mainly, research focused on map generation includes VL-Maps~\cite{huang23vlmaps}, AVL-Maps~\cite{huang23avlmaps}, CLIP on Wheels~\cite{gadre2022clip}, ConceptFusion~\cite{jatavallabhula2023conceptfusion}, ConceptGraphs~\cite{gu2023conceptgraphs}, and CLIP-Fields~\cite{shafiullah2022clipfields}.
These studies use embeddings of CLIP~\citep{radford2021clip}, LSeg~\citep{li2022languagedriven}, AudioCLIP~\citep{guzhov2022audioclip}, etc. to generate maps and obtain suitable destinations from language instruction.
On the other hand, LM-Nav~\cite{shah2022lmnav} utilizes an LLM and VLM to perform path planning.
In efforts to create foundation models for navigation, there are also studies like GNM~\cite{shah2022gnm} and ViNT~\cite{shah2023vint}.
Additionally, in ~\cite{vemprala2023chatgpt}, code generation is conducted for the navigation of the drone Tello.
While there is extensive research on map construction indoors, LM-Nav, GNM and ViNT conduct experiments outdoors.
Robots such as Clearpath Ridgeback, Clearpath Jackal, Clearpath Warthog, Stretch, HSR, DJI Tello, TurtleBot, Spot, and Unitree are commonly used in these studies.

\subsection{Manipulation}
Research on manipulation is the most extensive.
Most studies focus on single-arm manipulation and grasping, but there are also studies that deal with dual-arm systems~\cite{ren2023knowno, open_x_embodiment_rt_x_2023} and in-hand manipulation~\cite{xiao2022masked}.
Additionally, the research often takes place in everyday environments, including kitchens and tabletops, although some studies focus on laboratory automation~\cite{yoshikawa2023chemistry}.
The tasks covered are primarily pick-and-place, but many studies also report on obstacle avoidance, path planning, and flexible object manipulation.
Commonly used robotic platforms include Panda, Kinova Series, UR Series, xarm, EDR, mycobot, KUKA, sawyer, and arms using dynamixel servo motors (WidowX, ViperX, etc.).
In terms of end effectors, researchers use not only the robot's built-in hands but also robotiq hands, allegro hands, and custom soft hands.
There are relatively few studies that utilize custom-designed individual arms.

\subsection{Navigation with Manipulation}
Research that combines navigation and manipulation is actively conducted.
All of this research is conducted indoors and deals with both navigation and pick-and-place grasping, either by combining an LLM and VLM \cite{ahn2022saycan, wu2023tidybot, huang2023grounded, ren2023knowno, rana2023sayplan, huang2022inner, chen2023nlmapsaycan, liang2023codeaspolicies, obinata2023gpsr, shirasaka2023self, kawaharazuka2023ptvlm, kawaharazuka2023vqa, liu2024okrobot, yenamandra2023homerobot}, or by simultaneously handling mobile base motion and grasping through the construction of Robotic Foundation Models \cite{brohan2023rt, zitkovich2023rt2, driess2023palm}.
Various robots are used for these purposes, including EDR, single-arm manipulators (e.g. Kinova or Panda) with UGV (e.g. custom mobile platform or Omron mobile platform), Stretch, Fetch, HSR, etc.

\subsection{Locomotion}
There is relatively little research on using foundation models for locomotion.
This is likely to be due to the difficulty of expressing leg movements in language.
Among them, two studies, SayTap~\cite{tang2023saytap} and Language to Rewards~\cite{yu2023language}, use foundation models for locomotion.
SayTap generates footstep plans, while Language to Rewards generates rewards or constraints for reinforcement learning or model predictive control, by utilizing LLM.
Instead of directly handling leg movements, it is common to implicitly adjust leg movements through some abstract layers.
Currently, most of the robots under research in the field of locomotion are quadruped robots, such as Unitree, and research on bipedal robots like humans is lacking.

\subsection{Communication}
There is a considerable amount of research on communication, but among them, studies involving actual robots, not just devices, for communication are relatively scarce.
\fixed{Here, we introduce interactions between embodied robots and humans involving physical contact through LLMs.}
Among these studies, Automatic Diary Generation~\cite{ichikura2023diary, ichikura2023automatic} stands out for extracting appropriate scenes and imbuing emotions from data collected when robots and humans go for a walk, generating diaries to enhance intimacy with humans.
In~\citet{cho2023story}, diaries are utilized where robots convey their experiences and perspectives to humans as an approach to form social recognition for robots.
This diary generation can be considered as a form of high-level perception in our classification.
Also, some studies consider human preferences~\cite{ren2023knowno, wu2023tidybot}.

\section{Conclusion and Future Challenges}
\subsection{Conclusion}
The application of foundation models to robotics is an exciting and fast-changing field.
In this study, we have compiled various studies regarding the real-world robot applications of foundation models.
Especially, we focused on how existing robot systems can be replaced by foundational models.
First, to apply foundation models to real robots, we categorized them based on the perspective of the input and output modalities and their transformations.
Next, we classified these foundation models based on five aspects of their application methodology to real robots: low-level perception, high-level perception, high-level planning, low-level planning, and data augmentation.
Then, we categorized the patterns of the combination of these five aspects.
Following that, we summarized the construction of robotic foundation models, including actions and more, from the perspectives of model architecture, datasets, and learning objectives.
Lastly, we summarized the aspects of robots, tasks, and environments that research on the real-world robot applications of foundation models 
covers.

Given the rapid development of this field, it is challenging to encompass all the research papers, but it allows us to observe the trends.
There are several active directions in current development.
One involves the practical application of existing foundational models to robot movements, with a focus on increasing the use of different modalities.
Additionally, there is a trend towards constructing robotic foundational models using more extensive data and larger models.
The direction also includes building foundational models that generalize across diverse body types and exploring the use of models beyond Transformers, such as Diffusion Models.
Furthermore, research on low-level motion control is still relatively scarce, and there is a limited presence of robots thriving in real household and outdoor environments.
Expectations for future advancements in these areas remain high.
We hope that this study will serve as a guide for future utilization of foundation models.

\subsection{Future Challenges}
First, we discuss the real-world robot application of existing foundation models.
Various foundation models for mutual conversion between different modalities have been developed, and their application to robots is advancing.
However, there are still many modalities that have not been fully utilized.
Particularly, depth information, force feedback, inertial sensors, and motion information of humans, objects, and robots are not fully exploited, despite the existence of diverse examples for language, image, and sound information.
Additionally, there are issues related to the granularity of skills for motion planning.
The hierarchical organization of robot skills, how to structure the skill API at what level of granularity, and the addition of new skills are expected to be significant focus areas in the future.

Second, we elaborate on the robotic foundation models.
While diverse robotic foundation models have been developed, their control frequencies are exceptionally slow.
These current models are not suitable for tasks that require fine-grained force control or collaboration with humans.
\fixed{There is room for studies about which action spaces are suitable for these end-to-end control policies and how to reduce safety risks with robotics foundation models.}

Next, we discuss practical robot applications across the entire spectrum.
Through various studies, it has been observed that there are few research examples in outdoor settings, and many setups lack reflection of real-world environments, often consisting of toy problems.
The use of off-the-shelf robots limits diversity.
Furthermore, most setups are primarily based on position control, with few examples utilizing torque control or soft robots.
It is anticipated that in the future, robots developed by various research institutions will integrate with foundation models, and there will be an increase in examples of operation in more realistic environments.

Next, the generalization ability of robots using foundation models to various environments and tasks is a future challenge.
Currently, examples of applying foundation models to open-world environments, such as OVMM~\cite{yenamandra2023homerobot}, OK-Robot~\cite{liu2024okrobot}, and GOAT~\cite{chang2023goat}, are increasing.
We believe this direction represents a significant advancement towards new robots that go beyond the capabilities of traditional robots.

Lastly, as tasks executed through language instructions become more prevalent, quantitative evaluation of performance becomes challenging.
Benchmarks such as CALVIN~\citep{mees2022calvin} and ARNOLD~\citep{gong2023arnold}, which perform language instruction tasks in simulators for evaluation, have emerged.
Achieving a fair evaluation in the real world, however, is not easy.
Additionally, there is a need to contemplate how to evaluate interactions with humans.

\clearpage

\bibliography{references}

\begin{thebibliography}{225}
\providecommand{\natexlab}[1]{#1}
\providecommand{\url}[1]{\texttt{#1}}
\expandafter\ifx\csname urlstyle\endcsname\relax
  \providecommand{\doi}[1]{doi: #1}\else
  \providecommand{\doi}{doi: \begingroup \urlstyle{rm}\Url}\fi

\bibitem[Collaboration et~al.(2023)Collaboration, Padalkar, Pooley, Jain, Bewley, Herzog, Irpan, Khazatsky, Rai, Singh, Brohan, Raffin, Wahid, Burgess-Limerick, Kim, Sch{\"{o}}lkopf, Ichter, Lu, Xu, Finn, Xu, Chi, Huang, Chan, Pan, Fu, Devin, Driess, Pathak, Shah, B{\"{u}}chler, Kalashnikov, Sadigh, Johns, Ceola, Xia, Stulp, Zhou, Sukhatme, Salhotra, Yan, Schiavi, Su, Fang, Shi, Amor, Christensen, Furuta, Walke, Fang, Mordatch, Radosavovic, Leal, Liang, Kim, Schneider, Hsu, Bohg, Bingham, Wu, Wu, Luo, Gu, Tan, Oh, Malik, Tompson, Yang, Lim, Silv{\'{e}}rio, Han, Rao, Pertsch, Hausman, Go, Gopalakrishnan, Goldberg, Byrne, Oslund, Kawaharazuka, Zhang, Majd, Rana, Srinivasan, Chen, Pinto, Tan, Ott, Lee, Tomizuka, Du, Ahn, Zhang, Ding, Srirama, Sharma, Kim, Kanazawa, Hansen, Heess, Joshi, Suenderhauf, Palo, Shafiullah, Mees, Kroemer, Sanketi, Wohlhart, Xu, Sermanet, Sundaresan, Vuong, Rafailov, Tian, Doshi, Mart{\'{i}}n-Mart{\'{i}}n, Mendonca, Shah, Hoque, Julian, Bustamante, Kirmani, Levine, Moore, Bahl, Dass,
  Song, Xu, Haldar, Adebola, Guist, Nasiriany, Schaal, Welker, Tian, Dasari, Belkhale, Osa, Harada, Matsushima, Xiao, Yu, Ding, Davchev, Zhao, Armstrong, Darrell, Jain, Vanhoucke, Zhan, Zhou, Burgard, Chen, Wang, Zhu, Li, Lu, Chebotar, Zhou, Zhu, Xu, Wang, Bisk, Cho, Lee, Cui, Wu, Tang, Zhu, Li, Iwasawa, Matsuo, Xu, and Cui]{open_x_embodiment_rt_x_2023}
Open X-Embodiment Collaboration, Abhishek Padalkar, Acorn Pooley, Ajinkya Jain, Alex Bewley, Alex Herzog, Alex Irpan, Alexander Khazatsky, Anant Rai, Anikait Singh, Anthony Brohan, Antonin Raffin, Ayzaan Wahid, Ben Burgess-Limerick, Beomjoon Kim, Bernhard Sch{\"{o}}lkopf, Brian Ichter, Cewu Lu, Charles Xu, Chelsea Finn, Chenfeng Xu, Cheng Chi, Chenguang Huang, Christine Chan, Chuer Pan, Chuyuan Fu, Coline Devin, Danny Driess, Deepak Pathak, Dhruv Shah, Dieter B{\"{u}}chler, Dmitry Kalashnikov, Dorsa Sadigh, Edward Johns, Federico Ceola, Fei Xia, Freek Stulp, Gaoyue Zhou, Gaurav~S Sukhatme, Gautam Salhotra, Ge~Yan, Giulio Schiavi, Hao Su, Hao-Shu Fang, Haochen Shi, Heni~Ben Amor, Henrik~I Christensen, Hiroki Furuta, Homer Walke, Hongjie Fang, Igor Mordatch, Ilija Radosavovic, Isabel Leal, Jacky Liang, Jaehyung Kim, Jan Schneider, Jasmine Hsu, Jeannette Bohg, Jeffrey Bingham, Jiajun Wu, Jialin Wu, Jianlan Luo, Jiayuan Gu, Jie Tan, Jihoon Oh, Jitendra Malik, Jonathan Tompson, Jonathan Yang, Joseph~J Lim, João
  Silv{\'{e}}rio, Junhyek Han, Kanishka Rao, Karl Pertsch, Karol Hausman, Keegan Go, Keerthana Gopalakrishnan, Ken Goldberg, Kendra Byrne, Kenneth Oslund, Kento Kawaharazuka, Kevin Zhang, Keyvan Majd, Krishan Rana, Krishnan Srinivasan, Lawrence~Yunliang Chen, Lerrel Pinto, Liam Tan, Lionel Ott, Lisa Lee, Masayoshi Tomizuka, Maximilian Du, Michael Ahn, Mingtong Zhang, Mingyu Ding, Mohan~Kumar Srirama, Mohit Sharma, Moo~Jin Kim, Naoaki Kanazawa, Nicklas Hansen, Nicolas Heess, Nikhil~J Joshi, Niko Suenderhauf, Norman~Di Palo, Nur Muhammad~Mahi Shafiullah, Oier Mees, Oliver Kroemer, Pannag~R Sanketi, Paul Wohlhart, Peng Xu, Pierre Sermanet, Priya Sundaresan, Quan Vuong, Rafael Rafailov, Ran Tian, Ria Doshi, Roberto Mart{\'{i}}n-Mart{\'{i}}n, Russell Mendonca, Rutav Shah, Ryan Hoque, Ryan Julian, Samuel Bustamante, Sean Kirmani, Sergey Levine, Sherry Moore, Shikhar Bahl, Shivin Dass, Shuran Song, Sichun Xu, Siddhant Haldar, Simeon Adebola, Simon Guist, Soroush Nasiriany, Stefan Schaal, Stefan Welker, Stephen Tian,
  Sudeep Dasari, Suneel Belkhale, Takayuki Osa, Tatsuya Harada, Tatsuya Matsushima, Ted Xiao, Tianhe Yu, Tianli Ding, Todor Davchev, Tony~Z Zhao, Travis Armstrong, Trevor Darrell, Vidhi Jain, Vincent Vanhoucke, Wei Zhan, Wenxuan Zhou, Wolfram Burgard, Xi~Chen, Xiaolong Wang, Xinghao Zhu, Xuanlin Li, Yao Lu, Yevgen Chebotar, Yifan Zhou, Yifeng Zhu, Ying Xu, Yixuan Wang, Yonatan Bisk, Yoonyoung Cho, Youngwoon Lee, Yuchen Cui, Yueh-hua Wu, Yujin Tang, Yuke Zhu, Yunzhu Li, Yusuke Iwasawa, Yutaka Matsuo, Zhuo Xu, and Zichen~Jeff Cui.
\newblock {Open X-Embodiment: Robotic Learning Datasets and RT-X Models}.
\newblock \emph{arXiv preprint arXiv:2310.08864}, 2023.

\bibitem[Brohan et~al.(2023{\natexlab{a}})Brohan, Brown, Carbajal, Chebotar, Chen, Choromanski, Ding, Driess, Dubey, Finn, Florence, Fu, Arenas, Gopalakrishnan, Han, Hausman, Herzog, Hsu, Ichter, Irpan, Joshi, Julian, Kalashnikov, Kuang, Leal, Lee, Lee, Levine, Lu, Michalewski, Mordatch, Pertsch, Rao, Reymann, Ryoo, Salazar, Sanketi, Sermanet, Singh, Singh, Soricut, Tran, Vanhoucke, Vuong, Wahid, Welker, Wohlhart, Wu, Xia, Xiao, Xu, Xu, Yu, and Zitkovich]{zitkovich2023rt2}
Anthony Brohan, Noah Brown, Justice Carbajal, Yevgen Chebotar, Xi~Chen, Krzysztof Choromanski, Tianli Ding, Danny Driess, Avinava Dubey, Chelsea Finn, Pete Florence, Chuyuan Fu, Montse~Gonzalez Arenas, Keerthana Gopalakrishnan, Kehang Han, Karol Hausman, Alex Herzog, Jasmine Hsu, Brian Ichter, Alex Irpan, Nikhil Joshi, Ryan Julian, Dmitry Kalashnikov, Yuheng Kuang, Isabel Leal, Lisa Lee, Tsang-Wei~Edward Lee, Sergey Levine, Yao Lu, Henryk Michalewski, Igor Mordatch, Karl Pertsch, Kanishka Rao, Krista Reymann, Michael Ryoo, Grecia Salazar, Pannag Sanketi, Pierre Sermanet, Jaspiar Singh, Anikait Singh, Radu Soricut, Huong Tran, Vincent Vanhoucke, Quan Vuong, Ayzaan Wahid, Stefan Welker, Paul Wohlhart, Jialin Wu, Fei Xia, Ted Xiao, Peng Xu, Sichun Xu, Tianhe Yu, and Brianna Zitkovich.
\newblock {RT-2: Vision-Language-Action Models Transfer Web Knowledge to Robotic Control}.
\newblock In \emph{Proceedings of The 7th Conference on Robot Learning}, pages 2165--2183, 2023{\natexlab{a}}.

\bibitem[Nair et~al.(2022)Nair, Rajeswaran, Kumar, Finn, and Gupta]{nair2022r3m}
Suraj Nair, Aravind Rajeswaran, Vikash Kumar, Chelsea Finn, and Abhinav Gupta.
\newblock {R3M: A Universal Visual Representation for Robot Manipulation}.
\newblock In \emph{6th Annual Conference on Robot Learning}, 2022.
\newblock URL \url{https://openreview.net/forum?id=tGbpgz6yOrI}.

\bibitem[Ahn et~al.(2022)Ahn, Brohan, Brown, Chebotar, Cortes, David, Finn, Fu, Gopalakrishnan, Hausman, Herzog, Ho, Hsu, Ibarz, Ichter, Irpan, Jang, Ruano, Jeffrey, Jesmonth, Joshi, Julian, Kalashnikov, Kuang, Lee, Levine, Lu, Luu, Parada, Pastor, Quiambao, Rao, Rettinghouse, Reyes, Sermanet, Sievers, Tan, Toshev, Vanhoucke, Xia, Xiao, Xu, Xu, Yan, and Zeng]{ahn2022saycan}
Michael Ahn, Anthony Brohan, Noah Brown, Yevgen Chebotar, Omar Cortes, Byron David, Chelsea Finn, Chuyuan Fu, Keerthana Gopalakrishnan, Karol Hausman, Alex Herzog, Daniel Ho, Jasmine Hsu, Julian Ibarz, Brian Ichter, Alex Irpan, Eric Jang, Rosario~Jauregui Ruano, Kyle Jeffrey, Sally Jesmonth, Nikhil~J Joshi, Ryan Julian, Dmitry Kalashnikov, Yuheng Kuang, Kuang-Huei Lee, Sergey Levine, Yao Lu, Linda Luu, Carolina Parada, Peter Pastor, Jornell Quiambao, Kanishka Rao, Jarek Rettinghouse, Diego Reyes, Pierre Sermanet, Nicolas Sievers, Clayton Tan, Alexander Toshev, Vincent Vanhoucke, Fei Xia, Ted Xiao, Peng Xu, Sichun Xu, Mengyuan Yan, and Andy Zeng.
\newblock {Do As I Can, Not As I Say: Grounding Language in Robotic Affordances}.
\newblock In \emph{Proceedings of The 6th Conference on Robot Learning}, pages 287--318. PMLR, 2022.

\bibitem[Liang et~al.(2023)Liang, Huang, Xia, Xu, Hausman, Ichter, Florence, and Zeng]{liang2023codeaspolicies}
Jacky Liang, Wenlong Huang, Fei Xia, Peng Xu, Karol Hausman, Brian Ichter, Pete Florence, and Andy Zeng.
\newblock {Code as Policies: Language Model Programs for Embodied Control}.
\newblock In \emph{IEEE International Conference on Robotics and Automation}, 2023.

\bibitem[Goel and Gupta(2020)]{goel2020robotics}
Ruchi Goel and Pooja Gupta.
\newblock {Robotics and industry 4.0}.
\newblock \emph{A Roadmap to Industry 4.0: Smart Production, Sharp Business and Sustainable Development}, pages 157--169, 2020.

\bibitem[H{\"{a}}gele et~al.(2016)H{\"{a}}gele, Nilsson, Pires, and Bischoff]{hagele2016industrial}
Martin H{\"{a}}gele, Klas Nilsson, J~Norberto Pires, and Rainer Bischoff.
\newblock {Industrial robotics}.
\newblock \emph{Springer handbook of robotics}, pages 1385--1422, 2016.

\bibitem[Kondo and Ting(1998)]{kondo1998robotics}
Naoshi Kondo and K~C Ting.
\newblock {Robotics for plant production}.
\newblock \emph{Artificial intelligence review}, 12:\penalty0 227--243, 1998.

\bibitem[Koren et~al.(1985)Koren, Koren, and {others}]{koren1985robotics}
Yoram Koren, Yoram Koren, and {others}.
\newblock \emph{{Robotics for engineers}}, volume 168.
\newblock McGraw-Hill New York, 1985.

\bibitem[De~Backer et~al.(2018)De~Backer, DeStefano, Menon, and Suh]{de2018industrial}
Koen De~Backer, Timothy DeStefano, Carlo Menon, and Jung~Ran Suh.
\newblock {Industrial robotics and the global organisation of production}.
\newblock \emph{OECD}, 2018.
\newblock URL \url{https://www.oecd-ilibrary.org/content/paper/dd98ff58-en}.

\bibitem[Ren et~al.(2020)Ren, Lin, Ying, Chowdhary, and Ting]{ren2020agricultural}
Guoqiang Ren, Tao Lin, Yibin Ying, Girish Chowdhary, and K~C Ting.
\newblock {Agricultural robotics research applicable to poultry production: A review}.
\newblock \emph{Computers and Electronics in Agriculture}, 169:\penalty0 105216, 2020.

\bibitem[Gates(2007)]{gates2007robot}
Bill Gates.
\newblock {A robot in every home}.
\newblock \emph{Scientific American}, 296\penalty0 (1):\penalty0 58--65, 2007.

\bibitem[Kidd and Breazeal(2008)]{kidd2008robots}
Cory~D Kidd and Cynthia Breazeal.
\newblock {Robots at home: Understanding long-term human-robot interaction}.
\newblock In \emph{2008 IEEE/RSJ International Conference on Intelligent Robots and Systems}, pages 3230--3235, 2008.

\bibitem[Murphy et~al.(2016)Murphy, Tadokoro, and Kleiner]{murphy2016disaster}
Robin~R Murphy, Satoshi Tadokoro, and Alexander Kleiner.
\newblock {Disaster robotics}.
\newblock \emph{Springer handbook of robotics}, pages 1577--1604, 2016.

\bibitem[Burke et~al.(2004)Burke, Murphy, Coovert, and Riddle]{burke2004moonlight}
Jennifer~L Burke, Robin~R Murphy, Michael~D Coovert, and Dawn~L Riddle.
\newblock {Moonlight in Miami: Field study of human-robot interaction in the context of an urban search and rescue disaster response training exercise}.
\newblock \emph{Human-Computer Interaction}, 19\penalty0 (1-2):\penalty0 85--116, 2004.

\bibitem[Abdo et~al.(2015)Abdo, Stachniss, Spinello, and Burgard]{abdo2015robot}
Nichola Abdo, Cyrill Stachniss, Luciano Spinello, and Wolfram Burgard.
\newblock {Robot, organize my shelves! Tidying up objects by predicting user preferences}.
\newblock In \emph{2015 IEEE international conference on robotics and automation (ICRA)}, pages 1557--1564, 2015.

\bibitem[Matsushima et~al.(2022)Matsushima, Noguchi, Arima, Aoki, Okita, Ikeda, Ishimoto, Taniguchi, Yamashita, Seto, and {others}]{matsushima2022world}
Tatsuya Matsushima, Yuki Noguchi, Jumpei Arima, Toshiki Aoki, Yuki Okita, Yuya Ikeda, Koki Ishimoto, Shohei Taniguchi, Yuki Yamashita, Shoichi Seto, and {others}.
\newblock {World robot challenge 2020–partner robot: a data-driven approach for room tidying with mobile manipulator}.
\newblock \emph{Advanced Robotics}, 36\penalty0 (17-18):\penalty0 850--869, 2022.

\bibitem[Wu et~al.(2023{\natexlab{a}})Wu, Antonova, Kan, Lepert, Zeng, Song, Bohg, Rusinkiewicz, and Funkhouser]{wu2023tidybot}
Jimmy Wu, Rika Antonova, Adam Kan, Marion Lepert, Andy Zeng, Shuran Song, Jeannette Bohg, Szymon Rusinkiewicz, and Thomas Funkhouser.
\newblock {TidyBot: Personalized Robot Assistance with Large Language Models}.
\newblock \emph{Autonomous Robots}, 2023{\natexlab{a}}.

\bibitem[Marchetti et~al.(2022)Marchetti, Grimme, Hornecker, Kollakidou, and Graf]{marchetti2022pet}
Emanuela Marchetti, Sophie Grimme, Eva Hornecker, Avgi Kollakidou, and Philipp Graf.
\newblock {Pet-robot or appliance? Care home residents with dementia respond to a zoomorphic floor washing robot}.
\newblock In \emph{Proceedings of the 2022 CHI Conference on Human Factors in Computing Systems}, pages 1--21, 2022.

\bibitem[Mir-Nasiri et~al.(2018)Mir-Nasiri, Siswoyo, and Ali]{mir2018portable}
Nazim Mir-Nasiri, Hudyjaya Siswoyo, and Md~Hazrat Ali.
\newblock {Portable autonomous window cleaning robot}.
\newblock \emph{Procedia computer science}, 133:\penalty0 197--204, 2018.

\bibitem[Leidner et~al.(2019)Leidner, Bartels, Bejjani, Albu-Sch{\"{a}}ffer, and Beetz]{leidner2019cognition}
Daniel Leidner, Georg Bartels, Wissam Bejjani, Alin Albu-Sch{\"{a}}ffer, and Michael Beetz.
\newblock {Cognition-enabled robotic wiping: Representation, planning, execution, and interpretation}.
\newblock \emph{Robotics and Autonomous Systems}, 114:\penalty0 199--216, 2019.

\bibitem[Thosar et~al.(2018)Thosar, Zug, Skaria, and Jain]{thosar2018review}
Madhura Thosar, Sebastian Zug, Alpha~Mary Skaria, and Akshay Jain.
\newblock {A Review of Knowledge Bases for Service Robots in Household Environments.}
\newblock In \emph{AIC}, pages 98--110, 2018.

\bibitem[Bollini et~al.(2013)Bollini, Tellex, Thompson, Roy, and Rus]{bollini2013interpreting}
Mario Bollini, Stefanie Tellex, Tyler Thompson, Nicholas Roy, and Daniela Rus.
\newblock {Interpreting and executing recipes with a cooking robot}.
\newblock In \emph{Experimental Robotics: The 13th International Symposium on Experimental Robotics}, pages 481--495, 2013.

\bibitem[Sugiura et~al.(2010)Sugiura, Sakamoto, Withana, Inami, and Igarashi]{sugiura2010cooking}
Yuta Sugiura, Daisuke Sakamoto, Anusha Withana, Masahiko Inami, and Takeo Igarashi.
\newblock {Cooking with robots: designing a household system working in open environments}.
\newblock In \emph{Proceedings of the SIGCHI conference on human factors in computing systems}, pages 2427--2430, 2010.

\bibitem[Bommasani et~al.(2021)Bommasani, Hudson, Adeli, Altman, Arora, von Arx, Bernstein, Bohg, Bosselut, Brunskill, Brynjolfsson, Buch, Card, Castellon, Chatterji, Chen, Creel, Davis, Demszky, Donahue, Doumbouya, Durmus, Ermon, Etchemendy, Ethayarajh, Fei-Fei, Finn, Gale, Gillespie, Goel, Goodman, Grossman, Guha, Hashimoto, Henderson, Hewitt, Ho, Hong, Hsu, Huang, Icard, Jain, Jurafsky, Kalluri, Karamcheti, Keeling, Khani, Khattab, Koh, Krass, Krishna, Kuditipudi, Kumar, Ladhak, Lee, Lee, Leskovec, Levent, Li, Li, Ma, Malik, Manning, Mirchandani, Mitchell, Munyikwa, Nair, Narayan, Narayanan, Newman, Nie, Niebles, Nilforoshan, Nyarko, Ogut, Orr, Papadimitriou, Park, Piech, Portelance, Potts, Raghunathan, Reich, Ren, Rong, Roohani, Ruiz, Ryan, R{\'{e}}, Sadigh, Sagawa, Santhanam, Shih, Srinivasan, Tamkin, Taori, Thomas, Tram{\`{e}}r, Wang, Wang, Wu, Wu, Wu, Xie, Yasunaga, You, Zaharia, Zhang, Zhang, Zhang, Zhang, Zheng, Zhou, and Liang]{bommasani2021opportunities}
Rishi Bommasani, Drew~A. Hudson, Ehsan Adeli, Russ Altman, Simran Arora, Sydney von Arx, Michael~S. Bernstein, Jeannette Bohg, Antoine Bosselut, Emma Brunskill, Erik Brynjolfsson, Shyamal Buch, Dallas Card, Rodrigo Castellon, Niladri Chatterji, Annie Chen, Kathleen Creel, Jared~Quincy Davis, Dora Demszky, Chris Donahue, Moussa Doumbouya, Esin Durmus, Stefano Ermon, John Etchemendy, Kawin Ethayarajh, Li~Fei-Fei, Chelsea Finn, Trevor Gale, Lauren Gillespie, Karan Goel, Noah Goodman, Shelby Grossman, Neel Guha, Tatsunori Hashimoto, Peter Henderson, John Hewitt, Daniel~E. Ho, Jenny Hong, Kyle Hsu, Jing Huang, Thomas Icard, Saahil Jain, Dan Jurafsky, Pratyusha Kalluri, Siddharth Karamcheti, Geoff Keeling, Fereshte Khani, Omar Khattab, Pang~Wei Koh, Mark Krass, Ranjay Krishna, Rohith Kuditipudi, Ananya Kumar, Faisal Ladhak, Mina Lee, Tony Lee, Jure Leskovec, Isabelle Levent, Xiang~Lisa Li, Xuechen Li, Tengyu Ma, Ali Malik, Christopher~D. Manning, Suvir Mirchandani, Eric Mitchell, Zanele Munyikwa, Suraj Nair,
  Avanika Narayan, Deepak Narayanan, Ben Newman, Allen Nie, Juan~Carlos Niebles, Hamed Nilforoshan, Julian Nyarko, Giray Ogut, Laurel Orr, Isabel Papadimitriou, Joon~Sung Park, Chris Piech, Eva Portelance, Christopher Potts, Aditi Raghunathan, Rob Reich, Hongyu Ren, Frieda Rong, Yusuf Roohani, Camilo Ruiz, Jack Ryan, Christopher R{\'{e}}, Dorsa Sadigh, Shiori Sagawa, Keshav Santhanam, Andy Shih, Krishnan Srinivasan, Alex Tamkin, Rohan Taori, Armin~W. Thomas, Florian Tram{\`{e}}r, Rose~E. Wang, William Wang, Bohan Wu, Jiajun Wu, Yuhuai Wu, Sang~Michael Xie, Michihiro Yasunaga, Jiaxuan You, Matei Zaharia, Michael Zhang, Tianyi Zhang, Xikun Zhang, Yuhui Zhang, Lucia Zheng, Kaitlyn Zhou, and Percy Liang.
\newblock {On the Opportunities and Risks of Foundation Models}.
\newblock \emph{arXiv preprint}, 2021.

\bibitem[Dong et~al.(2022)Dong, Li, Dai, Zheng, Wu, Chang, Sun, Xu, and Sui]{dong2022survey}
Qingxiu Dong, Lei Li, Damai Dai, Ce~Zheng, Zhiyong Wu, Baobao Chang, Xu~Sun, Jingjing Xu, and Zhifang Sui.
\newblock {A survey for in-context learning}.
\newblock \emph{arXiv preprint arXiv:2301.00234}, 2022.

\bibitem[{OpenAI}(2023)]{OpenAIGPT42023}
{OpenAI}.
\newblock {GPT-4 Technical Report}.
\newblock Technical report, 2023.

\bibitem[Radford et~al.(2021)Radford, Kim, Hallacy, Ramesh, Goh, Agarwal, Sastry, Askell, Mishkin, Clark, Krueger, and Sutskever]{radford2021clip}
Alec Radford, Jong~Wook Kim, Chris Hallacy, Aditya Ramesh, Gabriel Goh, Sandhini Agarwal, Girish Sastry, Amanda Askell, Pamela Mishkin, Jack Clark, Gretchen Krueger, and Ilya Sutskever.
\newblock {Learning Transferable Visual Models From Natural Language Supervision}.
\newblock In \emph{International conference on machine learning}, pages 8748--8763. PMLR, 2021.

\bibitem[Wang et~al.(2024)Wang, Wu, Li, Jiang, Shu, Shi, Hu, Ma, Liu, Wang, Yao, Liu, Zhao, Liu, Dai, Zhao, Ge, Li, Liu, and Zhang]{wang2024large}
Jiaqi Wang, Zihao Wu, Yiwei Li, Hanqi Jiang, Peng Shu, Enze Shi, Huawen Hu, Chong Ma, Yiheng Liu, Xuhui Wang, Yincheng Yao, Xuan Liu, Huaqin Zhao, Zhengliang Liu, Haixing Dai, Lin Zhao, Bao Ge, Xiang Li, Tianming Liu, and Shu Zhang.
\newblock {Large Language Models for Robotics: Opportunities, Challenges, and Perspectives}.
\newblock \emph{arXiv preprint}, 2024.

\bibitem[Zeng et~al.(2023{\natexlab{a}})Zeng, Gan, Wang, Liu, and Yu]{zeng2023large}
Fanlong Zeng, Wensheng Gan, Yongheng Wang, Ning Liu, and Philip~S Yu.
\newblock {Large Language Models for Robotics: A Survey}.
\newblock \emph{arXiv preprint}, 2023{\natexlab{a}}.

\bibitem[Firoozi et~al.(2023)Firoozi, Tucker, Tian, Majumdar, Sun, Liu, Zhu, Song, Kapoor, Hausman, Ichter, Driess, Wu, Lu, and Schwager]{firoozi2023foundation}
Roya Firoozi, Johnathan Tucker, Stephen Tian, Anirudha Majumdar, Jiankai Sun, Weiyu Liu, Yuke Zhu, Shuran Song, Ashish Kapoor, Karol Hausman, Brian Ichter, Danny Driess, Jiajun Wu, Cewu Lu, and Mac Schwager.
\newblock {Foundation Models in Robotics: Applications, Challenges, and the Future}.
\newblock \emph{arXiv preprint}, 2023.

\bibitem[Hu et~al.(2023)Hu, Xie, Jain, Francis, Patrikar, Keetha, Kim, Xie, Zhang, Zhao, Chong, Wang, Sycara, Johnson-Roberson, Batra, Wang, Scherer, Kira, Xia, and Bisk]{hu2023generalpurpose}
Yafei Hu, Quanting Xie, Vidhi Jain, Jonathan Francis, Jay Patrikar, Nikhil Keetha, Seungchan Kim, Yaqi Xie, Tianyi Zhang, Shibo Zhao, Yu~Quan Chong, Chen Wang, Katia Sycara, Matthew Johnson-Roberson, Dhruv Batra, Xiaolong Wang, Sebastian Scherer, Zsolt Kira, Fei Xia, and Yonatan Bisk.
\newblock {Toward General-Purpose Robots via Foundation Models: A Survey and Meta-Analysis}.
\newblock \emph{arXiv preprint}, 2023.

\bibitem[Yang et~al.(2022)Yang, Zhang, Song, Hong, Xu, Zhao, Zhang, Cui, and Yang]{yang2022diffusion}
Ling Yang, Zhilong Zhang, Yang Song, Shenda Hong, Runsheng Xu, Yue Zhao, Wentao Zhang, Bin Cui, and Ming-Hsuan Yang.
\newblock {Diffusion models: A comprehensive survey of methods and applications}.
\newblock \emph{ACM Computing Surveys}, 2022.

\bibitem[Yuan et~al.(2021)Yuan, Chen, Chen, Codella, Dai, Gao, Hu, Huang, Li, Li, and {others}]{yuan2021florence}
Lu~Yuan, Dongdong Chen, Yi-Ling Chen, Noel Codella, Xiyang Dai, Jianfeng Gao, Houdong Hu, Xuedong Huang, Boxin Li, Chunyuan Li, and {others}.
\newblock {Florence: A new foundation model for computer vision}.
\newblock \emph{arXiv preprint arXiv:2111.11432}, 2021.

\bibitem[Brown et~al.(2020)Brown, Mann, Ryder, Subbiah, Kaplan, Dhariwal, Neelakantan, Shyam, Sastry, Askell, Agarwal, Herbert-Voss, Krueger, Henighan, Child, Ramesh, Ziegler, Wu, Winter, Hesse, Chen, Sigler, Litwin, Gray, Chess, Clark, Berner, McCandlish, Radford, Sutskever, and Amodei]{brown2020language}
Tom~B. Brown, Benjamin Mann, Nick Ryder, Melanie Subbiah, Jared Kaplan, Prafulla Dhariwal, Arvind Neelakantan, Pranav Shyam, Girish Sastry, Amanda Askell, Sandhini Agarwal, Ariel Herbert-Voss, Gretchen Krueger, Tom Henighan, Rewon Child, Aditya Ramesh, Daniel~M. Ziegler, Jeffrey Wu, Clemens Winter, Christopher Hesse, Mark Chen, Eric Sigler, Mateusz Litwin, Scott Gray, Benjamin Chess, Jack Clark, Christopher Berner, Sam McCandlish, Alec Radford, Ilya Sutskever, and Dario Amodei.
\newblock {Language models are few-shot learners}.
\newblock In \emph{Advances in neural information processing systems 33}, pages 1877--1901, 2020.

\bibitem[Chowdhery et~al.(2023)Chowdhery, Narang, Devlin, Bosma, Mishra, Roberts, Barham, Chung, Sutton, Gehrmann, Schuh, Shi, Tsvyashchenko, Maynez, Rao, Barnes, Tay, Shazeer, Prabhakaran, Reif, Du, Hutchinson, Pope, Bradbury, Austin, Isard, Gur-Ari, Yin, Duke, Levskaya, Ghemawat, Dev, Michalewski, Garcia, Misra, Robinson, Fedus, Zhou, Ippolito, Luan, Lim, Zoph, Spiridonov, Sepassi, Dohan, Agrawal, Omernick, Dai, Pillai, Pellat, Lewkowycz, Moreira, Child, Polozov, Lee, Zhou, Wang, Saeta, Diaz, Firat, Catasta, Wei, Meier-Hellstern, Eck, Dean, Petrov, and Fiedel]{chowdhery2023PaLM}
Aakanksha Chowdhery, Sharan Narang, Jacob Devlin, Maarten Bosma, Gaurav Mishra, Adam Roberts, Paul Barham, Hyung~Won Chung, Charles Sutton, Sebastian Gehrmann, Parker Schuh, Kensen Shi, Sasha Tsvyashchenko, Joshua Maynez, Abhishek Rao, Parker Barnes, Yi~Tay, Noam Shazeer, Vinodkumar Prabhakaran, Emily Reif, Nan Du, Ben Hutchinson, Reiner Pope, James Bradbury, Jacob Austin, Michael Isard, Guy Gur-Ari, Pengcheng Yin, Toju Duke, Anselm Levskaya, Sanjay Ghemawat, Sunipa Dev, Henryk Michalewski, Xavier Garcia, Vedant Misra, Kevin Robinson, Liam Fedus, Denny Zhou, Daphne Ippolito, David Luan, Hyeontaek Lim, Barret Zoph, Alexander Spiridonov, Ryan Sepassi, David Dohan, Shivani Agrawal, Mark Omernick, Andrew~M. Dai, Thanumalayan~Sankaranarayana Pillai, Marie Pellat, Aitor Lewkowycz, Erica Moreira, Rewon Child, Oleksandr Polozov, Katherine Lee, Zongwei Zhou, Xuezhi Wang, Brennan Saeta, Mark Diaz, Orhan Firat, Michele Catasta, Jason Wei, Kathy Meier-Hellstern, Douglas Eck, Jeff Dean, Slav Petrov, and Noah Fiedel.
\newblock {PaLM: Scaling Language Modeling with Pathways}.
\newblock \emph{J. Mach. Learn. Res.}, 24\penalty0 (240):\penalty0 1--113, 2023.

\bibitem[Zhang et~al.(2022)Zhang, Roller, Goyal, Artetxe, Chen, Chen, Dewan, Diab, Li, Lin, and {others}]{zhang2022opt}
Susan Zhang, Stephen Roller, Naman Goyal, Mikel Artetxe, Moya Chen, Shuohui Chen, Christopher Dewan, Mona Diab, Xian Li, Xi~Victoria Lin, and {others}.
\newblock {Opt: Open pre-trained transformer language models}.
\newblock \emph{arXiv preprint arXiv:2205.01068}, 2022.

\bibitem[Zeng et~al.(2022)Zeng, Liu, Du, Wang, Lai, Ding, Yang, Xu, Zheng, Xia, and {others}]{zeng2022glm}
Aohan Zeng, Xiao Liu, Zhengxiao Du, Zihan Wang, Hanyu Lai, Ming Ding, Zhuoyi Yang, Yifan Xu, Wendi Zheng, Xiao Xia, and {others}.
\newblock {Glm-130b: An open bilingual pre-trained model}.
\newblock \emph{arXiv preprint arXiv:2210.02414}, 2022.

\bibitem[Scao et~al.(2022)Scao, Fan, Akiki, Pavlick, Ili{\'{c}}, Hesslow, Castagn{\'{e}}, Luccioni, Yvon, Gall{\'{e}}, and {others}]{scao2022bloom}
Teven~Le Scao, Angela Fan, Christopher Akiki, Ellie Pavlick, Suzana Ili{\'{c}}, Daniel Hesslow, Roman Castagn{\'{e}}, Alexandra~Sasha Luccioni, François Yvon, Matthias Gall{\'{e}}, and {others}.
\newblock {Bloom: A 176b-parameter open-access multilingual language model}.
\newblock \emph{arXiv preprint arXiv:2211.05100}, 2022.

\bibitem[Touvron et~al.(2023{\natexlab{a}})Touvron, Martin, Stone, Albert, Almahairi, Babaei, Bashlykov, Batra, Bhargava, Bhosale, and {others}]{touvron2023llama}
Hugo Touvron, Louis Martin, Kevin Stone, Peter Albert, Amjad Almahairi, Yasmine Babaei, Nikolay Bashlykov, Soumya Batra, Prajjwal Bhargava, Shruti Bhosale, and {others}.
\newblock {Llama 2: Open foundation and fine-tuned chat models}.
\newblock \emph{arXiv preprint arXiv:2307.09288}, 2023{\natexlab{a}}.

\bibitem[Anil et~al.(2023)Anil, Dai, Firat, Johnson, Lepikhin, Passos, Shakeri, Taropa, Bailey, Chen, and {others}]{anil2023palm}
Rohan Anil, Andrew~M Dai, Orhan Firat, Melvin Johnson, Dmitry Lepikhin, Alexandre Passos, Siamak Shakeri, Emanuel Taropa, Paige Bailey, Zhifeng Chen, and {others}.
\newblock {Palm 2 technical report}.
\newblock \emph{arXiv preprint arXiv:2305.10403}, 2023.

\bibitem[Thibault et~al.(2021)Thibault, Py, Gervasi, Salemme, Koun, L{\"{o}}vden, Boulenger, Roy, and Brozzoli]{Thibault2021ToolLanguage}
Simon Thibault, Raphaël Py, Angelo~Mattia Gervasi, Romeo Salemme, Eric Koun, Martin L{\"{o}}vden, Véronique Boulenger, Alice~C. Roy, and Claudio Brozzoli.
\newblock {Tool use and language share syntactic processes and neural patterns in the basal ganglia}.
\newblock \emph{Science}, 374\penalty0 (6569), 2021.
\newblock ISSN 0036-8075.
\newblock \doi{10.1126/science.abe0874}.

\bibitem[Krizhevsky et~al.(2012)Krizhevsky, Sutskever, and Hinton]{krizhevsky2012imagenet}
Alex Krizhevsky, Ilya Sutskever, and Geoffrey~E. Hinton.
\newblock {ImageNet classification with deep convolutional neural networks}.
\newblock In \emph{Advances in neural information processing systems}, volume~60, 2012.

\bibitem[Vaswani et~al.(2017)Vaswani, Shazeer, Parmar, Uszkoreit, Jones, Gomez, Kaiser, and Polosukhin]{vaswani2017attention}
Ashish Vaswani, Noam Shazeer, Niki Parmar, Jakob Uszkoreit, Llion Jones, Aidan~N. Gomez, Lukasz Kaiser, and Illia Polosukhin.
\newblock {Attention Is All You Need}.
\newblock In \emph{Proceedings of the 31st International Conference on Neural Information Processing Systems}, pages 6000--6010, 2017.

\bibitem[Devlin et~al.(2019)Devlin, Chang, Lee, and Toutanova]{Devlin2019}
Jacob Devlin, Ming-Wei Chang, Kenton Lee, and Kristina Toutanova.
\newblock {BERT: Pre-training of Deep Bidirectional Transformers for Language Understanding}.
\newblock In \emph{Proceedings of NAACL-HLT}, pages 4171--4186, 2019.

\bibitem[Dosovitskiy et~al.(2021)Dosovitskiy, Beyer, Kolesnikov, Weissenborn, Zhai, Unterthiner, Dehghani, Minderer, Heigold, Gelly, and {others}]{dosovitskiy2021an}
Alexey Dosovitskiy, Lucas Beyer, Alexander Kolesnikov, Dirk Weissenborn, Xiaohua Zhai, Thomas Unterthiner, Mostafa Dehghani, Matthias Minderer, Georg Heigold, Sylvain Gelly, and {others}.
\newblock {An image is worth 16x16 words: Transformers for image recognition at scale}.
\newblock In \emph{International Conference on Learning Representations}, 2021.
\newblock URL \url{https://openreview.net/forum?id=YicbFdNTTy}.

\bibitem[{Gwern}(2021)]{GwernGPT3}
{Gwern}.
\newblock {GPT-3 Creative Fiction}, 2021.
\newblock URL \url{https://www.gwern.net/GPT-3}.

\bibitem[Chen et~al.(2021)Chen, Tworek, Jun, Yuan, Pinto, Kaplan, Edwards, Burda, Joseph, Brockman, Ray, Puri, Krueger, Petrov, Khlaaf, Sastry, Mishkin, Chan, Gray, Ryder, Pavlov, Power, Kaiser, Bavarian, Winter, Tillet, Such, Cummings, Plappert, Chantzis, Barnes, Herbert-Voss, Guss, Nichol, Paino, Tezak, Tang, Babuschkin, Balaji, Jain, Saunders, Hesse, Carr, Leike, Achiam, Misra, Morikawa, Radford, Knight, Brundage, Murati, Mayer, Welinder, McGrew, Amodei, McCandlish, Sutskever, and Zaremba]{Codex2021}
Mark Chen, Jerry Tworek, Heewoo Jun, Qiming Yuan, Henrique Ponde de~Oliveira Pinto, Jared Kaplan, Harri Edwards, Yuri Burda, Nicholas Joseph, Greg Brockman, Alex Ray, Raul Puri, Gretchen Krueger, Michael Petrov, Heidy Khlaaf, Girish Sastry, Pamela Mishkin, Brooke Chan, Scott Gray, Nick Ryder, Mikhail Pavlov, Alethea Power, Lukasz Kaiser, Mohammad Bavarian, Clemens Winter, Philippe Tillet, Felipe~Petroski Such, Dave Cummings, Matthias Plappert, Fotios Chantzis, Elizabeth Barnes, Ariel Herbert-Voss, William~Hebgen Guss, Alex Nichol, Alex Paino, Nikolas Tezak, Jie Tang, Igor Babuschkin, Suchir Balaji, Shantanu Jain, William Saunders, Christopher Hesse, Andrew~N. Carr, Jan Leike, Josh Achiam, Vedant Misra, Evan Morikawa, Alec Radford, Matthew Knight, Miles Brundage, Mira Murati, Katie Mayer, Peter Welinder, Bob McGrew, Dario Amodei, Sam McCandlish, Ilya Sutskever, and Wojciech Zaremba.
\newblock {Evaluating Large Language Models Trained on Code}.
\newblock \emph{arXiv preprint}, 2021.

\bibitem[Touvron et~al.(2023{\natexlab{b}})Touvron, Lavril, Izacard, Martinet, Lachaux, Lacroix, Rozi{\`{e}}re, Goyal, Hambro, Azhar, and {others}]{touvron202302llama}
Hugo Touvron, Thibaut Lavril, Gautier Izacard, Xavier Martinet, Marie-Anne Lachaux, Timothée Lacroix, Baptiste Rozi{\`{e}}re, Naman Goyal, Eric Hambro, Faisal Azhar, and {others}.
\newblock {Llama: Open and efficient foundation language models}.
\newblock \emph{arXiv preprint arXiv:2302.13971}, 2023{\natexlab{b}}.

\bibitem[Kojima et~al.(2022)Kojima, Gu, Reid, Matsuo, and Iwasawa]{kojima2022large}
Takeshi Kojima, Shixiang~Shane Gu, Machel Reid, Yutaka Matsuo, and Yusuke Iwasawa.
\newblock {Large language models are zero-shot reasoners}.
\newblock \emph{Advances in neural information processing systems}, 35:\penalty0 22199--22213, 2022.

\bibitem[Wei et~al.(2022)Wei, Wang, Schuurmans, Bosma, Xia, Chi, Le, Zhou, and {others}]{wei2022chain}
Jason Wei, Xuezhi Wang, Dale Schuurmans, Maarten Bosma, Fei Xia, Ed~Chi, Quoc~V Le, Denny Zhou, and {others}.
\newblock {Chain-of-thought prompting elicits reasoning in large language models}.
\newblock \emph{Advances in Neural Information Processing Systems}, 35:\penalty0 24824--24837, 2022.

\bibitem[Liu et~al.(2019)Liu, Ott, Goyal, Du, Joshi, Chen, Levy, Lewis, Zettlemoyer, and Stoyanov]{Liu2019a}
Yinhan Liu, Myle Ott, Naman Goyal, Jingfei Du, Mandar Joshi, Danqi Chen, Omer Levy, Mike Lewis, Luke Zettlemoyer, and Veselin Stoyanov.
\newblock {RoBERTa: A robustly optimized BERT pretraining approach}.
\newblock \emph{arXiv preprint}, 2019.
\newblock ISSN 23318422.

\bibitem[Majumdar et~al.(2023)Majumdar, Yadav, Arnaud, Ma, Chen, Silwal, Jain, Berges, Abbeel, Malik, and {others}]{majumdar2023where}
Arjun Majumdar, Karmesh Yadav, Sergio Arnaud, Yecheng~Jason Ma, Claire Chen, Sneha Silwal, Aryan Jain, Vincent-Pierre Berges, Pieter Abbeel, Jitendra Malik, and {others}.
\newblock {Where are we in the search for an Artificial Visual Cortex for Embodied Intelligence?}
\newblock In \emph{Workshop on Reincarnating Reinforcement Learning at ICLR 2023}, 2023.
\newblock URL \url{https://openreview.net/forum?id=NJtSbIWmt2T}.

\bibitem[Kirillov et~al.(2023)Kirillov, Mintun, Ravi, Mao, Rolland, Gustafson, Xiao, Whitehead, Berg, Lo, Doll{\'{a}}r, and Girshick]{kirillov2023segany}
Alexander Kirillov, Eric Mintun, Nikhila Ravi, Hanzi Mao, Chloe Rolland, Laura Gustafson, Tete Xiao, Spencer Whitehead, Alexander~C. Berg, Wan-Yen Lo, Piotr Doll{\'{a}}r, and Ross Girshick.
\newblock {Segment Anything}.
\newblock \emph{arXiv preprint}, 2023.

\bibitem[Yang et~al.(2023{\natexlab{a}})Yang, Gao, Li, Gao, Wang, and Zheng]{yang2023track}
Jinyu Yang, Mingqi Gao, Zhe Li, Shang Gao, Fangjing Wang, and Feng Zheng.
\newblock {Track anything: Segment anything meets videos}.
\newblock \emph{arXiv preprint}, 2023{\natexlab{a}}.

\bibitem[Zhang et~al.(2023{\natexlab{a}})Zhang, Han, Qiao, Kim, Bae, Lee, and Hong]{zhang2023faster}
Chaoning Zhang, Dongshen Han, Yu~Qiao, Jung~Uk Kim, Sung-Ho Bae, Seungkyu Lee, and Choong~Seon Hong.
\newblock {Faster Segment Anything: Towards Lightweight SAM for Mobile Applications}.
\newblock \emph{arXiv preprint}, 2023{\natexlab{a}}.

\bibitem[Li et~al.(2022{\natexlab{a}})Li, Zhang, Zhang, Yang, Li, Zhong, Wang, Yuan, Zhang, Hwang, and {others}]{li2022grounded}
Liunian~Harold Li, Pengchuan Zhang, Haotian Zhang, Jianwei Yang, Chunyuan Li, Yiwu Zhong, Lijuan Wang, Lu~Yuan, Lei Zhang, Jenq-Neng Hwang, and {others}.
\newblock {Grounded language-image pre-training}.
\newblock In \emph{Proceedings of the IEEE/CVF Conference on Computer Vision and Pattern Recognition}, pages 10965--10975, 2022{\natexlab{a}}.

\bibitem[Li et~al.(2023)Li, Li, Savarese, and Hoi]{li2023blip}
Junnan Li, Dongxu Li, Silvio Savarese, and Steven Hoi.
\newblock {Blip-2: Bootstrapping language-image pre-training with frozen image encoders and large language models}.
\newblock \emph{arXiv preprint arXiv:2301.12597}, 2023.

\bibitem[Alayrac et~al.(2022)Alayrac, Donahue, Luc, Miech, Barr, Hasson, Lenc, Mensch, Millican, Reynolds, Ring, Rutherford, Cabi, Han, Gong, Samangooei, Monteiro, Menick, Borgeaud, Brock, Nematzadeh, Sharifzadeh, Binkowski, Barreira, Vinyals, Zisserman, and Simonyan]{alayrac2022flamingo}
Jean-Baptiste Alayrac, Jeff Donahue, Pauline Luc, Antoine Miech, Iain Barr, Yana Hasson, Karel Lenc, Arthur Mensch, Katie Millican, Malcolm Reynolds, Roman Ring, Eliza Rutherford, Serkan Cabi, Tengda Han, Zhitao Gong, Sina Samangooei, Marianne Monteiro, Jacob Menick, Sebastian Borgeaud, Andrew Brock, Aida Nematzadeh, Sahand Sharifzadeh, Mikolaj Binkowski, Ricardo Barreira, Oriol Vinyals, Andrew Zisserman, and Karen Simonyan.
\newblock {Flamingo: a Visual Language Model for Few-Shot Learning}.
\newblock In \emph{Advances in Neural Information Processing Systems}, pages 23716--23736, 2022.

\bibitem[Wang et~al.(2022)Wang, Yang, Men, Lin, Bai, Li, Ma, Zhou, Zhou, and Yang]{wang2022ofa}
Peng Wang, An~Yang, Rui Men, Junyang Lin, Shuai Bai, Zhikang Li, Jianxin Ma, Chang Zhou, Jingren Zhou, and Hongxia Yang.
\newblock {Ofa: Unifying architectures, tasks, and modalities through a simple sequence-to-sequence learning framework}.
\newblock In \emph{International Conference on Machine Learning}, pages 23318--23340, 2022.

\bibitem[Lu et~al.(2023)Lu, Clark, Zellers, Mottaghi, and Kembhavi]{lu2023unifiedio}
Jiasen Lu, Christopher Clark, Rowan Zellers, Roozbeh Mottaghi, and Aniruddha Kembhavi.
\newblock {UNIFIED-IO: A Unified Model for Vision, Language, and Multi-modal Tasks}.
\newblock In \emph{The Eleventh International Conference on Learning Representations}, 2023.
\newblock URL \url{https://openreview.net/forum?id=E01k9048soZ}.

\bibitem[Rombach et~al.(2022)Rombach, Blattmann, Lorenz, Esser, and Ommer]{StableDiffusion}
Robin Rombach, Andreas Blattmann, Dominik Lorenz, Patrick Esser, and Björn Ommer.
\newblock {High-Resolution Image Synthesis with Latent Diffusion Models}.
\newblock In \emph{Proceedings of the IEEE/CVF Conference on Computer Vision and Pattern Recognition}, 2022.

\bibitem[Ramesh et~al.(2021)Ramesh, Pavlov, Goh, Gray, Voss, Radford, Chen, and Sutskever]{ramesh2021zero}
Aditya Ramesh, Mikhail Pavlov, Gabriel Goh, Scott Gray, Chelsea Voss, Alec Radford, Mark Chen, and Ilya Sutskever.
\newblock {Zero-Shot Text-to-Image Generation}.
\newblock In \emph{International Conference on Machine Learning}, pages 8821--8831. PMLR, 2021.

\bibitem[Zhou et~al.(2022)Zhou, Girdhar, Joulin, Kr{\"{a}}henb{\"{u}}hl, and Misra]{zhou2022detecting}
Xingyi Zhou, Rohit Girdhar, Armand Joulin, Philipp Kr{\"{a}}henb{\"{u}}hl, and Ishan Misra.
\newblock {Detecting Twenty-thousand Classes using Image-level Supervision}.
\newblock In \emph{ECCV}, 2022.

\bibitem[Minderer et~al.(2022)Minderer, Gritsenko, Stone, Neumann, Weissenborn, Dosovitskiy, Mahendran, Arnab, Dehghani, Shen, Wang, Zhai, Kipf, and Houlsby]{minderer2022simple}
Matthias Minderer, Alexey Gritsenko, Austin Stone, Maxim Neumann, Dirk Weissenborn, Alexey Dosovitskiy, Aravindh Mahendran, Anurag Arnab, Mostafa Dehghani, Zhuoran Shen, Xiao Wang, Xiaohua Zhai, Thomas Kipf, and Neil Houlsby.
\newblock {Simple Open-Vocabulary Object Detection with Vision Transformers}.
\newblock In \emph{European Conference on Computer Vision}, 2022.

\bibitem[Liu et~al.(2021)Liu, Wu, Xiong, Chen, and Jiang]{liu2021unified}
Tianyi Liu, Zuxuan Wu, Wenhan Xiong, Jingjing Chen, and Yu-Gang Jiang.
\newblock Unified multimodal pre-training and prompt-based tuning for vision-language understanding and generation.
\newblock \emph{arXiv preprint arXiv:2112.05587}, 2021.

\bibitem[Li et~al.(2022{\natexlab{b}})Li, Weinberger, Belongie, Koltun, and Ranftl]{li2022languagedriven}
Boyi Li, Kilian~Q Weinberger, Serge Belongie, Vladlen Koltun, and Rene Ranftl.
\newblock {Language-driven Semantic Segmentation}.
\newblock In \emph{International Conference on Learning Representations}, 2022{\natexlab{b}}.
\newblock URL \url{https://openreview.net/forum?id=RriDjddCLN}.

\bibitem[Oquab et~al.(2023)Oquab, Darcet, Moutakanni, Vo, Szafraniec, Khalidov, Fernandez, Haziza, Massa, El-Nouby, Assran, Ballas, Galuba, Howes, Huang, Li, Misra, Rabbat, Sharma, Synnaeve, Xu, Jegou, Mairal, Labatut, Joulin, Bojanowski, and Research]{oquab2023dinov2}
Maxime Oquab, Timothée Darcet, Théo Moutakanni, Huy Vo, Marc Szafraniec, Vasil Khalidov, Pierre Fernandez, Daniel Haziza, Francisco Massa, Alaaeldin El-Nouby, Mahmoud Assran, Nicolas Ballas, Wojciech Galuba, Russell Howes, Po-Yao Huang, Shang-Wen Li, Ishan Misra, Michael Rabbat, Vasu Sharma, Gabriel Synnaeve, Hu~Xu, Hervé Jegou, Julien Mairal, Patrick Labatut, Armand Joulin, Piotr Bojanowski, and Meta~Ai Research.
\newblock {DINOv2: Learning Robust Visual Features without Supervision}.
\newblock \emph{arXiv preprint}, 2023.

\bibitem[Ma et~al.(2022)Ma, Xu, Sun, Yan, Zhang, and Ji]{ma2022xclip}
Yiwei Ma, Guohai Xu, Xiaoshuai Sun, Ming Yan, Ji~Zhang, and Rongrong Ji.
\newblock {X-clip: End-to-end multi-grained contrastive learning for video-text retrieval}.
\newblock In \emph{Proceedings of the 30th ACM International Conference on Multimedia}, pages 638--647, 2022.

\bibitem[Chai et~al.(2023)Chai, Guo, Wang, and Lu]{chai2023stablevideo}
Wenhao Chai, Xun Guo, Gaoang Wang, and Yan Lu.
\newblock {Stablevideo: Text-driven consistency-aware diffusion video editing}.
\newblock In \emph{Proceedings of the IEEE/CVF International Conference on Computer Vision}, pages 23040--23050, 2023.

\bibitem[Elizalde et~al.(2023)Elizalde, Deshmukh, Al~Ismail, and Wang]{elizalde2023clap}
Benjamin Elizalde, Soham Deshmukh, Mahmoud Al~Ismail, and Huaming Wang.
\newblock {CLAP: Learning Audio Concepts from Natural Language Supervision}.
\newblock In \emph{ICASSP 2023-2023 IEEE International Conference on Acoustics, Speech and Signal Processing (ICASSP)}, pages 1--5, 2023.

\bibitem[Guzhov et~al.(2022)Guzhov, Raue, Hees, and Dengel]{guzhov2022audioclip}
Andrey Guzhov, Federico Raue, Jörn Hees, and Andreas Dengel.
\newblock {Audioclip: Extending clip to image, text and audio}.
\newblock In \emph{ICASSP 2022-2022 IEEE International Conference on Acoustics, Speech and Signal Processing (ICASSP)}, pages 976--980, 2022.

\bibitem[Radford et~al.(2023)Radford, Kim, Xu, Brockman, McLeavey, and Sutskever]{radford2023robust}
Alec Radford, Jong~Wook Kim, Tao Xu, Greg Brockman, Christine McLeavey, and Ilya Sutskever.
\newblock {Robust speech recognition via large-scale weak supervision}.
\newblock In \emph{International Conference on Machine Learning}, pages 28492--28518, 2023.

\bibitem[Agostinelli et~al.(2023)Agostinelli, Denk, Borsos, Engel, Verzetti, Caillon, Huang, Jansen, Roberts, Tagliasacchi, and {others}]{agostinelli2023musiclm}
Andrea Agostinelli, Timo~I Denk, Zalán Borsos, Jesse Engel, Mauro Verzetti, Antoine Caillon, Qingqing Huang, Aren Jansen, Adam Roberts, Marco Tagliasacchi, and {others}.
\newblock {MusicLM: Generating music from text}.
\newblock \emph{arXiv preprint}, 2023.

\bibitem[Wang et~al.(2023{\natexlab{a}})Wang, Chen, Wu, Zhang, Zhou, Liu, Chen, Liu, Wang, Li, and {others}]{wang2023neural}
Chengyi Wang, Sanyuan Chen, Yu~Wu, Ziqiang Zhang, Long Zhou, Shujie Liu, Zhuo Chen, Yanqing Liu, Huaming Wang, Jinyu Li, and {others}.
\newblock {Neural codec language models are zero-shot text to speech synthesizers}.
\newblock \emph{arXiv preprint}, 2023{\natexlab{a}}.

\bibitem[Chen et~al.(2024{\natexlab{a}})Chen, Li, Wang, Zhao, Sun, Zhu, and Liu]{chen2024vast}
Sihan Chen, Handong Li, Qunbo Wang, Zijia Zhao, Mingzhen Sun, Xinxin Zhu, and Jing Liu.
\newblock Vast: A vision-audio-subtitle-text omni-modality foundation model and dataset.
\newblock \emph{Advances in Neural Information Processing Systems}, 36, 2024{\natexlab{a}}.

\bibitem[Xue et~al.(2023)Xue, Gao, Xing, Mart{\'{i}}n-Mart{\'{i}}n, Wu, Xiong, Xu, Niebles, and Savarese]{xue2023ulip}
Le~Xue, Mingfei Gao, Chen Xing, Roberto Mart{\'{i}}n-Mart{\'{i}}n, Jiajun Wu, Caiming Xiong, Ran Xu, Juan~Carlos Niebles, and Silvio Savarese.
\newblock {ULIP: Learning a unified representation of language, images, and point clouds for 3D understanding}.
\newblock In \emph{Proceedings of the IEEE/CVF Conference on Computer Vision and Pattern Recognition}, pages 1179--1189, 2023.

\bibitem[Hegde et~al.(2023)Hegde, Valanarasu, and Patel]{hegde2023clip}
Deepti Hegde, Jeya Maria~Jose Valanarasu, and Vishal Patel.
\newblock {Clip goes 3d: Leveraging prompt tuning for language grounded 3d recognition}.
\newblock In \emph{Proceedings of the IEEE/CVF International Conference on Computer Vision}, pages 2028--2038, 2023.

\bibitem[Wu et~al.(2023{\natexlab{b}})Wu, Zhang, Fu, Wang, Ren, Pan, Wu, Yang, Wang, Qian, and {others}]{wu2023omniobject3d}
Tong Wu, Jiarui Zhang, Xiao Fu, Yuxin Wang, Jiawei Ren, Liang Pan, Wayne Wu, Lei Yang, Jiaqi Wang, Chen Qian, and {others}.
\newblock {Omniobject3d: Large-vocabulary 3d object dataset for realistic perception, reconstruction and generation}.
\newblock In \emph{Proceedings of the IEEE/CVF Conference on Computer Vision and Pattern Recognition}, pages 803--814, 2023{\natexlab{b}}.

\bibitem[Nichol et~al.(2022)Nichol, Jun, Dhariwal, Mishkin, and Chen]{nichol2022point}
Alex Nichol, Heewoo Jun, Prafulla Dhariwal, Pamela Mishkin, and Mark Chen.
\newblock {Point-e: A system for generating 3d point clouds from complex prompts}.
\newblock \emph{arXiv preprint arXiv:2212.08751}, 2022.

\bibitem[Cao et~al.(2023)Cao, Kreis, Fidler, Sharp, and Yin]{cao2023texfusion}
Tianshi Cao, Karsten Kreis, Sanja Fidler, Nicholas Sharp, and Kangxue Yin.
\newblock Texfusion: Synthesizing 3d textures with text-guided image diffusion models.
\newblock In \emph{Proceedings of the IEEE/CVF International Conference on Computer Vision}, pages 4169--4181, 2023.

\bibitem[Hong et~al.(2023)Hong, Zhen, Chen, Zheng, Du, Chen, and Gan]{hong20233d}
Yining Hong, Haoyu Zhen, Peihao Chen, Shuhong Zheng, Yilun Du, Zhenfang Chen, and Chuang Gan.
\newblock {3D-LLM: Injecting the 3d world into large language models}.
\newblock \emph{arXiv preprint}, 2023.

\bibitem[Peng et~al.(2023)Peng, Genova, Jiang, Tagliasacchi, Pollefeys, Funkhouser, and {others}]{peng2023openscene}
Songyou Peng, Kyle Genova, Chiyu Jiang, Andrea Tagliasacchi, Marc Pollefeys, Thomas Funkhouser, and {others}.
\newblock {Openscene: 3d scene understanding with open vocabularies}.
\newblock In \emph{Proceedings of the IEEE/CVF Conference on Computer Vision and Pattern Recognition}, pages 815--824, 2023.

\bibitem[Chen et~al.(2024{\natexlab{b}})Chen, Xu, Kirmani, Driess, Florence, Ichter, Sadigh, Guibas, and Xia]{chen2024spatialvlm}
Boyuan Chen, Zhuo Xu, Sean Kirmani, Danny Driess, Pete Florence, Brian Ichter, Dorsa Sadigh, Leonidas Guibas, and Fei Xia.
\newblock {SpatialVLM: Endowing Vision-Language Models with Spatial Reasoning Capabilities}.
\newblock \emph{arXiv preprint}, 2024{\natexlab{b}}.

\bibitem[Girdhar et~al.(2023)Girdhar, El-Nouby, Liu, Singh, Alwala, Joulin, and Misra]{girdhar2023imagebind}
Rohit Girdhar, Alaaeldin El-Nouby, Zhuang Liu, Mannat Singh, Kalyan~Vasudev Alwala, Armand Joulin, and Ishan Misra.
\newblock {Imagebind: One embedding space to bind them all}.
\newblock In \emph{Proceedings of the IEEE/CVF Conference on Computer Vision and Pattern Recognition}, pages 15180--15190, 2023.

\bibitem[Zhang et~al.(2023{\natexlab{b}})Zhang, Gong, Zhang, Li, Qiao, Ouyang, and Yue]{zhang2023meta}
Yiyuan Zhang, Kaixiong Gong, Kaipeng Zhang, Hongsheng Li, Yu~Qiao, Wanli Ouyang, and Xiangyu Yue.
\newblock {Meta-transformer: A unified framework for multimodal learning}.
\newblock \emph{arXiv preprint}, 2023{\natexlab{b}}.

\bibitem[Wen et~al.(2023)Wen, Yang, Kautz, and Nvidia]{wen2023foundationpose}
Bowen Wen, Wei Yang, Jan Kautz, and Stan~Birchfield Nvidia.
\newblock {FoundationPose: Unified 6D Pose Estimation and Tracking of Novel Objects}.
\newblock \emph{arXiv preprint}, 2023.

\bibitem[Tevet et~al.(2023)Tevet, Raab, Gordon, Shafir, Cohen-or, and Bermano]{tevet2023human}
Guy Tevet, Sigal Raab, Brian Gordon, Yoni Shafir, Daniel Cohen-or, and Amit~Haim Bermano.
\newblock {Human Motion Diffusion Model}.
\newblock In \emph{The Eleventh International Conference on Learning Representations}, 2023.

\bibitem[Zhang et~al.(2023{\natexlab{c}})Zhang, Zhang, Cun, Huang, Zhang, Zhao, Lu, and Shen]{zhang2023generating}
Jianrong Zhang, Yangsong Zhang, Xiaodong Cun, Shaoli Huang, Yong Zhang, Hongwei Zhao, Hongtao Lu, and Xi~Shen.
\newblock {T2M-GPT: Generating Human Motion from Textual Descriptions with Discrete Representations}.
\newblock In \emph{Proceedings of the IEEE/CVF Conference on Computer Vision and Pattern Recognition}, 2023{\natexlab{c}}.

\bibitem[Ao et~al.()Ao, Zhang, and Liu]{Ao2023GestureDiffuCLIP}
Tenglong Ao, Zeyi Zhang, and Libin Liu.
\newblock {GestureDiffuCLIP: Gesture Diffusion Model with CLIP Latents}.
\newblock \emph{ACM Trans. Graph.}, 42\penalty0 (4).

\bibitem[Shridhar et~al.(2021)Shridhar, Manuelli, and Fox]{shridhar2021cliport}
Mohit Shridhar, Lucas Manuelli, and Dieter Fox.
\newblock {CLIPort: What and Where Pathways for Robotic Manipulation}.
\newblock In \emph{Proceedings of the 5th Conference on Robot Learning (CoRL)}, 2021.

\bibitem[Liu et~al.(2023{\natexlab{a}})Liu, Bahety, and Song]{liu2023reflect}
Zeyi Liu, Arpit Bahety, and Shuran Song.
\newblock {REFLECT: Summarizing Robot Experiences for FaiLure Explanation and CorrecTion Examples in the RoboFail Benchmark}.
\newblock In \emph{Conference on Robot Learning (CoRL)}, 2023{\natexlab{a}}.
\newblock URL \url{https://roboreflect.github.io/}.

\bibitem[Zeng et~al.(2020)Zeng, Florence, Tompson, Welker, Chien, Attarian, Armstrong, Krasin, Duong, Wahid, Sindhwani, and Lee]{zeng2020transporter}
Andy Zeng, Pete Florence, Jonathan Tompson, Stefan Welker, Jonathan Chien, Maria Attarian, Travis Armstrong, Ivan Krasin, Dan Duong, Ayzaan Wahid, Vikas Sindhwani, and Johnny Lee.
\newblock {Transporter Networks: Rearranging the Visual World for Robotic Manipulation}.
\newblock In \emph{Conference on Robot Learning (CoRL)}, 2020.

\bibitem[Kamath et~al.(2021)Kamath, Singh, LeCun, Synnaeve, Misra, and Carion]{kamath2021mdetr}
Aishwarya Kamath, Mannat Singh, Yann LeCun, Gabriel Synnaeve, Ishan Misra, and Nicolas Carion.
\newblock {Mdetr-modulated detection for end-to-end multi-modal understanding}.
\newblock In \emph{Proceedings of the IEEE/CVF International Conference on Computer Vision}, pages 1780--1790, 2021.

\bibitem[Huang et~al.(2023{\natexlab{a}})Huang, Wang, Zhang, Li, Wu, and Fei-Fei]{huang2023voxposer}
Wenlong Huang, Chen Wang, Ruohan Zhang, Yunzhu Li, Jiajun Wu, and Li~Fei-Fei.
\newblock {VoxPoser: Composable 3D Value Maps for Robotic Manipulation with Language Models}.
\newblock \emph{arXiv preprint}, 2023{\natexlab{a}}.

\bibitem[Shafiullah et~al.(2023)Shafiullah, Paxton, Pinto, Chintala, and Szlam]{shafiullah2022clipfields}
Nur Muhammad~Mahi Shafiullah, Chris Paxton, Lerrel Pinto, Soumith Chintala, and Arthur Szlam.
\newblock {CLIP-Fields: Weakly Supervised Semantic Fields for Robotic Memory}.
\newblock In \emph{Robotics: Science and Systems}, 2023.

\bibitem[Ma et~al.(2023{\natexlab{a}})Ma, Liang, Wang, Huang, Bastani, Jayaraman, Zhu, Fan, and Anandkumar]{ma2023eureka}
Yecheng~Jason Ma, William Liang, Guanzhi Wang, De-An Huang, Osbert Bastani, Dinesh Jayaraman, Yuke Zhu, Linxi Fan, and Anima Anandkumar.
\newblock {Eureka: Human-Level Reward Design via Coding Large Language Models}.
\newblock \emph{arXiv preprint arXiv:2310.12931}, 2023{\natexlab{a}}.

\bibitem[Reimers and Gurevych(2019)]{reimers2019sentence}
Nils Reimers and Iryna Gurevych.
\newblock {Sentence-bert: Sentence embeddings using siamese bert-networks}.
\newblock \emph{Proceedings of the 2019 Conference on Empirical Methods in Natural Language Processing and the 9th International Joint Conference on Natural Language Processing (EMNLP-IJCNLP)}, pages 3982--3992, 2019.
\newblock URL \url{https://aclanthology.org/D19-1410}.

\bibitem[Gu et~al.(2022)Gu, Lin, Kuo, and Cui]{gu2022open}
Xiuye Gu, Tsung-Yi Lin, Weicheng Kuo, and Yin Cui.
\newblock {Open-vocabulary Object Detection via Vision and Language Knowledge Distillation}.
\newblock In \emph{International Conference on Learning Representations}, 2022.

\bibitem[Tang et~al.(2023)Tang, Yu, Tan, Zen, Faust, and Harada]{tang2023saytap}
Yujin Tang, Wenhao Yu, Jie Tan, Heiga Zen, Aleksandra Faust, and Tatsuya Harada.
\newblock {SayTap: Language to Quadrupedal Locomotion}.
\newblock In \emph{7th Annual Conference on Robot Learning}, 2023.
\newblock URL \url{https://openreview.net/forum?id=7TYeO2XVqI}.

\bibitem[Mirchandani et~al.(2023)Mirchandani, Xia, Florence, Ichter, Driess, Arenas, Rao, Sadigh, and Zeng]{mirchandani2023generalpatternmachines}
Suvir Mirchandani, Fei Xia, Pete Florence, Brian Ichter, Danny Driess, Montserrat~Gonzalez Arenas, Kanishka Rao, Dorsa Sadigh, and Andy Zeng.
\newblock {Large Language Models as General Pattern Machines}.
\newblock In \emph{Proceedings of the 7th Conference on Robot Learning (CoRL)}, 2023.

\bibitem[Chen et~al.(2023{\natexlab{a}})Chen, Kiami, Gupta, and Kumar]{chen2023genaug}
Zoey Chen, Sho Kiami, Abhishek Gupta, and Vikash Kumar.
\newblock {GenAug: Retargeting behaviors to unseen situations via Generative Augmentation}.
\newblock In \emph{Robotics: Science and Systems}, 2023{\natexlab{a}}.

\bibitem[Xiao et~al.(2023)Xiao, Chan, Sermanet, Wahid, Brohan, Hausman, Levine, and Tompson]{xiao2023robotic}
Ted Xiao, Harris Chan, Pierre Sermanet, Ayzaan Wahid, Anthony Brohan, Karol Hausman, Sergey Levine, and Jonathan Tompson.
\newblock {Robotic skill acquisition via instruction augmentation with vision-language models}.
\newblock In \emph{Robotics: Science and Systems}, 2023.

\bibitem[He et~al.(2016)He, Zhang, Ren, and Sun]{he2016residual}
Kaiming He, Xiangyu Zhang, Shaoqing Ren, and Jian Sun.
\newblock {Deep Residual Learning for Image Recognition}.
\newblock In \emph{Proceedings of the IEEE conference on Computer Vision and Pattern Recognition}, pages 770--778, 2016.

\bibitem[He et~al.(2020)He, Fan, Wu, Xie, and Girshick]{he2020momentum}
Kaiming He, Haoqi Fan, Yuxin Wu, Saining Xie, and Ross Girshick.
\newblock {Momentum contrast for unsupervised visual representation learning}.
\newblock In \emph{Proceedings of the IEEE/CVF conference on computer vision and pattern recognition}, pages 9729--9738, 2020.

\bibitem[Shah and Kumar(2021)]{shah2021rrl}
Rutav Shah and Vikash Kumar.
\newblock {RRL: Resnet as representation for Reinforcement Learning}.
\newblock In \emph{International Conference on Machine Learning}. PMLR, 2021.

\bibitem[{Jia Deng} et~al.(2009){Jia Deng}, {Wei Dong}, Socher, {Li-Jia Li}, {Kai Li}, and {Li Fei-Fei}]{JiaDeng2009}
{Jia Deng}, {Wei Dong}, R.~Socher, {Li-Jia Li}, {Kai Li}, and {Li Fei-Fei}.
\newblock {ImageNet: A large-scale hierarchical image database}.
\newblock In \emph{2009 IEEE Conference on Computer Vision and Pattern Recognition}, pages 248--255, 2009.

\bibitem[Grauman et~al.(2022)Grauman, Westbury, Byrne, Chavis, Furnari, Girdhar, Hamburger, Jiang, Liu, Liu, Martin, Nagarajan, Radosavovic, Ramakrishnan, Ryan, Sharma, Wray, Xu, Xu, Zhao, Bansal, Batra, Cartillier, Crane, Do, Doulaty, Erapalli, Feichtenhofer, Fragomeni, Fu, Gebreselasie, Gonzalez, Hillis, Huang, Huang, Jia, Khoo, Kolar, Kottur, Kumar, Landini, Li, Li, Li, Mangalam, Modhugu, Munro, Murrell, Nishiyasu, Price, Puentes, Ramazanova, Sari, Somasundaram, Southerland, Sugano, Tao, Vo, Wang, Wu, Yagi, Zhao, Zhu, Arbelaez, Crandall, Damen, Farinella, Fuegen, Ghanem, Ithapu, Jawahar, Joo, Kitani, Li, Newcombe, Oliva, Park, Rehg, Sato, Shi, Shou, Torralba, Torresani, Yan, and Malik]{grauman2022ego4d}
Kristen Grauman, Andrew Westbury, Eugene Byrne, Zachary Chavis, Antonino Furnari, Rohit Girdhar, Jackson Hamburger, Hao Jiang, Miao Liu, Xingyu Liu, Miguel Martin, Tushar Nagarajan, Ilija Radosavovic, Santhosh~Kumar Ramakrishnan, Fiona Ryan, Jayant Sharma, Michael Wray, Mengmeng Xu, Eric~Zhongcong Xu, Chen Zhao, Siddhant Bansal, Dhruv Batra, Vincent Cartillier, Sean Crane, Tien Do, Morrie Doulaty, Akshay Erapalli, Christoph Feichtenhofer, Adriano Fragomeni, Qichen Fu, Abrham Gebreselasie, Cristina Gonzalez, James Hillis, Xuhua Huang, Yifei Huang, Wenqi Jia, Weslie Khoo, Jachym Kolar, Satwik Kottur, Anurag Kumar, Federico Landini, Chao Li, Yanghao Li, Zhenqiang Li, Karttikeya Mangalam, Raghava Modhugu, Jonathan Munro, Tullie Murrell, Takumi Nishiyasu, Will Price, Paola~Ruiz Puentes, Merey Ramazanova, Leda Sari, Kiran Somasundaram, Audrey Southerland, Yusuke Sugano, Ruijie Tao, Minh Vo, Yuchen Wang, Xindi Wu, Takuma Yagi, Ziwei Zhao, Yunyi Zhu, Pablo Arbelaez, David Crandall, Dima Damen, Giovanni~Maria
  Farinella, Christian Fuegen, Bernard Ghanem, Vamsi~Krishna Ithapu, C.~V. Jawahar, Hanbyul Joo, Kris Kitani, Haizhou Li, Richard Newcombe, Aude Oliva, Hyun~Soo Park, James~M. Rehg, Yoichi Sato, Jianbo Shi, Mike~Zheng Shou, Antonio Torralba, Lorenzo Torresani, Mingfei Yan, and Jitendra Malik.
\newblock {Ego4D: Around the World in 3,000 Hours of Egocentric Video}.
\newblock In \emph{IEEE/CVF Conference on Computer Vision and Pattern Recognition (CVPR)}, pages 18995--19012, 2022.

\bibitem[Mandi et~al.(2022)Mandi, Bharadhwaj, Moens, Song, Rajeswaran, and Kumar]{mandi2022cacti}
Zhao Mandi, Homanga Bharadhwaj, Vincent Moens, Shuran Song, Aravind Rajeswaran, and Vikash Kumar.
\newblock {CACTI: A Framework for Scalable Multi-Task Multi-Scene Visual Imitation Learning}.
\newblock In \emph{CoRL 2022 Workshop on Pre-training Robot Learning}, 2022.
\newblock URL \url{https://openreview.net/forum?id=dRHW9-QFj9}.

\bibitem[Zeng et~al.(2023{\natexlab{b}})Zeng, Attarian, Ichter, Choromanski, Wong, Welker, Tombari, Purohit, Ryoo, Sindhwani, Lee, Vanhoucke, and Florence]{zeng2023socraticmodels}
Andy Zeng, Maria Attarian, Brian Ichter, Krzysztof Choromanski, Adrian Wong, Stefan Welker, Federico Tombari, Aveek Purohit, Michael Ryoo, Vikas Sindhwani, Johnny Lee, Vincent Vanhoucke, and Pete Florence.
\newblock {Socratic Models: Composing Zero-Shot Multimodal Reasoning with Language}.
\newblock In \emph{International Conference on Learning Representations}, 2023{\natexlab{b}}.

\bibitem[Chen et~al.(2023{\natexlab{b}})Chen, Xia, Ichter, Rao, Gopalakrishnan, Ryoo, Stone, and Kappler]{chen2023nlmapsaycan}
Boyuan Chen, Fei Xia, Brian Ichter, Kanishka Rao, Keerthana Gopalakrishnan, Michael~S. Ryoo, Austin Stone, and Daniel Kappler.
\newblock {Open-vocabulary Queryable Scene Representations for Real World Planning}.
\newblock In \emph{International Conference on Robotics and Automation}, 2023{\natexlab{b}}.

\bibitem[Lin et~al.(2023{\natexlab{a}})Lin, Cui, Hao, Xia, and Sadigh]{lin2023gestureinformed}
Li-Heng Lin, Yuchen Cui, Yilun Hao, Fei Xia, and Dorsa Sadigh.
\newblock {Gesture-Informed Robot Assistance via Foundation Models}.
\newblock In \emph{7th Annual Conference on Robot Learning}, 2023{\natexlab{a}}.
\newblock URL \url{https://openreview.net/forum?id=Ffn8Z4Q-zU}.

\bibitem[Ilharco et~al.(2021)Ilharco, Wortsman, Wightman, Gordon, Carlini, Taori, Dave, Shankar, Namkoong, Miller, Hajishirzi, Farhadi, and Schmidt]{openclip2021}
Gabriel Ilharco, Mitchell Wortsman, Ross Wightman, Cade Gordon, Nicholas Carlini, Rohan Taori, Achal Dave, Vaishaal Shankar, Hongseok Namkoong, John Miller, Hannaneh Hajishirzi, Ali Farhadi, and Ludwig Schmidt.
\newblock {OpenCLIP}, 2021.
\newblock URL \url{https://doi.org/10.5281/zenodo.5143773}.

\bibitem[Kawaharazuka et~al.(2023{\natexlab{a}})Kawaharazuka, Obinata, Kanazawa, Okada, and Inaba]{kawaharazuka2023vqa}
Kento Kawaharazuka, Yoshiki Obinata, Naoaki Kanazawa, Kei Okada, and Masayuki Inaba.
\newblock {VQA-based Robotic State Recognition Optimized with Genetic Algorithm}.
\newblock In \emph{2023 IEEE International Conference on Robotics and Automation (ICRA)}, pages 8306--8311, 2023{\natexlab{a}}.

\bibitem[Kanazawa et~al.(2023)Kanazawa, Kawaharazuka, Obinata, Okada, and Inaba]{kanazawa2023recognition}
Naoaki Kanazawa, Kento Kawaharazuka, Yoshiki Obinata, Kei Okada, and Masayuki Inaba.
\newblock {Recognition of Heat-Induced Food State Changes by Time-Series Use of Vision-Language Model for Cooking Robot}.
\newblock \emph{arXiv preprint arXiv:2309.01528}, 2023.

\bibitem[Kapelyukh et~al.(2023)Kapelyukh, Vosylius, and Johns]{kapelyukh2023dallebot}
Ivan Kapelyukh, Vitalis Vosylius, and Edward Johns.
\newblock {DALL-E-Bot: Introducing Web-Scale Diffusion Models to Robotics}.
\newblock \emph{IEEE Robotics and Automation Letters (RA-L)}, 2023.

\bibitem[Liu et~al.(2023{\natexlab{b}})Liu, Du, Hermans, Chernova, and Paxton]{structdiffusion2023}
Weiyu Liu, Yilun Du, Tucker Hermans, Sonia Chernova, and Chris Paxton.
\newblock {StructDiffusion: Language-Guided Creation of Physically-Valid Structures using Unseen Objects}.
\newblock In \emph{RSS 2023}, 2023{\natexlab{b}}.

\bibitem[Black et~al.(2023)Black, Nakamoto, Atreya, Walke, Finn, Kumar, and Levine]{black2023zeroshot}
Kevin Black, Mitsuhiko Nakamoto, Pranav Atreya, Homer Walke, Chelsea Finn, Aviral Kumar, and Sergey Levine.
\newblock {Zero-Shot Robotic Manipulation with Pretrained Image-Editing Diffusion Models}.
\newblock \emph{arXiv preprint arXiv:2310.10639}, 2023.

\bibitem[Brooks et~al.(2023)Brooks, Holynski, and Efros]{brooks2023instructpix2pix}
Tim Brooks, Aleksander Holynski, and Alexei~A Efros.
\newblock {Instructpix2pix: Learning to follow image editing instructions}.
\newblock In \emph{Proceedings of the IEEE/CVF Conference on Computer Vision and Pattern Recognition}, pages 18392--18402, 2023.

\bibitem[Du et~al.(2023)Du, Yang, Dai, Dai, Nachum, Tenenbaum, Schuurmans, and Abbeel]{du2023learning}
Yilun Du, Mengjiao Yang, Bo~Dai, Hanjun Dai, Ofir Nachum, Joshua~B. Tenenbaum, Dale Schuurmans, and Pieter Abbeel.
\newblock {Learning Universal Policies via Text-Guided Video Generation}.
\newblock \emph{arXiv preprint}, 2023.

\bibitem[Kwon et~al.(2023)Kwon, Xie, Bullard, and Sadigh]{kwon2023rewarddesign}
Minae Kwon, Sang~Michael Xie, Kalesha Bullard, and Dorsa Sadigh.
\newblock {Reward Design with Language Models}.
\newblock In \emph{International Conference on Learning Representations}, 2023.

\bibitem[Yu et~al.(2023{\natexlab{a}})Yu, Gileadi, Fu, Kirmani, Lee, Arenas, Chiang, Erez, Hasenclever, Humplik, Ichter, Xiao, Xu, Zeng, Zhang, Heess, Sadigh, Tan, Tassa, and Xia]{yu2023language}
Wenhao Yu, Nimrod Gileadi, Chuyuan Fu, Sean Kirmani, Kuang-Huei Lee, Montse~Gonzalez Arenas, Hao-Tien~Lewis Chiang, Tom Erez, Leonard Hasenclever, Jan Humplik, Brian Ichter, Ted Xiao, Peng Xu, Andy Zeng, Tingnan Zhang, Nicolas Heess, Dorsa Sadigh, Jie Tan, Yuval Tassa, and Fei Xia.
\newblock {Language to Rewards for Robotic Skill Synthesis}.
\newblock \emph{7th Annual Conference on Robot Learning}, 2023{\natexlab{a}}.
\newblock URL \url{https://openreview.net/forum?id=SgTPdyehXMA}.

\bibitem[Lin et~al.(2023{\natexlab{b}})Lin, Du, Watkins, Hafner, Abbeel, Klein, and Dragan]{lin2023learning}
Jessy Lin, Yuqing Du, Olivia Watkins, Danijar Hafner, Pieter Abbeel, Dan Klein, and Anca Dragan.
\newblock {Learning to Model the World with Language}.
\newblock \emph{arXiv preprint}, 2023{\natexlab{b}}.

\bibitem[Hafner et~al.(2023)Hafner, Pasukonis, Ba, and Lillicrap]{hafner2023dreamerv3}
Danijar Hafner, Jurgis Pasukonis, Jimmy Ba, and Timothy Lillicrap.
\newblock {Mastering Diverse Domains through World Models}.
\newblock \emph{arXiv preprint}, 2023.

\bibitem[Cui et~al.(2022)Cui, Niekum, Gupta, Kumar, and Rajeswaran]{cui2022can}
Yuchen Cui, Scott Niekum, Abhinav Gupta, Vikash Kumar, and Aravind Rajeswaran.
\newblock {Can Foundation Models Perform Zero-Shot Task Specification For Robot Manipulation?}
\newblock In \emph{Learning for Dynamics and Control Conference}, pages 893--905, 2022.

\bibitem[Damen et~al.(2022)Damen, Doughty, Farinella, Furnari, Kazakos, Ma, Moltisanti, Munro, Perrett, Price, and Wray]{damen2022rescaling}
Dima Damen, Hazel Doughty, Giovanni~Maria Farinella, Antonino Furnari, Evangelos Kazakos, Jian Ma, Davide Moltisanti, Jonathan Munro, Toby Perrett, Will Price, and Michael Wray.
\newblock {Rescaling Egocentric Vision: Collection, Pipeline and Challenges for EPIC-KITCHENS-100}.
\newblock \emph{International Journal of Computer Vision}, 130\penalty0 (1):\penalty0 33--55, 2022.
\newblock ISSN 0920-5691.
\newblock \doi{10.1007/s11263-021-01531-2}.

\bibitem[Bahl et~al.(2023)Bahl, Mendonca, Chen, Jain, and Pathak]{bahl2023affordances}
Shikhar Bahl, Russell Mendonca, Lili Chen, Unnat Jain, and Deepak Pathak.
\newblock {Affordances from Human Videos as a Versatile Representation for Robotics}.
\newblock In \emph{CVPR}, 2023.

\bibitem[Ma et~al.(2023{\natexlab{b}})Ma, Sodhani, Jayaraman, Bastani, Kumar, and Zhang]{ma2023vip}
Yecheng~Jason Ma, Shagun Sodhani, Dinesh Jayaraman, Osbert Bastani, Vikash Kumar, and Amy Zhang.
\newblock {VIP: Towards Universal Visual Reward and Representation via Value-Implicit Pre-Training}.
\newblock In \emph{International Conference on Learning Representations}, 2023{\natexlab{b}}.

\bibitem[Ma et~al.(2023{\natexlab{c}})Ma, Liang, Som, Kumar, Zhang, Bastani, and Jayaraman]{ma2023liv}
Yecheng~Jason Ma, William Liang, Vaidehi Som, Vikash Kumar, Amy Zhang, Osbert Bastani, and Dinesh Jayaraman.
\newblock {LIV: Language-Image Representations and Rewards for Robotic Control}.
\newblock In \emph{International Conference on Machine Learning}, 2023{\natexlab{c}}.

\bibitem[Shah et~al.(2022)Shah, Osinski, Ichter, and Levine]{shah2022lmnav}
Dhruv Shah, Blazej Osinski, Brian Ichter, and Sergey Levine.
\newblock {LM-Nav: Robotic Navigation with Large Pre-Trained Models of Language, Vision, and Action}.
\newblock In \emph{6th Annual Conference on Robot Learning}, 2022.

\bibitem[Gadre et~al.(2022)Gadre, Wortsman, Ilharco, Schmidt, and Song]{gadre2022clip}
Samir~Yitzhak Gadre, Mitchell Wortsman, Gabriel Ilharco, Ludwig Schmidt, and Shuran Song.
\newblock {Clip on wheels: Zero-shot object navigation as object localization and exploration}.
\newblock \emph{arXiv preprint arXiv:2203.10421}, 3\penalty0 (4):\penalty0 7, 2022.

\bibitem[Mildenhall et~al.(2020)Mildenhall, Srinivasan, Tancik, Barron, Ramamoorthi, and Ng]{mildenhall2020nerf}
Ben Mildenhall, Pratul~P. Srinivasan, Matthew Tancik, Jonathan~T. Barron, Ravi Ramamoorthi, and Ren Ng.
\newblock {NeRF: Representing Scenes as Neural Radiance Fields for View Synthesis}.
\newblock In \emph{ECCV}, volume 12346 LNCS, 2020.
\newblock \doi{10.1007/978-3-030-58452-8{\_}24}.

\bibitem[M{\"{u}}ller et~al.(2022)M{\"{u}}ller, Evans, Schied, and Keller]{mueller2022instant}
Thomas M{\"{u}}ller, Alex Evans, Christoph Schied, and Alexander Keller.
\newblock {Instant neural graphics primitives with a multiresolution hash encoding}.
\newblock \emph{ACM Transactions on Graphics}, 41\penalty0 (4):\penalty0 1--15, 2022.

\bibitem[Huang et~al.(2023{\natexlab{b}})Huang, Mees, Zeng, and Burgard]{huang23avlmaps}
Chenguang Huang, Oier Mees, Andy Zeng, and Wolfram Burgard.
\newblock {Audio Visual Language Maps for Robot Navigation}.
\newblock In \emph{International Symposium on Experimental Robotics (ISER)}, 2023{\natexlab{b}}.

\bibitem[Huang et~al.(2023{\natexlab{c}})Huang, Mees, Zeng, and Burgard]{huang23vlmaps}
Chenguang Huang, Oier Mees, Andy Zeng, and Wolfram Burgard.
\newblock {Visual Language Maps for Robot Navigation}.
\newblock In \emph{IEEE International Conference on Robotics and Automation (ICRA)}, 2023{\natexlab{c}}.

\bibitem[Jatavallabhula et~al.(2023)Jatavallabhula, Kuwajerwala, Gu, Omama, Chen, Maalouf, Li, Iyer, Keetha, Tewari, Tenenbaum, de~Melo, Krishna, Paull, Shkurti, and Torralba]{jatavallabhula2023conceptfusion}
Krishna~Murthy Jatavallabhula, Alihusein Kuwajerwala, Qiao Gu, Mohd Omama, Tao Chen, Alaa Maalouf, Shuang Li, Ganesh~Subramanian Iyer, Nikhil~Varma Keetha, Ayush Tewari, Joshua~B Tenenbaum, Celso~M de~Melo, Madhava Krishna, Liam Paull, Florian Shkurti, and Antonio Torralba.
\newblock {ConceptFusion: Open-set Multimodal 3D Mapping}.
\newblock In \emph{Robotics: Science and Systems (RSS) 2023}, 2023.

\bibitem[Yamazaki et~al.(2024)Yamazaki, Hanyu, Vo, Pham, Tran, Doretto, Nguyen, and Le]{yamazaki2024openfusion}
Kashu Yamazaki, Taisei Hanyu, Khoa Vo, Thang Pham, Minh Tran, Gianfranco Doretto, Anh Nguyen, and Ngan Le.
\newblock {Open-Fusion: Real-time Open-Vocabulary 3D Mapping and Queryable Scene Representation}.
\newblock In \emph{IEEE International Conference on Robotics and Automation (ICRA)}, 2024.

\bibitem[Gu et~al.(2023{\natexlab{a}})Gu, Kuwajerwala, Jatavallabhula, Sen, Agarwal, Rivera, Paul, Chellappa, Gan, de~Melo, Tenenbaum, Torralba, Shkurti, and Paull]{gu2023conceptgraphs}
Qiao Gu, Alihusein Kuwajerwala, Krishna~Murthy Jatavallabhula, Bipasha Sen, Aditya Agarwal, Corban Rivera, William Paul, Rama Chellappa, Chuang Gan, Celso~M de~Melo, Joshua~B Tenenbaum, Antonio Torralba, Florian Shkurti, and Liam Paull.
\newblock {ConceptGraphs: Open-Vocabulary 3D Scene Graphs for Perception and Planning}.
\newblock In \emph{2nd Workshop on Language and Robot Learning: Language as Grounding}, 2023{\natexlab{a}}.
\newblock URL \url{https://openreview.net/forum?id=QVbHQ33R0x}.

\bibitem[Huang et~al.(2022{\natexlab{a}})Huang, Abbeel, Pathak, and Mordatch]{huang2022language}
Wenlong Huang, Pieter Abbeel, Deepak Pathak, and Igor Mordatch.
\newblock {Language models as zero-shot planners: Extracting actionable knowledge for embodied agents}.
\newblock In \emph{International Conference on Machine Learning}, pages 9118--9147, 2022{\natexlab{a}}.

\bibitem[Huang et~al.(2022{\natexlab{b}})Huang, Xia, Xiao, Chan, Liang, Florence, Zeng, Tompson, Mordatch, Chebotar, Sermanet, Brown, Jackson, Luu, Levine, and Hausman]{huang2022inner}
Wenlong Huang, Fei Xia, Ted Xiao, Harris Chan, Jacky Liang, Pete Florence, Andy Zeng, Jonathan Tompson, Igor Mordatch, Yevgen Chebotar, Pierre Sermanet, Noah Brown, Tomas Jackson, Linda Luu, Sergey Levine, and Karol Hausman.
\newblock {Inner Monologue: Embodied Reasoning through Planning with Language Models}.
\newblock In \emph{Conference on Robot Learning}, 2022{\natexlab{b}}.

\bibitem[Mees et~al.(2023)Mees, Borja-Diaz, and Burgard]{mees2023hulc2}
Oier Mees, Jessica Borja-Diaz, and Wolfram Burgard.
\newblock {Grounding Language with Visual Affordances over Unstructured Data}.
\newblock In \emph{IEEE International Conference on Robotics and Automation (ICRA)}, 2023.

\bibitem[Yang et~al.(2023{\natexlab{b}})Yang, Garrett, Lozano-P{\'{e}}rez, Kaelbling, and Fox]{yang2023piginet}
Zhutian Yang, Caelan~Reed Garrett, Tomás Lozano-P{\'{e}}rez, Leslie Kaelbling, and Dieter Fox.
\newblock {Sequence-Based Plan Feasibility Prediction for Efficient Task and Motion Planning}.
\newblock In \emph{Robotics: Science and Systems}, 2023{\natexlab{b}}.

\bibitem[Huang et~al.(2023{\natexlab{d}})Huang, Xia, Shah, Driess, Zeng, Lu, Florence, Mordatch, Levine, Hausman, and Ichter]{huang2023grounded}
Wenlong Huang, Fei Xia, Dhruv Shah, Danny Driess, Andy Zeng, Yao Lu, Pete Florence, Igor Mordatch, Sergey Levine, Karol Hausman, and Brian Ichter.
\newblock {Grounded Decoding: Guiding Text Generation with Grounded Models for Robot Control}.
\newblock In \emph{Thirty-seventh Conference on Neural Information Processing Systems}, 2023{\natexlab{d}}.
\newblock URL \url{https://openreview.net/forum?id=JCCi58IUsh}.

\bibitem[Ding et~al.(2023)Ding, Zhang, Paxton, and Zhang]{ding2023llmgrop}
Yan Ding, Xiaohan Zhang, Chris Paxton, and Shiqi Zhang.
\newblock {Task and Motion Planning with Large Language Models for Object Rearrangement}.
\newblock \emph{arXiv preprint}, 2023.

\bibitem[Lin et~al.(2023{\natexlab{c}})Lin, Agia, Migimatsu, Pavone, and Bohg]{lin2023text2motion}
Kevin Lin, Christopher Agia, Toki Migimatsu, Marco Pavone, and Jeannette Bohg.
\newblock {Text2Motion: From Natural Language Instructions to Feasible Plans}.
\newblock \emph{ICRA 2023 Workshop Pretraining4Robotics}, 2023{\natexlab{c}}.

\bibitem[Wu et~al.(2023{\natexlab{c}})Wu, Min, Bisk, Salakhutdinov, Azaria, Li, Mitchell, and Prabhumoye]{wu2023plan}
Yue Wu, So~Yeon Min, Yonatan Bisk, Ruslan Salakhutdinov, Amos Azaria, Yuanzhi Li, Tom Mitchell, and Shrimai Prabhumoye.
\newblock {Plan, Eliminate, and Track -- Language Models are Good Teachers for Embodied Agents}.
\newblock \emph{arXiv preprint}, 2023{\natexlab{c}}.

\bibitem[Sundaresan et~al.(2023)Sundaresan, Belkhale, Sadigh, and Bohg]{sundaresan2023kite}
Priya Sundaresan, Suneel Belkhale, Dorsa Sadigh, and Jeannette Bohg.
\newblock {KITE: Keypoint-Conditioned Policies for Semantic Manipulation}.
\newblock In \emph{7th Annual Conference on Robot Learning}, 2023.
\newblock URL \url{https://openreview.net/forum?id=veGdf4L4Xz}.

\bibitem[Ren et~al.(2023)Ren, Dixit, Bodrova, Singh, Tu, Brown, Xu, Takayama, Xia, Varley, Xu, Sadigh, Zeng, and Majumdar]{ren2023knowno}
Allen~Z. Ren, Anushri Dixit, Alexandra Bodrova, Sumeet Singh, Stephen Tu, Noah Brown, Peng Xu, Leila Takayama, Fei Xia, Jake Varley, Zhenjia Xu, Dorsa Sadigh, Andy Zeng, and Anirudha Majumdar.
\newblock {Robots That Ask For Help: Uncertainty Alignment for Large Language Model Planners}.
\newblock In \emph{Conference on Robot Learning (CoRL)}, 2023.

\bibitem[Rana et~al.(2023)Rana, Haviland, Garg, Abou-Chakra, Reid, and Suenderhauf]{rana2023sayplan}
Krishan Rana, Jesse Haviland, Sourav Garg, Jad Abou-Chakra, Ian Reid, and Niko Suenderhauf.
\newblock {SayPlan: Grounding Large Language Models using 3D Scene Graphs for Scalable Task Planning}.
\newblock In \emph{7th Annual Conference on Robot Learning}, 2023.
\newblock URL \url{https://openreview.net/forum?id=wMpOMO0Ss7a}.

\bibitem[Ichikura et~al.(2023{\natexlab{a}})Ichikura, Kawaharazuka, Obinata, Okada, and Inaba]{ichikura2023diary}
Aiko Ichikura, Kento Kawaharazuka, Yoshiki Obinata, Kei Okada, and Masayuki Inaba.
\newblock {A method for Selecting Scenes and Emotion-based Descriptions for a Robot's Diary}.
\newblock In \emph{IEEE RO-MAN}, 2023{\natexlab{a}}.

\bibitem[Liu et~al.(2024)Liu, Orru, Paxton, Shafiullah, and Pinto]{liu2024okrobot}
Peiqi Liu, Yaswanth Orru, Chris Paxton, Nur Muhammad~Mahi Shafiullah, and Lerrel Pinto.
\newblock {OK-Robot: What Really Matters in Integrating Open-Knowledge Models for Robotics}.
\newblock \emph{arXiv preprint arXiv:2401.12202}, 2024.

\bibitem[Fang et~al.(2023)Fang, Wang, Fang, Gou, Liu, Yan, Liu, Xie, and Lu]{fang2023anygrasp}
Hao-Shu Fang, Chenxi Wang, Hongjie Fang, Minghao Gou, Jirong Liu, Hengxu Yan, Wenhai Liu, Yichen Xie, and Cewu Lu.
\newblock {AnyGrasp: Robust and Efficient Grasp Perception in Spatial and Temporal Domains}.
\newblock \emph{IEEE Transactions on Robotics (T-RO)}, 2023.

\bibitem[Singh et~al.(2023)Singh, Blukis, Mousavian, Goyal, Xu, Tremblay, Fox, Thomason, and Garg]{singh2023progprompt}
Ishika Singh, Valts Blukis, Arsalan Mousavian, Ankit Goyal, Danfei Xu, Jonathan Tremblay, Dieter Fox, Jesse Thomason, and Animesh Garg.
\newblock {ProgPrompt: Generating Situated Robot Task Plans using Large Language Models}.
\newblock In \emph{2023 IEEE International Conference on Robotics and Automation (ICRA)}, 2023.

\bibitem[Wake et~al.(2023)Wake, Kanehira, Sasabuchi, Takamatsu, and Ikeuchi]{wake2023chatgpt}
Naoki Wake, Atsushi Kanehira, Kazuhiro Sasabuchi, Jun Takamatsu, and Katsushi Ikeuchi.
\newblock {Chatgpt empowered long-step robot control in various environments: A case application}.
\newblock \emph{arXiv preprint}, 2023.

\bibitem[Shirasaka et~al.(2024)Shirasaka, Matsushima, Tsunashima, Ikeda, Horo, Ikoma, Tsuji, Wada, Omija, Komukai, et~al.]{shirasaka2023self}
Mimo Shirasaka, Tatsuya Matsushima, Soshi Tsunashima, Yuya Ikeda, Aoi Horo, So~Ikoma, Chikaha Tsuji, Hikaru Wada, Tsunekazu Omija, Dai Komukai, et~al.
\newblock Self-recovery prompting: Promptable general purpose service robot system with foundation models and self-recovery.
\newblock In \emph{2024 IEEE International Conference on Robotics and Automation (ICRA)}, pages 17395--17402. IEEE, 2024.

\bibitem[Liu et~al.(2023{\natexlab{c}})Liu, Jiang, Zhang, Liu, Zhang, Biswas, and Stone]{liu2023llmplusp}
Bo~Liu, Yuqian Jiang, Xiaohan Zhang, Qiang Liu, Shiqi Zhang, Joydeep Biswas, and Peter Stone.
\newblock {LLM+P: Empowering Large Language Models with Optimal Planning Proficiency}.
\newblock \emph{arXiv preprint}, 2023{\natexlab{c}}.

\bibitem[Aeronautiques et~al.(1998)Aeronautiques, Howe, Knoblock, McDermott, Ram, Veloso, Weld, SRI, Barrett, Christianson, and {others}]{aeronautiques1998pddl}
Constructions Aeronautiques, Adele Howe, Craig Knoblock, I~S I~Drew McDermott, Ashwin Ram, Manuela Veloso, Daniel Weld, David~Wilkins SRI, Anthony Barrett, Dave Christianson, and {others}.
\newblock {PDDL - the planning domain definition language}.
\newblock \emph{Technical Report, Tech. Rep.}, 1998.

\bibitem[Wang et~al.(2023{\natexlab{b}})Wang, Xie, Jiang, Mandlekar, Xiao, Zhu, Fan, and Anandkumar]{wang2023voyager}
Guanzhi Wang, Yuqi Xie, Yunfan Jiang, Ajay Mandlekar, Chaowei Xiao, Yuke Zhu, Linxi Fan, and Anima Anandkumar.
\newblock {Voyager: An Open-Ended Embodied Agent with Large Language Models}.
\newblock \emph{arXiv preprint arXiv:2305.16291}, 2023{\natexlab{b}}.

\bibitem[Vemprala et~al.(2023)Vemprala, Bonatti, Bucker, and Kapoor]{vemprala2023chatgpt}
Sai Vemprala, Rogerio Bonatti, Arthur Bucker, and Ashish Kapoor.
\newblock {ChatGPT for Robotics: Design Principles and Model Abilities}.
\newblock \emph{arXiv preprint}, 2023.

\bibitem[Puig et~al.(2023)Puig, Undersander, Szot, Cote, Yang, Partsey, Desai, Clegg, Hlavac, Min, and {others}]{puig2023habitat}
Xavier Puig, Eric Undersander, Andrew Szot, Mikael~Dallaire Cote, Tsung-Yen Yang, Ruslan Partsey, Ruta Desai, Alexander~William Clegg, Michal Hlavac, So~Yeon Min, and {others}.
\newblock {Habitat 3.0: A Co-Habitat for Humans, Avatars and Robots}.
\newblock \emph{arXiv preprint arXiv:2310.13724}, 2023.

\bibitem[Yoneda et~al.(2023)Yoneda, Fang, Li, Zhang, Jiang, Lin, Picker, Yunis, Mei, and Walter]{yoneda2023statler}
Takuma Yoneda, Jiading Fang, Peng Li, Huanyu Zhang, Tianchong Jiang, Shengjie Lin, Ben Picker, David Yunis, Hongyuan Mei, and Matthew~R. Walter.
\newblock {Statler: State-Maintaining Language Models for Embodied Reasoning}.
\newblock \emph{arXiv preprint}, 2023.

\bibitem[Obinata et~al.(2023)Obinata, Kanazawa, Kawaharazuka, Yanokura, Kim, Okada, and Inaba]{obinata2023gpsr}
Yoshiki Obinata, Naoaki Kanazawa, Kento Kawaharazuka, Iori Yanokura, Soonhyo Kim, Kei Okada, and Masayuki Inaba.
\newblock {Foundation Model based Open Vocabulary Task Planning and Executive System for General Purpose Service Robots}.
\newblock \emph{arXiv preprint}, 2023.

\bibitem[Shirai et~al.(2023)Shirai, Beltran-Hernandez, Hamaya, Hashimoto, Tanaka, Kawaharazuka, Tanaka, Ushiku, and Mori]{shirai2023vision}
Keisuke Shirai, Cristian~C Beltran-Hernandez, Masashi Hamaya, Atsushi Hashimoto, Shohei Tanaka, Kento Kawaharazuka, Kazutoshi Tanaka, Yoshitaka Ushiku, and Shinsuke Mori.
\newblock {Vision-Language Interpreter for Robot Task Planning}.
\newblock \emph{arXiv preprint arXiv:2311.00967}, 2023.

\bibitem[Wang et~al.(2023{\natexlab{c}})Wang, Zhang, Chen, and Sreenath]{wang2023prompt}
Yen-Jen Wang, Bike Zhang, Jianyu Chen, and Koushil Sreenath.
\newblock {Prompt a Robot to Walk with Large Language Models}.
\newblock \emph{arXiv preprint}, 2023{\natexlab{c}}.

\bibitem[Yu et~al.(2023{\natexlab{b}})Yu, Xiao, Stone, Tompson, Brohan, Wang, Singh, Tan, M, Peralta, Ichter, Hausman, and Xia]{yu2023scaling}
Tianhe Yu, Ted Xiao, Austin Stone, Jonathan Tompson, Anthony Brohan, Su~Wang, Jaspiar Singh, Clayton Tan, Dee M, Jodilyn Peralta, Brian Ichter, Karol Hausman, and Fei Xia.
\newblock {Scaling Robot Learning with Semantically Imagined Experience}.
\newblock In \emph{arXiv preprint arXiv:2302.11550}, 2023{\natexlab{b}}.

\bibitem[Brohan et~al.(2023{\natexlab{b}})Brohan, Brown, Carbajal, Chebotar, Dabis, Finn, Gopalakrishnan, Hausman, Herzog, Hsu, and {others}]{brohan2023rt}
Anthony Brohan, Noah Brown, Justice Carbajal, Yevgen Chebotar, Joseph Dabis, Chelsea Finn, Keerthana Gopalakrishnan, Karol Hausman, Alex Herzog, Jasmine Hsu, and {others}.
\newblock {Rt-1: Robotics transformer for real-world control at scale}.
\newblock In \emph{Proceedings of Robotics: Science and Systems}, 2023{\natexlab{b}}.

\bibitem[Wang et~al.(2023{\natexlab{d}})Wang, Saharia, Montgomery, Pont-Tuset, Noy, Pellegrini, Onoe, Laszlo, Fleet, Soricut, and {others}]{wang2023imagen}
Su~Wang, Chitwan Saharia, Ceslee Montgomery, Jordi Pont-Tuset, Shai Noy, Stefano Pellegrini, Yasumasa Onoe, Sarah Laszlo, David~J Fleet, Radu Soricut, and {others}.
\newblock {Imagen editor and editbench: Advancing and evaluating text-guided image inpainting}.
\newblock In \emph{Proceedings of the IEEE/CVF Conference on Computer Vision and Pattern Recognition}, pages 18359--18369, 2023{\natexlab{d}}.

\bibitem[Stone et~al.(2023)Stone, Xiao, Lu, Gopalakrishnan, Lee, Vuong, Wohlhart, Kirmani, Zitkovich, Xia, Finn, and Hausman]{stone2023moo}
Austin Stone, Ted Xiao, Yao Lu, Keerthana Gopalakrishnan, Kuang-Huei Lee, Quan Vuong, Paul Wohlhart, Sean Kirmani, Brianna Zitkovich, Fei Xia, Chelsea Finn, and Karol Hausman.
\newblock {Open-World Object Manipulation using Pre-Trained Vision-Language Models}.
\newblock \emph{arXiv preprint}, 2023.

\bibitem[Sermanet et~al.(2018)Sermanet, Lynch, Chebotar, Hsu, Jang, Schaal, Levine, and Brain]{sermanet2018time}
Pierre Sermanet, Corey Lynch, Yevgen Chebotar, Jasmine Hsu, Eric Jang, Stefan Schaal, Sergey Levine, and Google Brain.
\newblock Time-contrastive networks: Self-supervised learning from video.
\newblock In \emph{2018 IEEE international conference on robotics and automation (ICRA)}, pages 1134--1141. IEEE, 2018.

\bibitem[Xiao et~al.(2022)Xiao, Radosavovic, Darrell, and Malik]{xiao2022masked}
Tete Xiao, Ilija Radosavovic, Trevor Darrell, and Jitendra Malik.
\newblock {Masked visual pre-training for motor control}.
\newblock \emph{arXiv preprint arXiv:2203.06173}, 2022.

\bibitem[Radosavovic et~al.(2022)Radosavovic, Xiao, James, Abbeel, Malik, and Darrell]{radosavovic2022realworld}
Ilija Radosavovic, Tete Xiao, Stephen James, Pieter Abbeel, Jitendra Malik, and Trevor Darrell.
\newblock {Real-World Robot Learning with Masked Visual Pre-training}.
\newblock In \emph{6th Annual Conference on Robot Learning}, 2022.
\newblock URL \url{https://openreview.net/forum?id=KWCZfuqshd}.

\bibitem[He et~al.(2022)He, Chen, Xie, Li, Doll{\'{a}}r, and Girshick]{he2022masked}
Kaiming He, Xinlei Chen, Saining Xie, Yanghao Li, Piotr Doll{\'{a}}r, and Ross Girshick.
\newblock {Masked autoencoders are scalable vision learners}.
\newblock In \emph{Proceedings of the IEEE/CVF conference on computer vision and pattern recognition}, pages 16000--16009, 2022.

\bibitem[Driess et~al.(2023)Driess, Xia, Sajjadi, Lynch, Chowdhery, Ichter, Wahid, Tompson, Vuong, Yu, and {others}]{driess2023palm}
Danny Driess, Fei Xia, Mehdi S~M Sajjadi, Corey Lynch, Aakanksha Chowdhery, Brian Ichter, Ayzaan Wahid, Jonathan Tompson, Quan Vuong, Tianhe Yu, and {others}.
\newblock {Palm-e: An embodied multimodal language model}.
\newblock \emph{arXiv preprint arXiv:2303.03378}, 2023.

\bibitem[Sermanet et~al.(2023)Sermanet, Ding, Zhao, Xia, Dwibedi, Gopalakrishnan, Chan, Dulac-Arnold, Maddineni, Joshi, Florence, Han, Baruch, Lu, Mirchandani, Xu, Sanketi, Hausman, Shafran, Ichter, and Cao]{sermanet2023robovqa}
Pierre Sermanet, Tianli Ding, Jeffrey Zhao, Fei Xia, Debidatta Dwibedi, Keerthana Gopalakrishnan, Christine Chan, Gabriel Dulac-Arnold, Sharath Maddineni, Nikhil~J Joshi, Pete Florence, Wei Han, Robert Baruch, Yao Lu, Suvir Mirchandani, Peng Xu, Pannag Sanketi, Karol Hausman, Izhak Shafran, Brian Ichter, and Yuan Cao.
\newblock {RoboVQA: Multimodal Long-Horizon Reasoning for Robotics}.
\newblock In \emph{2nd Workshop on Language and Robot Learning: Language as Grounding}, 2023.
\newblock URL \url{https://openreview.net/forum?id=z3zBluGh1L}.

\bibitem[Kalashnikov et~al.(2021)Kalashnikov, Varley, Chebotar, Swanson, Jonschkowski, Finn, Levine, and Hausman]{kalashnikov2021scaling}
Dmitry Kalashnikov, Jacob Varley, Yevgen Chebotar, Benjamin Swanson, Rico Jonschkowski, Chelsea Finn, Sergey Levine, and Karol Hausman.
\newblock {Scaling Up Multi-Task Robotic Reinforcement Learning}.
\newblock \emph{5th Annual Conference on Robot Learning}, 2021.
\newblock URL \url{https://openreview.net/forum?id=p9Pe-l9MMEq}.

\bibitem[Jang et~al.(2021)Jang, Irpan, Khansari, Kappler, Ebert, Lynch, Levine, and Finn]{jang2022bc}
Eric Jang, Alex Irpan, Mohi Khansari, Daniel Kappler, Frederik Ebert, Corey Lynch, Sergey Levine, and Chelsea Finn.
\newblock {BC-Z: Zero-Shot Task Generalization with Robotic Imitation Learning}.
\newblock In \emph{5th Annual Conference on Robot Learning}, 2021.
\newblock URL \url{https://openreview.net/forum?id=8kbp23tSGYv}.

\bibitem[Reed et~al.(2022)Reed, Zolna, Parisotto, Colmenarejo, Novikov, Barth-Maron, Gimenez, Sulsky, Kay, Springenberg, and {others}]{reed2022generalist}
Scott Reed, Konrad Zolna, Emilio Parisotto, Sergio~Gomez Colmenarejo, Alexander Novikov, Gabriel Barth-Maron, Mai Gimenez, Yury Sulsky, Jackie Kay, Jost~Tobias Springenberg, and {others}.
\newblock {A generalist agent}.
\newblock \emph{Transactions on Machine Learning Research}, 2022.
\newblock URL \url{https://openreview.net/forum?id=1ikK0kHjvj}.

\bibitem[Chebotar et~al.(2023)Chebotar, Vuong, Hausman, Xia, Lu, Irpan, Kumar, Yu, Herzog, Pertsch, Gopalakrishnan, Ibarz, Nachum, Sontakke, Salazar, Tran, Peralta, Tan, Manjunath, Singh, Zitkovich, Jackson, Rao, Finn, and Levine]{chebotar2023qtransformer}
Yevgen Chebotar, Quan Vuong, Karol Hausman, Fei Xia, Yao Lu, Alex Irpan, Aviral Kumar, Tianhe Yu, Alexander Herzog, Karl Pertsch, Keerthana Gopalakrishnan, Julian Ibarz, Ofir Nachum, Sumedh~Anand Sontakke, Grecia Salazar, Huong~T Tran, Jodilyn Peralta, Clayton Tan, Deeksha Manjunath, Jaspiar Singh, Brianna Zitkovich, Tomas Jackson, Kanishka Rao, Chelsea Finn, and Sergey Levine.
\newblock {Q-Transformer: Scalable Offline Reinforcement Learning via Autoregressive Q-Functions}.
\newblock In \emph{7th Annual Conference on Robot Learning}, 2023.

\bibitem[Bousmalis et~al.(2023)Bousmalis, Vezzani, Rao, Devin, Lee, Bauza, Davchev, Zhou, Gupta, Raju, and {others}]{bousmalis2023robocat}
Konstantinos Bousmalis, Giulia Vezzani, Dushyant Rao, Coline Devin, Alex~X Lee, Maria Bauza, Todor Davchev, Yuxiang Zhou, Agrim Gupta, Akhil Raju, and {others}.
\newblock {RoboCat: A Self-Improving Foundation Agent for Robotic Manipulation}.
\newblock \emph{arXiv preprint arXiv:2306.11706}, 2023.

\bibitem[Schubert et~al.(2023)Schubert, Zhang, Bruce, Bechtle, Parisotto, Riedmiller, Springenberg, Byravan, Hasenclever, and Heess]{schubert2023generalist}
Ingmar Schubert, Jingwei Zhang, Jake Bruce, Sarah Bechtle, Emilio Parisotto, Martin Riedmiller, Jost~Tobias Springenberg, Arunkumar Byravan, Leonard Hasenclever, and Nicolas Heess.
\newblock {A Generalist Dynamics Model for Control}.
\newblock \emph{arXiv preprint arXiv:2305.10912}, 2023.

\bibitem[Kelly et~al.(2019)Kelly, Sidrane, Driggs-Campbell, and Kochenderfer]{kelly2019hg}
Michael Kelly, Chelsea Sidrane, Katherine Driggs-Campbell, and Mykel~J Kochenderfer.
\newblock {Hg-dagger: Interactive imitation learning with human experts}.
\newblock In \emph{2019 International Conference on Robotics and Automation (ICRA)}, pages 8077--8083, 2019.

\bibitem[Jiang et~al.(2023)Jiang, Gupta, Zhang, Wang, Dou, Chen, Fei-Fei, Anandkumar, Zhu, and Fan]{jiang2023vima}
Yunfan Jiang, Agrim Gupta, Zichen Zhang, Guanzhi Wang, Yongqiang Dou, Yanjun Chen, Li~Fei-Fei, Anima Anandkumar, Yuke Zhu, and Linxi Fan.
\newblock {VIMA: Robot Manipulation with Multimodal Prompts}.
\newblock In \emph{Fortieth International Conference on Machine Learning}, 2023.

\bibitem[Chen et~al.(2023{\natexlab{c}})Chen, Djolonga, Padlewski, Mustafa, Changpinyo, Wu, Ruiz, Goodman, Wang, Tay, and {others}]{chen2023pali}
Xi~Chen, Josip Djolonga, Piotr Padlewski, Basil Mustafa, Soravit Changpinyo, Jialin Wu, Carlos~Riquelme Ruiz, Sebastian Goodman, Xiao Wang, Yi~Tay, and {others}.
\newblock {PaLI-X: On Scaling up a Multilingual Vision and Language Model}.
\newblock \emph{arXiv preprint arXiv:2305.18565}, 2023{\natexlab{c}}.

\bibitem[Gu et~al.(2023{\natexlab{b}})Gu, Kirmani, Wohlhart, Lu, Arenas, Rao, Yu, Fu, Gopalakrishnan, Xu, and {others}]{gu2023rt}
Jiayuan Gu, Sean Kirmani, Paul Wohlhart, Yao Lu, Montserrat~Gonzalez Arenas, Kanishka Rao, Wenhao Yu, Chuyuan Fu, Keerthana Gopalakrishnan, Zhuo Xu, and {others}.
\newblock {RT-Trajectory: Robotic Task Generalization via Hindsight Trajectory Sketches}.
\newblock \emph{arXiv preprint arXiv:2311.01977}, 2023{\natexlab{b}}.

\bibitem[Anonymous(2024)]{anonymous2024rtsketch}
Anonymous.
\newblock {RT}-sketch: Goal-conditioned imitation learning from hand-drawn sketches, 2024.
\newblock URL \url{https://openreview.net/forum?id=YxvmODVWny}.

\bibitem[Ahn et~al.(2024)Ahn, Dwibedi, Finn, Arenas, Gopalakrishnan, Hausman, Ichter, Irpan, Joshi, Julian, et~al.]{ahn2024autort}
Michael Ahn, Debidatta Dwibedi, Chelsea Finn, Montse~Gonzalez Arenas, Keerthana Gopalakrishnan, Karol Hausman, Brian Ichter, Alex Irpan, Nikhil Joshi, Ryan Julian, et~al.
\newblock Autort: Embodied foundation models for large scale orchestration of robotic agents.
\newblock \emph{arXiv preprint arXiv:2401.12963}, 2024.

\bibitem[Leal et~al.(2023)Leal, Choromanski, Jain, Dubey, Varley, Ryoo, Lu, Liu, Sindhwani, Vuong, et~al.]{leal2023sara}
Isabel Leal, Krzysztof Choromanski, Deepali Jain, Avinava Dubey, Jake Varley, Michael Ryoo, Yao Lu, Frederick Liu, Vikas Sindhwani, Quan Vuong, et~al.
\newblock Sara-rt: Scaling up robotics transformers with self-adaptive robust attention.
\newblock \emph{arXiv preprint arXiv:2312.01990}, 2023.

\bibitem[{Octo Model Team} et~al.(2023){Octo Model Team}, Ghosh, Walke, Pertsch, Black, Mees, Dasari, Hejna, Xu, Luo, Kreiman, Tan, Sadigh, Finn, and Levine]{octo_2023}
{Octo Model Team}, Dibya Ghosh, Homer Walke, Karl Pertsch, Kevin Black, Oier Mees, Sudeep Dasari, Joey Hejna, Charles Xu, Jianlan Luo, Tobias Kreiman, {You Liang} Tan, Dorsa Sadigh, Chelsea Finn, and Sergey Levine.
\newblock Octo: An open-source generalist robot policy.
\newblock \url{https://octo-models.github.io}, 2023.

\bibitem[Chi et~al.(2023)Chi, Feng, Du, Xu, Cousineau, Burchfiel, and Song]{chi2023diffusion}
Cheng Chi, Siyuan Feng, Yilun Du, Zhenjia Xu, Eric Cousineau, Benjamin Burchfiel, and Shuran Song.
\newblock Diffusion policy: Visuomotor policy learning via action diffusion.
\newblock \emph{arXiv preprint arXiv:2303.04137}, 2023.

\bibitem[Mnih et~al.(2015)Mnih, Kavukcuoglu, Silver, Rusu, Veness, Bellemare, Graves, Riedmiller, Fidjeland, Ostrovski, Petersen, Beattie, Sadik, Antonoglou, King, Kumaran, Wierstra, Legg, and Hassabis]{mnih2015atari}
Volodymyr Mnih, Koray Kavukcuoglu, David Silver, Andrei~A. Rusu, Joel Veness, Marc~G. Bellemare, Alex Graves, Martin Riedmiller, Andreas~K. Fidjeland, Georg Ostrovski, Stig Petersen, Charles Beattie, Amir Sadik, Ioannis Antonoglou, Helen King, Dharshan Kumaran, Daan Wierstra, Shane Legg, and Demis Hassabis.
\newblock {Human-level control through deep reinforcement learning}.
\newblock \emph{Nature}, 518\penalty0 (7540):\penalty0 529--533, 2015.

\bibitem[Tassa et~al.(2018)Tassa, Doron, Muldal, Erez, Li, Casas, Budden, Abdolmaleki, Merel, Lefrancq, and {others}]{tassa2018deepmind}
Yuval Tassa, Yotam Doron, Alistair Muldal, Tom Erez, Yazhe Li, Diego de~Las Casas, David Budden, Abbas Abdolmaleki, Josh Merel, Andrew Lefrancq, and {others}.
\newblock {Deepmind control suite}.
\newblock \emph{arXiv preprint arXiv:1801.00690}, 2018.

\bibitem[Geng et~al.(2019)Geng, Cao, and Tulyakov]{geng20193d}
Zhenglin Geng, Chen Cao, and Sergey Tulyakov.
\newblock {3D Guided Fine-Grained Face Manipulation}.
\newblock In \emph{Proceedings of the IEEE/CVF conference on computer vision and pattern recognition}, pages 9821--9830, 2019.

\bibitem[Gu et~al.(2017)Gu, Holly, Lillicrap, and Levine]{gu2017deep}
Shixiang Gu, Ethan Holly, Timothy Lillicrap, and Sergey Levine.
\newblock {Deep Reinforcement Learning for Robotic Manipulation}.
\newblock In \emph{2017 IEEE international conference on robotics and automation (ICRA)}, pages 3389--3396, 2017.

\bibitem[Yang et~al.(2023{\natexlab{c}})Yang, Du, Ghasemipour, Tompson, Schuurmans, and Abbeel]{yang2023learning}
Mengjiao Yang, Yilun Du, Kamyar Ghasemipour, Jonathan Tompson, Dale Schuurmans, and Pieter Abbeel.
\newblock {Learning Interactive Real-World Simulators}.
\newblock \emph{arXiv preprint arXiv:2310.06114}, 2023{\natexlab{c}}.

\bibitem[Jaegle et~al.(2022)Jaegle, Borgeaud, Alayrac, Doersch, Ionescu, Ding, Koppula, Zoran, Brock, Shelhamer, and {others}]{jaegle2022perceiver}
Andrew Jaegle, Sebastian Borgeaud, Jean-Baptiste Alayrac, Carl Doersch, Catalin Ionescu, David Ding, Skanda Koppula, Daniel Zoran, Andrew Brock, Evan Shelhamer, and {others}.
\newblock {Perceiver io: A general architecture for structured inputs {\&} outputs}.
\newblock In \emph{International Conference on Learning Representations}, 2022.
\newblock URL \url{https://openreview.net/forum?id=fILj7WpI-g}.

\bibitem[Shridhar et~al.(2022)Shridhar, Manuelli, and Fox]{shridhar2023perceiver}
Mohit Shridhar, Lucas Manuelli, and Dieter Fox.
\newblock {Perceiver-Actor: A Multi-Task Transformer for Robotic Manipulation}.
\newblock In \emph{6th Annual Conference on Robot Learning}, 2022.
\newblock URL \url{https://openreview.net/forum?id=PS_eCS_WCvD}.

\bibitem[Perez et~al.(2018)Perez, Strub, De~Vries, Dumoulin, and Courville]{perez2018film}
Ethan Perez, Florian Strub, Harm De~Vries, Vincent Dumoulin, and Aaron Courville.
\newblock {Film: Visual reasoning with a general conditioning layer}.
\newblock In \emph{Proceedings of the AAAI conference on artificial intelligence}, volume~32, 2018.

\bibitem[Bharadhwaj et~al.(2023)Bharadhwaj, Vakil, Sharma, Gupta, Tulsiani, and Kumar]{bharadhwaj2023roboagent}
Homanga Bharadhwaj, Jay Vakil, Mohit Sharma, Abhinav Gupta, Shubham Tulsiani, and Vikash Kumar.
\newblock {Roboagent: Generalization and efficiency in robot manipulation via semantic augmentations and action chunking}.
\newblock \emph{arXiv preprint arXiv:2309.01918}, 2023.

\bibitem[Tan and Le(2019)]{tan2019efficientnet}
Mingxing Tan and Quoc Le.
\newblock Efficientnet: Rethinking model scaling for convolutional neural networks.
\newblock In \emph{International conference on machine learning}, pages 6105--6114. PMLR, 2019.

\bibitem[Ryoo et~al.(2021)Ryoo, Piergiovanni, Arnab, Dehghani, and Angelova]{ryoo2021tokenlearner}
Michael Ryoo, A~J Piergiovanni, Anurag Arnab, Mostafa Dehghani, and Anelia Angelova.
\newblock {Tokenlearner: Adaptive space-time tokenization for videos}.
\newblock \emph{Advances in Neural Information Processing Systems}, 34:\penalty0 12786--12797, 2021.

\bibitem[Dasari et~al.(2019)Dasari, Ebert, Tian, Nair, Bucher, Schmeckpeper, Singh, Levine, and Finn]{dasari2019robonet}
Sudeep Dasari, Frederik Ebert, Stephen Tian, Suraj Nair, Bernadette Bucher, Karl Schmeckpeper, Siddharth Singh, Sergey Levine, and Chelsea Finn.
\newblock {Robonet: Large-scale multi-robot learning}.
\newblock In \emph{Conference on Robot Learning (CoRL)}, 2019.

\bibitem[Fu et~al.(2020)Fu, Kumar, Nachum, Tucker, and Levine]{fu2020d4rl}
Justin Fu, Aviral Kumar, Ofir Nachum, George Tucker, and Sergey Levine.
\newblock {D4rl: Datasets for deep data-driven reinforcement learning}.
\newblock \emph{arXiv preprint arXiv:2004.07219}, 2020.

\bibitem[Kumar et~al.(2020)Kumar, Zhou, Tucker, and Levine]{kumar2020conservative}
Aviral Kumar, Aurick Zhou, George Tucker, and Sergey Levine.
\newblock {Conservative q-learning for offline reinforcement learning}.
\newblock \emph{Advances in Neural Information Processing Systems}, 33:\penalty0 1179--1191, 2020.

\bibitem[Yu et~al.(2020)Yu, Thomas, Yu, Ermon, Zou, Levine, Finn, and Ma]{yu2020mopo}
Tianhe Yu, Garrett Thomas, Lantao Yu, Stefano Ermon, James~Y Zou, Sergey Levine, Chelsea Finn, and Tengyu Ma.
\newblock {Mopo: Model-based offline policy optimization}.
\newblock \emph{Advances in Neural Information Processing Systems}, 33:\penalty0 14129--14142, 2020.

\bibitem[Matsushima et~al.(2021)Matsushima, Furuta, Matsuo, Nachum, and Gu]{matsushima2021deploymentefficient}
Tatsuya Matsushima, Hiroki Furuta, Yutaka Matsuo, Ofir Nachum, and Shixiang Gu.
\newblock {Deployment-efficient reinforcement learning via model-based offline optimization}.
\newblock In \emph{International Conference on Learning Representations}, 2021.
\newblock URL \url{https://openreview.net/forum?id=3hGNqpI4WS}.

\bibitem[Ebert et~al.(2022)Ebert, Yang, Schmeckpeper, Bucher, Georgakis, Daniilidis, Finn, and Levine]{ebert2022bridge}
Frederik Ebert, Yanlai Yang, Karl Schmeckpeper, Bernadette Bucher, Georgios Georgakis, Kostas Daniilidis, Chelsea Finn, and Sergey Levine.
\newblock {Bridge data: Boosting generalization of robotic skills with cross-domain datasets}.
\newblock In \emph{Robotics: Science and Systems}, 2022.

\bibitem[Walke et~al.(2023)Walke, Black, Lee, Kim, Du, Zheng, Zhao, Hansen-Estruch, Vuong, He, and {others}]{walke2023bridgedata}
Homer Walke, Kevin Black, Abraham Lee, Moo~Jin Kim, Max Du, Chongyi Zheng, Tony Zhao, Philippe Hansen-Estruch, Quan Vuong, Andre He, and {others}.
\newblock {Bridgedata v2: A dataset for robot learning at scale}.
\newblock In \emph{Conference on Robot Learning (CoRL)}, 2023.

\bibitem[Zhao et~al.(2023)Zhao, Kumar, Levine, and Finn]{zhao2023learning}
Tony~Z Zhao, Vikash Kumar, Sergey Levine, and Chelsea Finn.
\newblock Learning fine-grained bimanual manipulation with low-cost hardware.
\newblock \emph{arXiv preprint arXiv:2304.13705}, 2023.

\bibitem[Fu et~al.(2024)Fu, Zhao, and Finn]{fu2024mobile}
Zipeng Fu, Tony~Z Zhao, and Chelsea Finn.
\newblock {Mobile ALOHA}: Learning bimanual mobile manipulation with low-cost whole-body teleoperation.
\newblock \emph{arXiv preprint arXiv:2401.02117}, 2024.

\bibitem[Wu et~al.(2023{\natexlab{d}})Wu, Shentu, Yi, Lin, and Abbeel]{wu2023gello}
Philipp Wu, Yide Shentu, Zhongke Yi, Xingyu Lin, and Pieter Abbeel.
\newblock {GELLO}: A general, low-cost, and intuitive teleoperation framework for robot manipulators.
\newblock \emph{arXiv preprint arXiv:2309.13037}, 2023{\natexlab{d}}.

\bibitem[Sutton and Barto(2018)]{Sutton2018}
Richard~S Sutton and Andrew~G Barto.
\newblock \emph{{Reinforcement learning: An introduction}}.
\newblock 2018.

\bibitem[Carroll et~al.(2022)Carroll, Paradise, Lin, Georgescu, Sun, Bignell, Milani, Hofmann, Hausknecht, Dragan, and {others}]{carroll2022uni}
Micah Carroll, Orr Paradise, Jessy Lin, Raluca Georgescu, Mingfei Sun, David Bignell, Stephanie Milani, Katja Hofmann, Matthew Hausknecht, Anca Dragan, and {others}.
\newblock {Uni [mask]: Unified inference in sequential decision problems}.
\newblock \emph{Advances in neural information processing systems}, 35:\penalty0 35365--35378, 2022.

\bibitem[Wu et~al.(2023{\natexlab{e}})Wu, Majumdar, Stone, Lin, Mordatch, Abbeel, and Rajeswaran]{wu2023masked}
Philipp Wu, Arjun Majumdar, Kevin Stone, Yixin Lin, Igor Mordatch, Pieter Abbeel, and Aravind Rajeswaran.
\newblock {Masked Trajectory Models for Prediction, Representation, and Control}.
\newblock In \emph{Proceedings of the 40th International Conference on Machine Learning}, volume 202 of \emph{Proceedings of Machine Learning Research}, pages 37607--37623. PMLR, 1 2023{\natexlab{e}}.

\bibitem[Hatori et~al.(2018)Hatori, Kikuchi, Kobayashi, Takahashi, Tsuboi, Unno, Ko, and Tan]{hatori2018interactively}
Jun Hatori, Yuta Kikuchi, Sosuke Kobayashi, Kuniyuki Takahashi, Yuta Tsuboi, Yuya Unno, Wilson Ko, and Jethro Tan.
\newblock Interactively picking real-world objects with unconstrained spoken language instructions.
\newblock In \emph{2018 IEEE International Conference on Robotics and Automation (ICRA)}, pages 3774--3781. IEEE, 2018.

\bibitem[Taniguchi et~al.(2016)Taniguchi, Nagai, Nakamura, Iwahashi, Ogata, and Asoh]{taniguchi2016symbol}
Tadahiro Taniguchi, Takayuki Nagai, Tomoaki Nakamura, Naoto Iwahashi, Tetsuya Ogata, and Hideki Asoh.
\newblock Symbol emergence in robotics: a survey.
\newblock \emph{Advanced Robotics}, 30\penalty0 (11-12):\penalty0 706--728, 2016.

\bibitem[Shah et~al.(2023{\natexlab{a}})Shah, Sridhar, Bhorkar, Hirose, and Levine]{shah2022gnm}
Dhruv Shah, Ajay Sridhar, Arjun Bhorkar, Noriaki Hirose, and Sergey Levine.
\newblock {GNM: A General Navigation Model to Drive Any Robot}.
\newblock In \emph{International Conference on Robotics and Automation (ICRA)}, 2023{\natexlab{a}}.

\bibitem[Shah et~al.(2023{\natexlab{b}})Shah, Sridhar, Dashora, Stachowicz, Black, Hirose, and Levine]{shah2023vint}
Dhruv Shah, Ajay Sridhar, Nitish Dashora, Kyle Stachowicz, Kevin Black, Noriaki Hirose, and Sergey Levine.
\newblock {ViNT: A Foundation Model for Visual Navigation}.
\newblock In \emph{7th Annual Conference on Robot Learning}, 2023{\natexlab{b}}.

\bibitem[Yoshikawa et~al.(2023)Yoshikawa, Skreta, Darvish, Arellano-Rubach, Ji, Bj{\o}rn~Kristensen, Li, Zhao, Xu, Kuramshin, and {others}]{yoshikawa2023chemistry}
Naruki Yoshikawa, Marta Skreta, Kourosh Darvish, Sebastian Arellano-Rubach, Zhi Ji, Lasse Bj{\o}rn~Kristensen, Andrew~Zou Li, Yuchi Zhao, Haoping Xu, Artur Kuramshin, and {others}.
\newblock {Large language models for chemistry robotics}.
\newblock \emph{Autonomous Robots}, 47:\penalty0 1057--1086, 2023.

\bibitem[Kawaharazuka et~al.(2023{\natexlab{b}})Kawaharazuka, Obinata, Kanazawa, Okada, and Inaba]{kawaharazuka2023ptvlm}
Kento Kawaharazuka, Yoshiki Obinata, Naoaki Kanazawa, Kei Okada, and Masayuki Inaba.
\newblock {Robotic Applications of Pre-Trained Vision-Language Models to Various Recognition Behaviors}.
\newblock \emph{arXiv preprint arXiv:2303.05674}, 2023{\natexlab{b}}.

\bibitem[Yenamandra et~al.(2023)Yenamandra, Ramachandran, Yadav, Wang, Khanna, Gervet, Yang, Jain, Clegg, Turner, Kira, Savva, Chang, Chaplot, Batra, Mottaghi, Bisk, and Paxton]{yenamandra2023homerobot}
Sriram Yenamandra, Arun Ramachandran, Karmesh Yadav, Austin~S Wang, Mukul Khanna, Theophile Gervet, Tsung-Yen Yang, Vidhi Jain, Alexander Clegg, John~M Turner, Zsolt Kira, Manolis Savva, Angel~X Chang, Devendra~Singh Chaplot, Dhruv Batra, Roozbeh Mottaghi, Yonatan Bisk, and Chris Paxton.
\newblock {HomeRobot: Open-Vocabulary Mobile Manipulation}.
\newblock In \emph{7th Annual Conference on Robot Learning}, 2023.
\newblock URL \url{https://openreview.net/forum?id=b-cto-fetlz}.

\bibitem[Ichikura et~al.(2023{\natexlab{b}})Ichikura, Kawaharazuka, Obinata, Shinjo, Okada, and Inaba]{ichikura2023automatic}
Aiko Ichikura, Kento Kawaharazuka, Yoshiki Obinata, Koki Shinjo, Kei Okada, and Masayuki Inaba.
\newblock {Automatic Diary Generation System including Information on Joint Experiences between Humans and Robots}.
\newblock \emph{arXiv preprint arXiv:2309.01948}, 2023{\natexlab{b}}.

\bibitem[Cho and Nam(2023)]{cho2023story}
Hyungjun Cho and Tek-Jin Nam.
\newblock {The story of Beau: Exploring the potential of generative diaries in shaping social perceptions of robots}.
\newblock \emph{International Journal of Design}, 17\penalty0 (1):\penalty0 1--15, 2023.

\bibitem[Chang et~al.(2023)Chang, Gervet, Khanna, Yenamandra, Shah, Min, Shah, Paxton, Gupta, Batra, and {others}]{chang2023goat}
Matthew Chang, Theophile Gervet, Mukul Khanna, Sriram Yenamandra, Dhruv Shah, So~Yeon Min, Kavit Shah, Chris Paxton, Saurabh Gupta, Dhruv Batra, and {others}.
\newblock {Goat: Go to any thing}.
\newblock \emph{arXiv preprint arXiv:2311.06430}, 2023.

\bibitem[Mees et~al.(2022)Mees, Hermann, Rosete-Beas, and Burgard]{mees2022calvin}
Oier Mees, Lukas Hermann, Erick Rosete-Beas, and Wolfram Burgard.
\newblock {Calvin: A benchmark for language-conditioned policy learning for long-horizon robot manipulation tasks}.
\newblock \emph{IEEE Robotics and Automation Letters}, 7\penalty0 (3):\penalty0 7327--7334, 2022.

\bibitem[Gong et~al.(2023)Gong, Huang, Zhao, Geng, Gao, Wu, Ai, Zhou, Terzopoulos, Zhu, and {others}]{gong2023arnold}
Ran Gong, Jiangyong Huang, Yizhou Zhao, Haoran Geng, Xiaofeng Gao, Qingyang Wu, Wensi Ai, Ziheng Zhou, Demetri Terzopoulos, Song-Chun Zhu, and {others}.
\newblock {ARNOLD: A Benchmark for Language-Grounded Task Learning With Continuous States in Realistic 3D Scenes}.
\newblock In \emph{Proceedings of the IEEE/CVF International Conference on Computer Vision (ICCV)}, 2023.

\end{thebibliography}
\bibliographystyle{unsrtnat}

\end{document}